\documentclass[twoside,11pt]{article}

\usepackage{jmlr2e}
\usepackage{subcaption}
\usepackage{booktabs}       

\usepackage{amsfonts,epsfig,graphicx}
\usepackage{amsmath,amssymb}

\usepackage{epsf}
\usepackage{epstopdf}
\usepackage{graphics}
\usepackage{amsfonts}
\usepackage{amsmath}
\usepackage{psfrag}
\usepackage{color}
\usepackage{mathtools}
\usepackage[colorinlistoftodos]{todonotes} 
\usepackage{hyperref}
\usepackage[margin=1in]{geometry}
\newcommand{\Exs}{\ensuremath{\mathbb{E}}}

\newcommand{\f}{\ensuremath{f}}
\DeclareMathOperator{\trace}{trace}
\newcommand{\conv}{\operatorname{conv}}

\newtheorem{lem}{Lemma}[section]

\newtheorem{nota}{Notation}[section]
\newtheorem{de}{Definition}[section]
\newtheorem{exa}{Example}[section]
\newtheorem{as}{Assumption}[section]
\newtheorem{alg}{Algorithm}[section]

\newcommand{\btheo}{\begin{theos}}
\newcommand{\bde}{\begin{de}}
\newcommand{\ble}{\begin{lem}}
\newcommand{\bpr}{\begin{props}}
\newcommand{\bno}{\begin{nota}}
\newcommand{\bex}{\begin{exa}}
\newcommand{\bcor}{\begin{cors}}
\newcommand{\spro}{\begin{proof}}
\newcommand{\bas}{\begin{as}}
\newcommand{\balg}{\begin{alg}}
\newcommand{\bremark}{\begin{remark}}

\newcommand{\etheo}{\end{theos}}
\newcommand{\ede}{\end{de}}
\newcommand{\ele}{\end{lem}}
\newcommand{\epr}{\end{props}}
\newcommand{\eno}{\end{nota}}
\newcommand{\eex}{\end{exa}}
\newcommand{\ecor}{\end{cors}}
\newcommand{\fpro}{\end{proof}}
\newcommand{\eas}{\end{as}}
\newcommand{\ealg}{\end{alg}}
\newcommand{\eremark}{\end{remark}}
\newcommand{\reals}{\mathbb{R}}

\newtheorem{theos}{Theorem}
\newtheorem{props}{Proposition}
\newtheorem{lems}{Lemma}
\newtheorem{cors}{Corollary}
\newtheorem{rems}{Remark}

\newtheorem{exas}{Example}
\newtheorem{algs}{Algorithm}
\newtheorem{asss}{Assumption}
\newtheorem{defns}{Definition}

\newcommand{\btheos}{\begin{theos}}
\newcommand{\etheos}{\end{theos}}
\newcommand{\brems}{\begin{rems}}
\newcommand{\erems}{\end{rems}}
\newcommand{\bprops}{\begin{props}}
\newcommand{\eprops}{\end{props}}
\newcommand{\bdes}{\begin{defns}}
\newcommand{\edes}{\end{defns}}
\newcommand{\blems}{\begin{lems}}
\newcommand{\elems}{\end{lems}}
\newcommand{\bcors}{\begin{cors}}
\newcommand{\ecors}{\end{cors}}
\newcommand{\bexs}{\begin{exas}}
\newcommand{\eexs}{\end{exas}}
\newcommand{\balgs}{\begin{algs}}
\newcommand{\ealgs}{\end{algs}}
\newcommand{\bass}{\begin{asss}}
\newcommand{\eass}{\end{asss}}
\newcommand{\bit}{\begin{itemize}}
\newcommand{\eit}{\end{itemize}}

\newcommand{\data}{{\bf A}}
\newcommand{\sample}{\vec{a}}
\newcommand{\samplescalar}{a}
\newcommand{\secondw}{w}
\newcommand{\secondwvec}{\vec{w}}
\newcommand{\secondwmat}{\vec{W}}
\newcommand{\firstw}{\vec{u}}
\newcommand{\firstwscalar}{u}
\newcommand{\firstwmat}{\vec{U}}
\newcommand{\bias}{b}
\newcommand{\biasvec}{\vec{b}}
\newcommand{\dual}{\vec{v}}
\newcommand{\dualmat}{\vec{V}}

\newcommand{\dualscalar}{v}

\newcommand{\rectset}{\mathcal{Q}_\data}
\newcommand{\relu}[1]{\big( #1 \big)_+}
\newcommand{\ball}{\mathcal{B}}
\usepackage{algorithm}
\usepackage{algpseudocode}
 \DeclareMathOperator*{\argmin}{\arg\!\min} 
  \DeclareMathOperator*{\argmax}{\arg\!\max} 
\newcommand{\sign}{\text{sign}}
\let\vec\mathbf
\usepackage{hyperref}
\usepackage{url}

\usepackage[toc,page,header]{appendix}
\usepackage{minitoc}

\usepackage{bbm}

\ShortHeadings{Convex Geometry and Duality of Over-parameterized Neural Networks}{Ergen and Pilanci}
\firstpageno{1}

\usepackage{lastpage}
\jmlrheading{22}{2021}{1-\pageref{LastPage}}{12/20; Revised
7/21}{8/21}{20-1447}{Tolga Ergen and Mert Pilanci}
\ShortHeadings{Convex Geometry and Duality of Over-parameterized Neural Networks}{Ergen and Pilanci}

\begin{document}

\doparttoc 
\faketableofcontents 

\title{Convex Geometry and Duality of Over-parameterized Neural Networks}

\author{\name Tolga Ergen \email ergen@stanford.edu \\
       \addr Department of Electrical Engineering\\
       Stanford University\\
       Stanford, CA 94305, USA
       \AND
    \name Mert Pilanci \email pilanci@stanford.edu \\
       \addr Department of Electrical Engineering\\
       Stanford University\\
       Stanford, CA 94305, USA}
\editor{Julien Mairal}   
\maketitle

\begin{abstract}
We develop a convex analytic approach to analyze finite width two-layer ReLU networks. We first prove that an optimal solution to the regularized training problem can be characterized as extreme points of a convex set, where simple solutions are encouraged via its convex geometrical properties. We then leverage this characterization to show that an optimal set of parameters yield linear spline interpolation for regression problems involving one dimensional or rank-one data. We also characterize the classification decision regions in terms of a kernel matrix and minimum $\ell_1$-norm solutions. This is in contrast to Neural Tangent Kernel which is unable to explain predictions of finite width networks. Our convex geometric characterization also provides intuitive explanations of hidden neurons as auto-encoders. In higher dimensions, we show that the training problem can be cast as a finite dimensional convex problem with infinitely many constraints. Then, we apply certain convex relaxations and introduce a cutting-plane algorithm to globally optimize the network. We further analyze the exactness of the relaxations to provide conditions for the convergence to a global optimum. Our analysis also shows that optimal network parameters can be also characterized as interpretable closed-form formulas in some practically relevant special cases.

\end{abstract}

%

 \section{Introduction}

Over-parameterized Deep Neural Networks (DNNs) have attracted significant attention due to their powerful representation and generalization capabilities. Recent studies empirically observed that NNs with ReLU activation achieve simple solutions as a result of training (see e.g., \citet{quantized_hartmut,infinite_width}), although a full theoretical understanding is yet to be developed. Particularly, \citep{infinite_width} showed that among two-layer ReLU networks that perfectly fit one dimensional training data, i.e., $d=1$, the minimum Euclidean norm ReLU network is a linear spline interpolator. Therefore, training over-parameterized networks with standard weight decay induces a bias towards simple solutions, which may result in good generalization performance for $d=1$. This work later extended to multi-variate functions ($d>1$) in \cite{Ongie2020A}, however, the later work lacks an explicit characterization for the optimal solutions. Therefore, in the general case $d>1$, characterizing the structure of optimal solutions and understanding the fundamental mechanism behind this implicit bias remain an open problem.

In this paper, we develop a convex analytic framework to reveal a fundamental convex geometric mechanism behind the bias towards simple solutions. More specifically, we show that over parameterized networks achieve simple solutions as the \emph{extreme points} of a certain convex set, where simplicity is enforced by an implicit regularizer analogous to $\ell_1$-norm regularization that promotes sparsity through extreme points of the unit $\ell_1$-ball, i.e., the cross-polytope. However,  unlike the conventional $\ell_1$-norm regularization, extreme points are data-adaptive and can be interpreted as \emph{convex autoenconders}. In the paper, we provide a complete characterization for extreme points via exact analytical expressions. As a corollary, for one dimensional and rank-one regression and classifications tasks, we prove that extreme points are in a specific form that yields linear spline interpolations, which explains the recent empirical observations in the $d=1$ case. We also extend this analysis to higher dimensions ($d>1$) to obtain exact characterizations or even closed-form solutions for the network parameters in some practically relevant cases.

\subsection{Related work}
\citet{quantized_hartmut,implicit_reg_blanc,understanding_zhang} previously studied the dynamics of ReLU networks with finite neurons. \citet{understanding_zhang} specifically showed that NNs are implicitly regularized so that training with Stochastic Gradient Descent (SGD) converges to small norm solutions. Later, \citet{implicit_reg_blanc} further elaborated the previous studies and proved that in a one dimensional case, SGD finds a solution that is linear over any set of three or more co-linear data points. Additionally, \citet{quantized_hartmut} proved that initialization magnitude of network parameters has a strong connection with implicit regularization. The authors further showed that in the regime where implicit regularization is effective, i.e., when initialization magnitude is small, network parameters align along certain directions characterized by the input data points. This observation shows that in fact there exist finitely many simple (or regularized) functions for a given training dataset. \citet{lazy_training_bach} then proved that ReLU networks converge to a point that generalizes when initialization magnitude is small, i.e., called active training. Otherwise, parameters do not tend to vary and stay very close their initialization so that network does not generalize as well as in the small initialization case, which is also known as lazy training. 

Another line of research in \citet{bengio2006convex,margin_theory_tengyu,bach2017breaking,lazy_training_bach}
studied infinitely wide two-layer ReLU networks. In particular, \citet{bengio2006convex} introduced a convex algorithm to train infinite width two-layer NNs. Although infinite dimensional training problems are not practical in higher dimensions, the analysis may shed light into the generalization properties. \citet{margin_theory_tengyu} proved that over-parameterization improves generalization bounds by analyzing weakly regularized NNs. In addition to this, recently, the connection between infinite width NNs and kernel methods has attracted significant attention \citep{ntk_jacot,arora_cntk}. Such kernel based methods, nowadays known as Neural Tangent Kernel (NTK), work in a regime where parameters barely change after initialization, coined the lazy regime, so that the dynamics of an NN training problem can be characterized via a deterministic fixed kernel matrix. Therefore, these studies showed that an NN trained with GD and infinitesimal step size in the lazy regime is equivalent to a kernel predictor with a fixed kernel matrix. 

Convexity of infinitely wide two-layer networks and kernel approximation is attractive due to the analytical tractability of convex optimization and tools from convex geometry, although these characterizations fall short of explaining the practical success of finite width networks. In \citet{ergen2019convexshallow,bartan2019convex} convex relaxations of one layer ReLU networks were studied, which have approximation guarantees under certain data assumptions. These architectures have a limited representation power due to the lack of a second layer.

\subsection{Our contributions}
Our contributions can be summarized as follows.
\begin{itemize}
    \item We develop a convex analytic framework for two-layer ReLU NNs with weight decay, i.e., $\ell_2^2$ regularization to provide a deeper insight into over-parameterization and implicit regularization. We prove that over-parameterized ReLU NNs behave like convex regularizers, which encourage simple solutions as the extreme points of a convex set which is termed \emph{rectified ellipsoids}. The polar dual of a rectified ellipsoid is a convex body that determines the optimal hidden layer weights in a similar spirit to the extreme points of an $\ell_1$-ball and its polar dual $\ell_{\infty}$-ball. We further show that the rectified ellipsoid is a data-dependent regularizer whose extreme points act as autoencoders (Lemma \ref{lemma:convex_mixtures} and Lemma \ref{lemma:extreme_general_closedform}).
    
    \item As a corollary of our analysis, we show that optimal NNs that perfectly fit the training data outputs a linear spline interpolation for one dimensional or rank-one data. We also derive a general characterization for the hidden layer weights in higher dimensions in terms of a representer theorem to develop convex geometric insights (Corollary \ref{cor:representer}).
    
    \item Using our convex analytic framework, we characterize the set of optimal solutions in some specific cases so that the training problem can be reformulated as a convex optimization problem. We also show that there can be multiple globally optimal NNs with minimal $\ell_2^2$ regularization, but leading to different predictions in these cases (Proposition \ref{prop:1d_uniqueness} and Figure \ref{fig:uniquness}).
    
    \item For whitened data matrices, we provide exact closed form expressions of the optimal first and second layer weights by leveraging the convex duality (Theorems \ref{theo:closedform_regularized}, \ref{theo:closedform_2neuron_hinge} and \ref{theo:closedform_regularized_multiclass}). These expressions exhibit an interesting thresholding effect in a similar spirit to soft-thresholding of $\ell_1$ penalty. 
 
    \item Based on our convex analytic description, we propose training algorithms relying on convex relaxations of the rectified ellipsoid set. We further prove that the relaxations are tight in certain regimes when a convex geometric condition holds, including whitened  and i.i.d random training data, and the algorithm globally optimizes the network. 
    
    \item We establish a connection between $\ell_0$-$\ell_1$ equivalence in compressed sensing and the training problem for ReLU networks (Lemma \ref{lemma:l1-l0 equivalence}). Using this connection, we then obtain closed-form solutions for the optimal ReLU network parameters in certain practically relevant cases.

    \item We leverage our convex analytic characterization to design convex optimization based training methods that perform well in standard datasets and validate our theoretical results. In contrast to standard non-convex training methods, our methods provide transparent and interpretable means to train neural models.
    
\end{itemize}

\begin{figure*}[t]
	\centering
	\captionsetup[subfigure]{oneside}
		\begin{subfigure}[t]{0.475\textwidth}
		\centering
		\includegraphics[width=1\textwidth, height=0.7\textwidth]{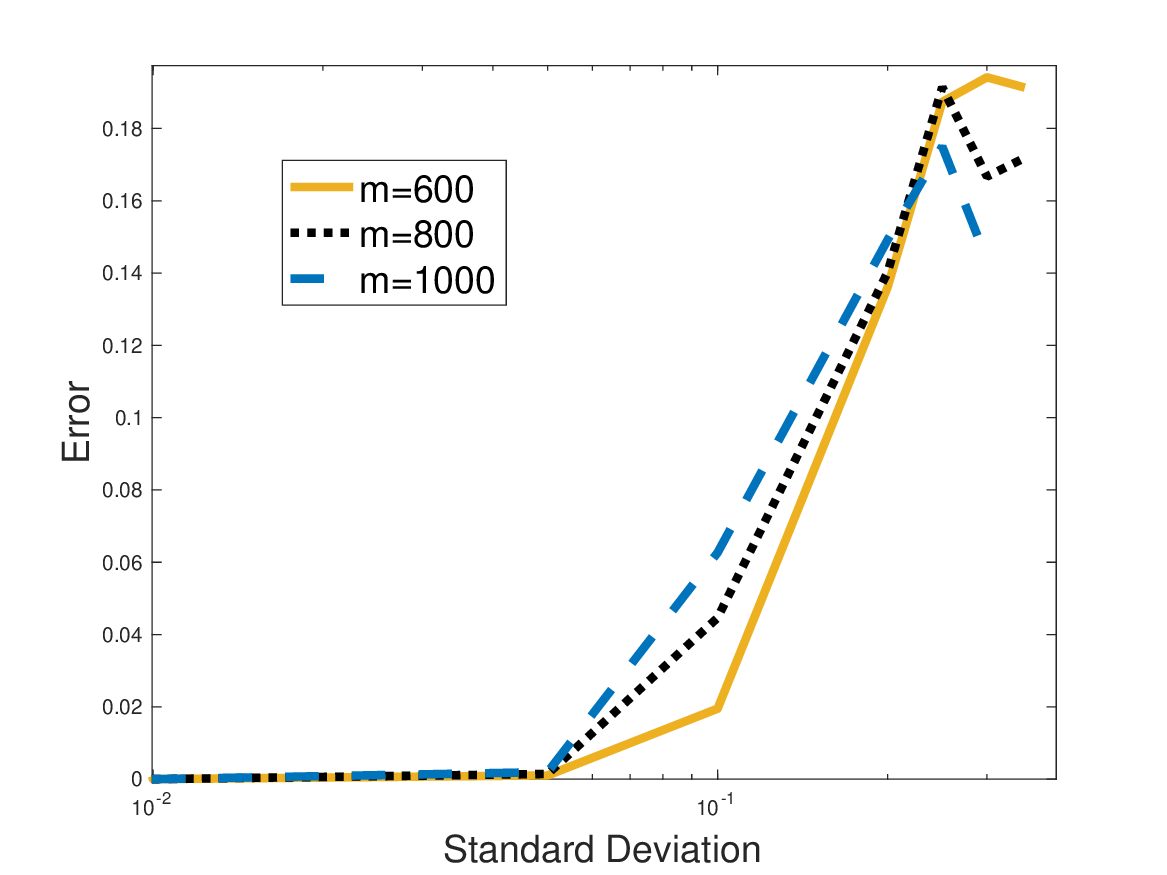}
		\caption{Deviation of the ReLU network output from piecewise linear spline vs standard deviation of initialization plotted for different number of hidden neurons $m$.} \label{fig:sigma}
	\end{subfigure} \hspace*{\fill}
			\begin{subfigure}[t]{0.475\textwidth}
		\centering
		\includegraphics[width=1\textwidth, height=0.7\textwidth]{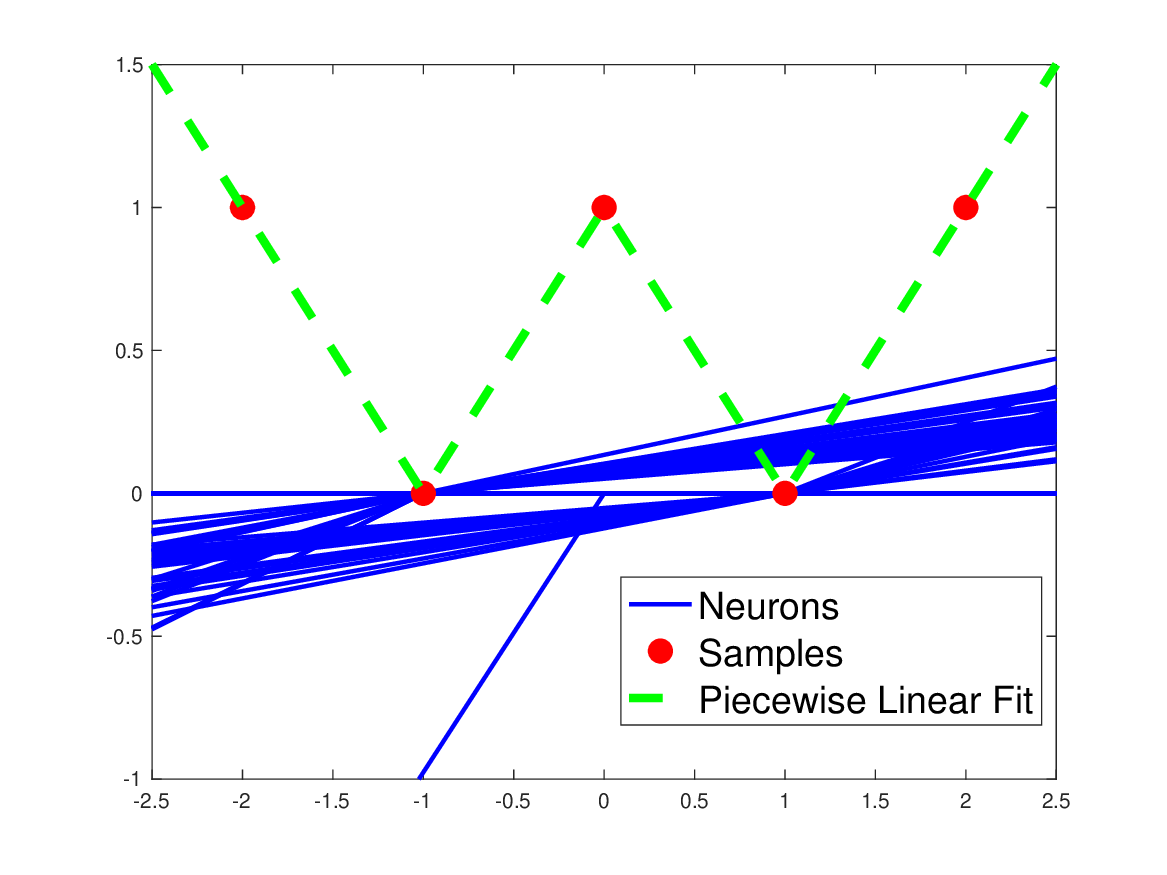}
		\caption{Contribution of each neuron along with the overall fit. Each activation point corresponds to a particular data sample.} \label{fig:neuron}
	\end{subfigure} \hspace*{\fill}
		\begin{subfigure}[t]{0.475\textwidth}
		\centering
		\includegraphics[width=1\textwidth, height=0.7\textwidth]{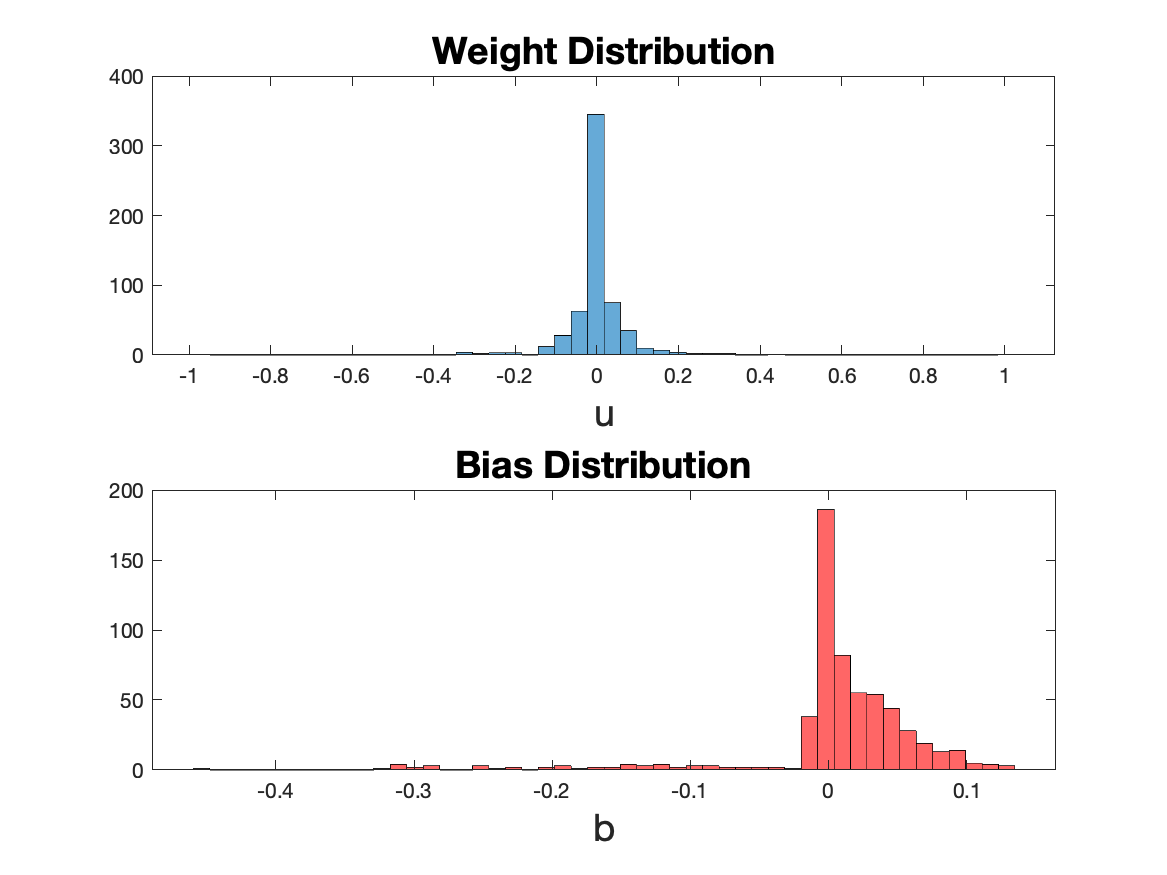}
		\caption{Weight and bias distributions for the network in Figure \ref{fig:neuron}.} \label{fig:hist}
	\end{subfigure} \hspace*{\fill}
		\begin{subfigure}[t]{0.475\textwidth}
		\centering
		\includegraphics[width=1\textwidth, height=0.7\textwidth]{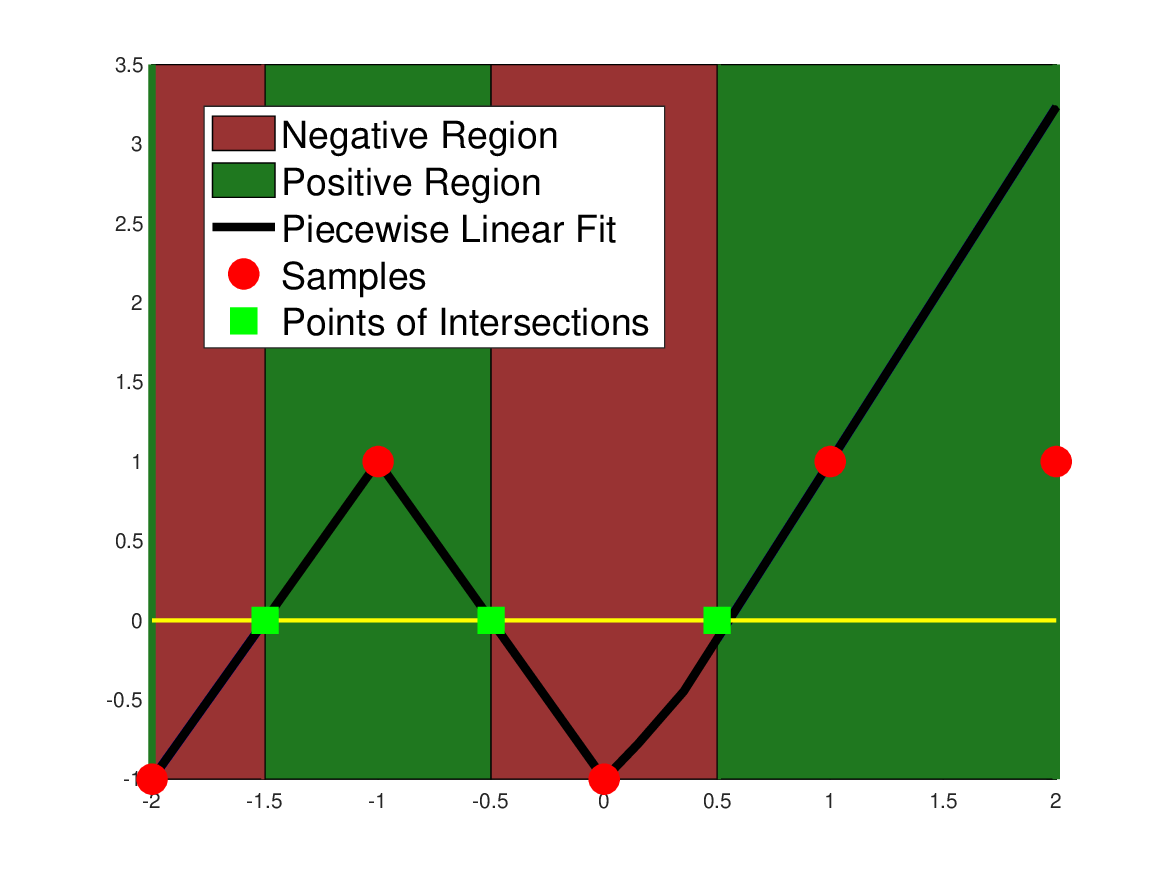}
		\caption{Binary classification using hinge loss. Network output is a linear spline interpolation, and decision regions are determined by zero crossings (see Lemma \ref{lemma:extreme_l2loss}).} \label{fig:hinge}
	\end{subfigure} \hspace*{\fill}
	\medskip
	\caption{Analysis of one dimensional regression and classification with a two-layer NN.}\label{figs_intro}
	\end{figure*}

\subsection{Overview of our results}
In order to understand the effects of initialization magnitude on implicit regularization, we train a two-layer ReLU network on one dimensional training data depicted in Figure \ref{fig:neuron} with different initialization magnitudes. In Figure \ref{fig:sigma}, we observe that the resulting optimal network output is linear spline interpolation when the initialization magnitude (i.e., the standard deviation of random initializations) is small, which matches with the empirical observations in recent work \citet{quantized_hartmut,lazy_training_bach}. We also provide the function fit by each neuron in Figure \ref{fig:neuron} in the case of small initialization magnitude. Here, we remark that the kink of each ReLU neuron, i.e., the point where the output of ReLU is exactly zero, completely aligns with one of the input data points, which is consistent with the alignment behavior observed in \citet{quantized_hartmut}. We also note that even though the weights and biases might take quite different values for each neuron as illustrated in Figure \ref{fig:hist}, their activation points (or kinks) correspond to the data samples.  The same analysis and conclusions also apply to binary classification scenarios with hinge loss as illustrated in Figure \ref{fig:hinge}. In this case, the resulting optimal networks are specific piecewise linear functions with kinks only at data points that determine the decision boundaries as the zero crossings. Based on these observations, the central questions we address in this paper are: \emph{Why are over-parameterized ReLU NNs fitting a linear spline interpolation for one dimensional datasets? Is there a general mechanism inducing simple solutions in higher dimensions?} In the sequel, we show that these questions can be addressed by our convex analytic framework based on \emph{convex geometry} and \emph{duality}.

Here, we show that the optimal ReLU network fits a linear spline interpolation whose kinks at the input data points because the convex approximation\footnote{Here $\samplescalar_i$ is an arbitrary data sample, and $\mathcal{S}$ is an arbitrary subset of data points. $\lambda_1,\ldots,\lambda_n$ are mixture weights, approximating $\samplescalar_i$ as a convex mixture of the input data points in $\mathcal{S}\backslash\{i\}$.} of a data point $\samplescalar_i$ given by 
\begin{align*}
    \min_{\boldsymbol{\lambda}\succcurlyeq 0,\sum_{j}\lambda_j=1} \left| \samplescalar_i-\sum_{j\in \mathcal{S},j\neq i} \lambda_j \samplescalar_j \right|
\end{align*}
is given by another data point, i.e., an extreme point of the convex hull of data points in $\mathcal{S}\backslash\{i\}$. Consequently,  input data points are optimal hidden neuron activation thresholds for one dimensional ReLU networks. Similar characterizations also extend to the hidden neurons in multivariate cases as detailed below.

 We also provide a \emph{representer theorem} for the optimal neurons in a general two-layer NN. In particular, in a finite training dataset with samples $\sample_1,\ldots,\sample_n \in \real^d$, the hidden neurons $\firstw_1,\ldots,\firstw_m\in\real^d$ obey
\begin{align*}
    \firstw_{j}=\sum_{i}\alpha_i (\sample_i-\sample_k) \text{ and } \bias=-\sample_k^T \firstw_{j} ~\forall j\in[m], 
\end{align*}
for some weight vector $\boldsymbol{\alpha}$ and index $k\in[n]$.

\textbf{Notation: }Matrices and vectors are denoted as uppercase and lowercase bold letters, respectively. $\vec{I}_k$ denotes the identity matrix of the size $k$. We use $(x)_+=\max\{x,0\}$ for the ReLU activation function. Furthermore, the set of integers from $1$ to $n$ are denoted as $[n]$ and the notation $\vec{e}_j$ is used for the $j^{th}$ ordinary basis vector. We also use $\ball_2$ to denote the $\ell_2$ unit ball in $\mathbb{R}^d$, i.e., $\ball_2=\{\vec{u} \in \mathbb{R}^d \;\vert \;\|\vec{u}\|_2 \leq 1\}$.

\subsection{Organization of the Paper}
The organization of this paper is as follows. In Section \ref{sec:main_prelims}, we first describe the problem setting with the required preliminary concepts and then define notions of spike-free matrices and extreme points. Based on the definitions in Section \ref{sec:main_prelims}, we then state our main results using convex duality in Section \ref{sec:main_results}, where we also analyze some special cases, e.g., rank-one/whitened data, and introduce a training algorithm to globally optimize two-layer ReLU networks. We extend these results to various cases with regularization, vector outputs, arbitrary convex loss functions in Section \ref{sec:main_regconvex}. Here, we also provide closed-form solutions and/or equivalent convex optimization formulations for regularized ReLU network training problems. In Section \ref{sec:main_ntk}, we briefly review the recently introduced NTK characterization and analytically compare it with our exact characterization. Then, Section \ref{sec:numerical_exps} follows with numerical experiments on both synthetic and real benchmark datasets to verify our analysis in the previous sections. Finally, we conclude the paper with some remarks and future research directions in  Section \ref{sec:main_conclusion}.

\section{Preliminaries}\label{sec:main_prelims}
Given $n$ data samples, i.e., $\{ \sample_i\}_{i=1}^n, \sample_i \in \mathbb{R}^d$, we consider two-layer NNs with $m$ hidden neurons and ReLU activations.  Initially, we focus on the scalar output case for simplicity, i.e., \footnote{We assume that the bias term for the output layer is zero without loss of generality, since we can still recover the general case as illustrated in \citet{quantized_hartmut}.}
\begin{align}
    f(\data) =\sum_{j=1}^m \secondw_j (\data \firstw_j+\bias_j \vec{1})_{+}, \label{eq:2layer_function}
\end{align}
where $\data \in \mathbb{R}^{n \times d}$ is the data matrix, $\firstw_j \in \mathbb{R}^d$ and $\bias_j \in \mathbb{R}$ are the parameters of the $j^{th}$ hidden neuron, and $\secondw_j \in \mathbb{R}$ is the corresponding output layer weight. For a more compact representation, we also define $\firstwmat \in \mathbb{R}^{d \times m}$, $\biasvec \in \mathbb{R}^m$, and $\secondwvec \in \mathbb{R}^m$ as the hidden layer weight matrix, the bias vector, and the output layer weight vector, respectively. Thus, \eqref{eq:2layer_function} can be written as $f(\data)=(\data\firstwmat + \vec{1} \biasvec^T)_{+}\secondwvec$.\footnote{We defer the discussion of the more general vector output case to Section \ref{sec:vectoroutput}.}

Given the data matrix $\data$ and the label vector $\vec{y} \in \mathbb{R}^n$, consider training the network by solving the following optimization problem
\begin{align}
    \min_{\secondw_j,\firstw_j, \bias_j} \big \| \sum_{j=1}^m \secondw_j(\data \firstw_j +\bias_j \vec{1})_+ -\vec{y} \big\|_2^2 +\beta \sum_{j=1}^m (\|\vec{u}_j\|_2^2+w_j^2)\,, \label{eq:problem_statement}
\end{align}
where $\beta$ is a regularization parameter. 
We define the overall parameter space $\Theta$ for \eqref{eq:2layer_function} as
$    \theta \in \Theta= \{(\firstwmat, \biasvec,\secondwvec,m )\,\vert\, \firstwmat \in \mathbb{R}^{d \times m}, \biasvec \in \mathbb{R}^m, \secondwvec \in \mathbb{R}^m, m \in \mathbb{Z}_+\}$. Based on our observations in Figure \ref{fig:sigma} and the results in \citet{infinite_width,lazy_training_bach,neyshabur_reg,parhi_minimum}, we first focus on a minimum norm variant of \eqref{eq:problem_statement}\footnote{This corresponds to a weak form of regularization as $\beta\rightarrow 0$ in \eqref{eq:problem_statement}.}. We define the squared Euclidean norm of the weights (without biases) as
$R(\theta)=\|\secondwvec\|_2^2+\|\firstwmat\|_F^2$. Then we consider the following optimization problem
\begin{align}
    & \min_{\theta \in \Theta} R(\theta) \text{ s.t. } f_{\theta}(\data)=\vec{y}, \label{eq:problem_def1}
\end{align}
where the over-parameterization allows us to reach zero training error over $\data$ via the ReLU network in \eqref{eq:2layer_function}. 
The next lemma shows that the minimum squared Euclidean norm is equivalent to minimum $\ell_1$-norm after a suitable rescaling. 
\begin{lems}\footnote{Proofs are presented in Appendix \ref{sec:proofs}.} \label{lemma:reg_equivalence}
The following two optimization problems are equivalent:
\begin{align*}
  &P^*=\min_{\theta \in \Theta} R(\theta)  \text{ s.t. } f_{\theta}(\data)=\vec{y} \hspace{1cm} = \hspace{1cm}\min_{\theta \in \Theta} \| \secondwvec \|_1 \text{ s.t. } f_{\theta}(\data)=\vec{y}, \| \firstw_j \|_2=1 , \forall j.
\end{align*}
\end{lems}
\begin{lems}
\label{lemma:constraint}
Replacing $\| \firstw_j \|_2=1$ with $\| \firstw_j \|_2 \leq 1$ does not change the value of the above problem.
\end{lems}
By Lemma \ref{lemma:reg_equivalence} and \ref{lemma:constraint}, we can express \eqref{eq:problem_def1} as
\begin{align}
   P^*= \min_{\theta \in \Theta} \| \secondwvec \|_1 \text{ s.t. } f_{\theta}(\data)=\vec{y}, \| \firstw_j \|_2 \leq 1 , \forall j.\label{eq:problem_def2}
\end{align}
However, both \eqref{eq:problem_statement} and \eqref{eq:problem_def2} are quite challenging optimization problems due to the optimization over hidden neurons and the ReLU activation. In particular, depending on the properties of $\data$, e.g., singular values, rank, and dimensions, the landscape of the non-convex objective in \eqref{eq:problem_statement} can be quite complex.

\subsection{Geometry of a single ReLU neuron in the function space} \label{sec:geometry1}
\begin{figure*}[t]
	\centering
	\captionsetup[subfigure]{oneside}
		\begin{subfigure}[t]{0.32\textwidth}
		\centering
		\includegraphics[width=1\textwidth, height=0.8\textwidth]{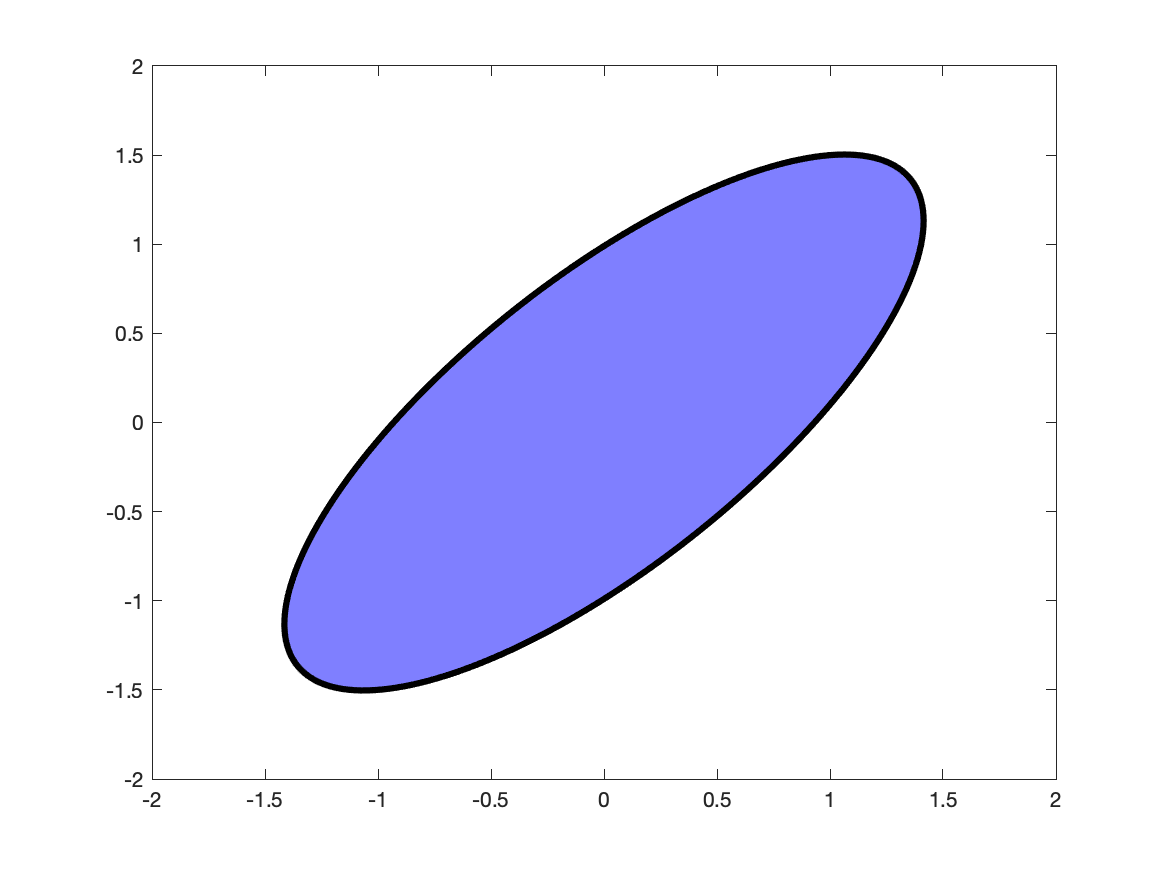}
		\caption{Ellipsoidal set: $\{\data\vec{u}\,\vert\,\firstw \in \mathbb{R}^d, \| \firstw\|_2 \leq 1\}$\centering} \label{fig:nospike_ellipsoid}
	\end{subfigure} \hspace*{\fill}
		\begin{subfigure}[t]{0.32\textwidth}
		\centering
		\includegraphics[width=1\textwidth, height=0.8\textwidth]{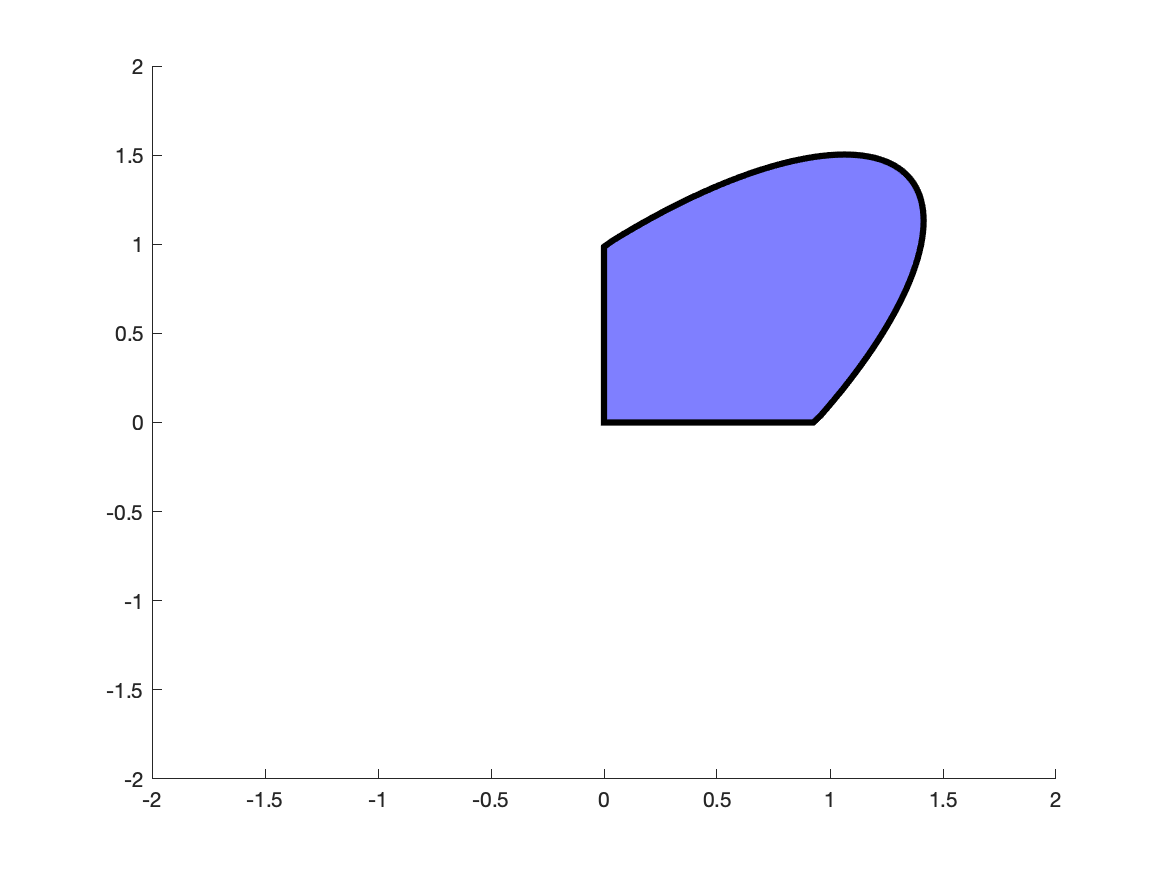}
		\caption{Rectified ellipsoidal set $\rectset$: $\big \{ \relu{ \data \vec{u} } \, | \firstw \in \mathbb{R}^d, \| \firstw\|_2 \leq 1   \big\}$\centering} \label{fig:nospike_recellipsoid}
	\end{subfigure} \hspace*{\fill}
			\begin{subfigure}[t]{0.32\textwidth}
		\centering
		\includegraphics[width=1\textwidth, height=0.8\textwidth]{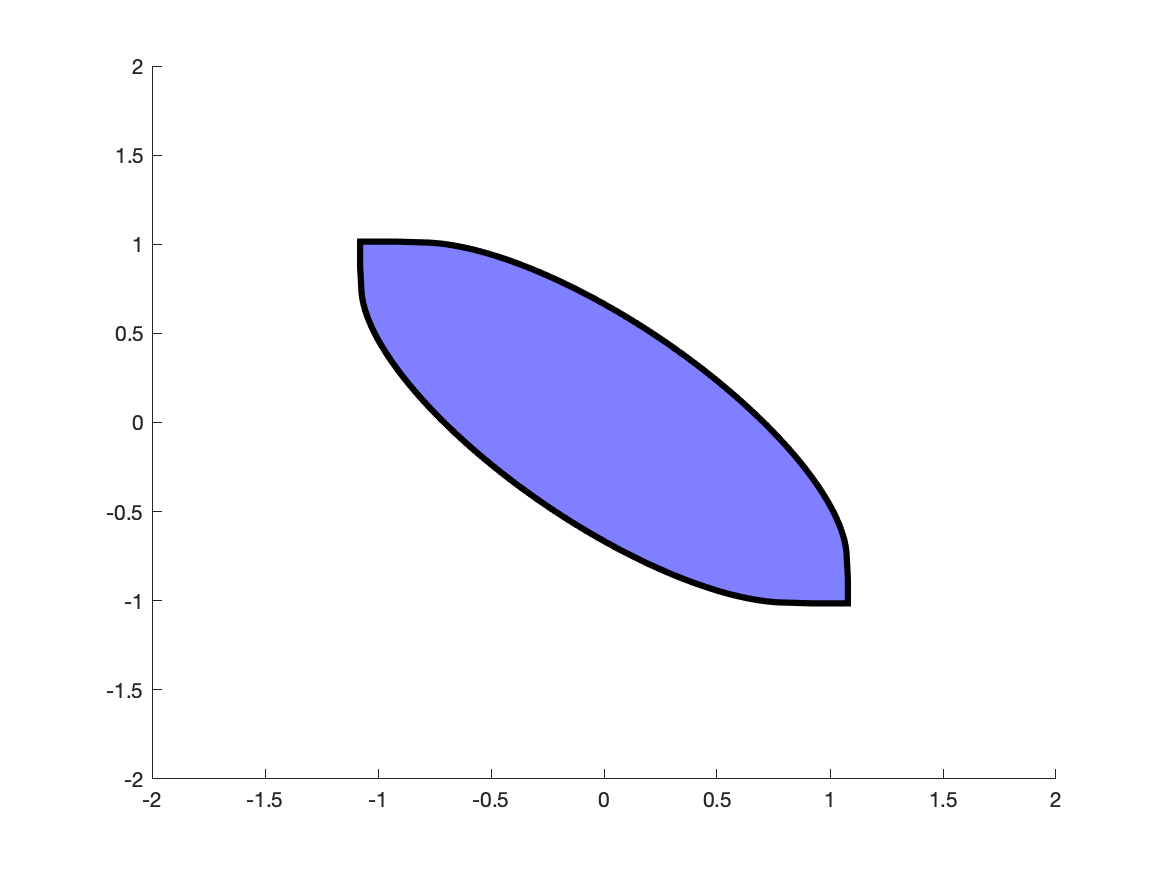}
		\caption{Polar set $\rectset^\circ$: $\{\dual | \dual^T \firstw\le 1\, \forall \firstw\in \rectset\}$\centering} \label{fig:nospike_polar}
	\end{subfigure} \hspace*{\fill}
	\medskip
	\caption{Two dimensional illustration of a spike-free case. Extreme points (spikes) induce the linear spline interpolation behavior in Figures \ref{fig:neuron} and \ref{fig:hinge} as predicted by our theory (see Lemma \ref{lemma:convex_mixtures}). The set shown in the middle figure acts as a regularizer analogous to a non-convex atomic norm.}
\label{fig:nospike}
	\end{figure*}

In order to illustrate the geometry of \eqref{eq:problem_statement}, we particularly focus on a simple case where we have a single neuron with no bias and regularization, i.e., $m=1$, $\bias_1=0$, and $\beta=0$. Thus, \eqref{eq:problem_statement} reduces to 
\begin{align}
      \min_{\firstw_1,\secondw_1} \big \|  \secondw_1 (\data \firstw_1)_+ -\vec{y} \big\|_2^2 \text{ s.t. }  \| \firstw_1\|_2 \leq 1. \label{eq:problem_def_geo}
\end{align}
The solution of \eqref{eq:problem_def_geo} is completely determined by the set $\rectset=\{(\data \firstw )_+ | \firstw \in \mathbb{R}^d, \| \firstw\|_2 \leq 1 \}$. It is evident that \eqref{eq:problem_def_geo} is solved via scaling this set by $|w_1|$ to minimize the distance to $+\vec{y}$ or $-\vec{y}$, depending on the sign of $w_1$. We note that since $\|\firstw\|_2 \leq 1$ describes a $d$-dimensional unit ball, $\data \firstw$ describes an ellipsoid whose shape and orientation is determined by the singular values and the output singular vectors of $\data$ as illustrated in Figure \ref{fig:nospike}. 

\subsection{Rectified ellipsoid and its geometric properties}
A central object in our analysis is the rectified ellipsoidal set introduced in the previous section, which is defined as 
$    \rectset = \big \{ \relu{ \data \vec{u} } \, \vert \firstw \in \mathbb{R}^d, \| \firstw\|_2 \leq 1   \big\}\,.
$
The set $\rectset$ is non-convex in general, as depicted in Figure \ref{fig:spike1}, \ref{fig:spike2}, and \ref{fig:3dspikenew}. However, there exists a family of data matrices $\data$ for which the set $\rectset$ is convex as illustrated in Figure \ref{fig:nospike}, e.g.,  diagonal data matrices. We note that the aforementioned set of matrices are, in fact, a more general class.
\subsubsection{Spike-free matrices}
We say that a matrix $\data$ is spike-free if it holds that
$
\rectset = \data \ball_2 \cap \reals_+^n\,,
$
where $\data\ball_2=\{\data\vec{u}\,\vert\,\vec{u}\in\ball_2\}$, and $\ball_2$ is the unit $\ell_2$ ball. Note that $\rectset$ is a convex set if $\data$ is spike-free. In this case we have an efficient description of this set given by $\rectset=\{ \data \vec{u} | \firstw \in \mathbb{R}^d,\|\vec{u}\|_2\le 1,\, \data\vec{u}\ge 0\}$.

\begin{figure*}[t]
	\centering
	\captionsetup[subfigure]{oneside}
		\begin{subfigure}[t]{0.32\textwidth}
		\centering
		\includegraphics[width=1\textwidth, height=0.8\textwidth]{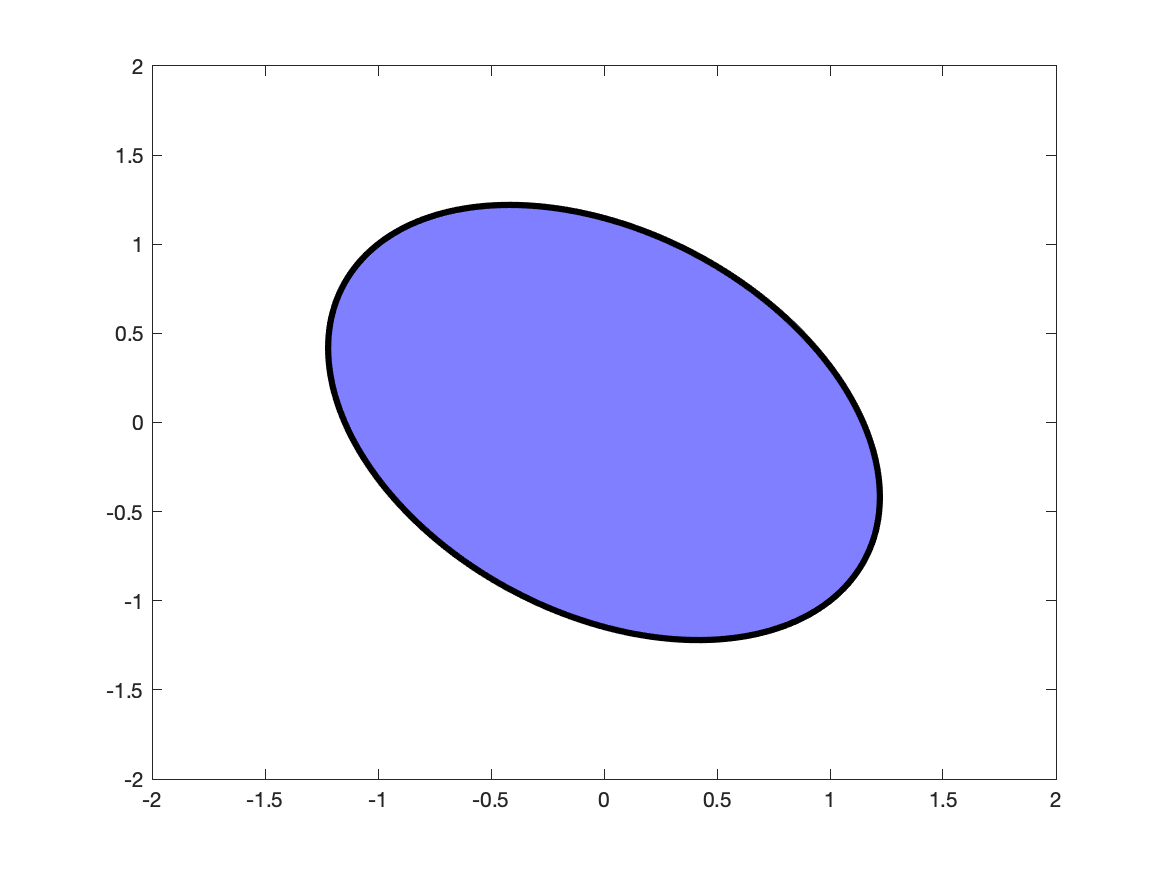}
		\caption{Ellipsoidal set: $\{\data\vec{u}\,\vert\,\firstw \in \mathbb{R}^d, \| \firstw\|_2 \leq 1\}$\centering} \label{fig:spike1_ellipsoid}
	\end{subfigure} \hspace*{\fill}
		\begin{subfigure}[t]{0.32\textwidth}
		\centering
		\includegraphics[width=1\textwidth, height=0.8\textwidth]{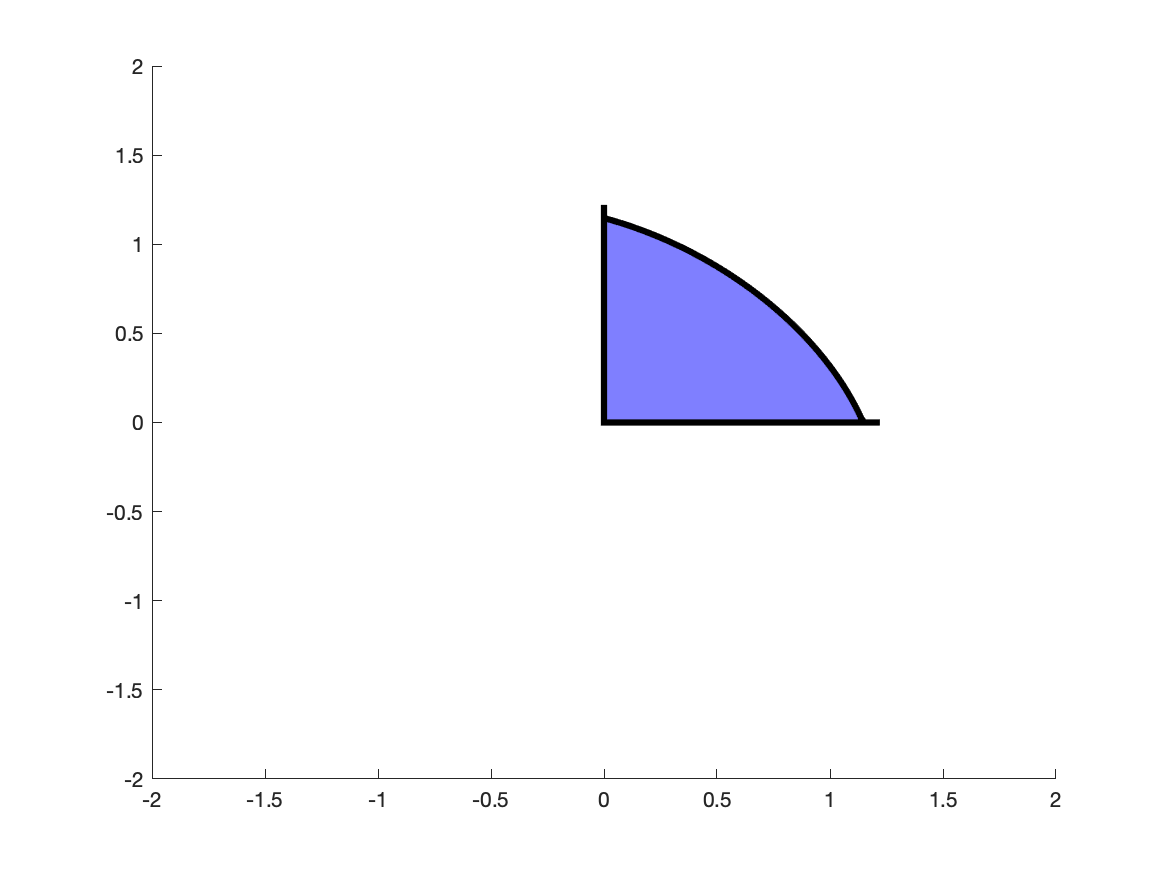}
	\caption{Rectified ellipsoidal set $\rectset$: $\big \{ \relu{ \data \vec{u} } \, | \firstw \in \mathbb{R}^d, \| \firstw\|_2 \leq 1   \big\}$\centering}\label{fig:spike1_recellipsoid}
	\end{subfigure} \hspace*{\fill}
			\begin{subfigure}[t]{0.32\textwidth}
		\centering
		\includegraphics[width=1\textwidth, height=0.8\textwidth]{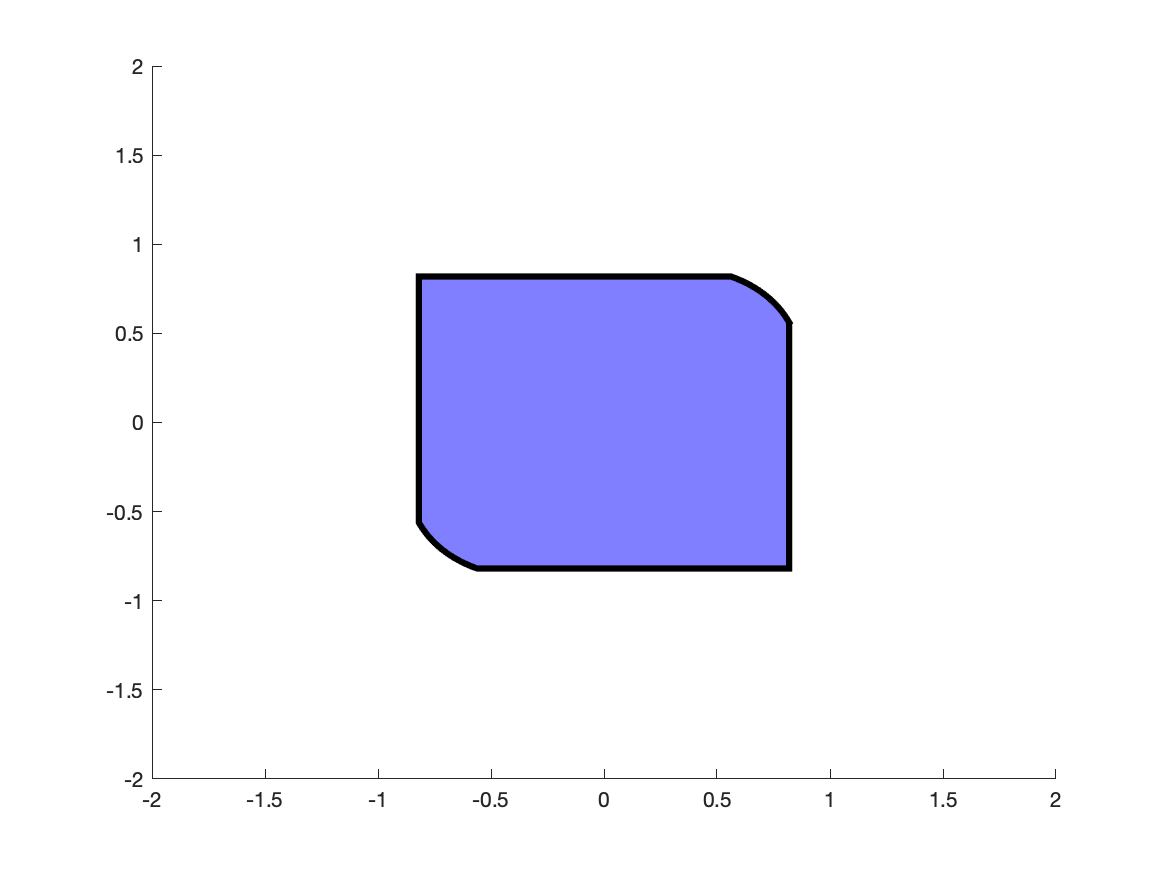}
		\caption{Polar set $\rectset^\circ$: $\{\dual | \dual^T \firstw\le 1\, \forall \firstw\in \rectset\}$\centering} \label{fig:spike1_polar}
	\end{subfigure} \hspace*{\fill}
	\medskip
	\caption{Two dimensional illustration of a the rectified ellipsoid that is not spike-free and its polar set. Note that the polar set is not polyhedral and exhibits a combination of smooth and non-smooth faces.}
\label{fig:spike1}
	\end{figure*}

\begin{figure*}[t]
	\centering
	\captionsetup[subfigure]{oneside}
		\begin{subfigure}[t]{0.32\textwidth}
		\centering
		\includegraphics[width=1\textwidth, height=0.8\textwidth]{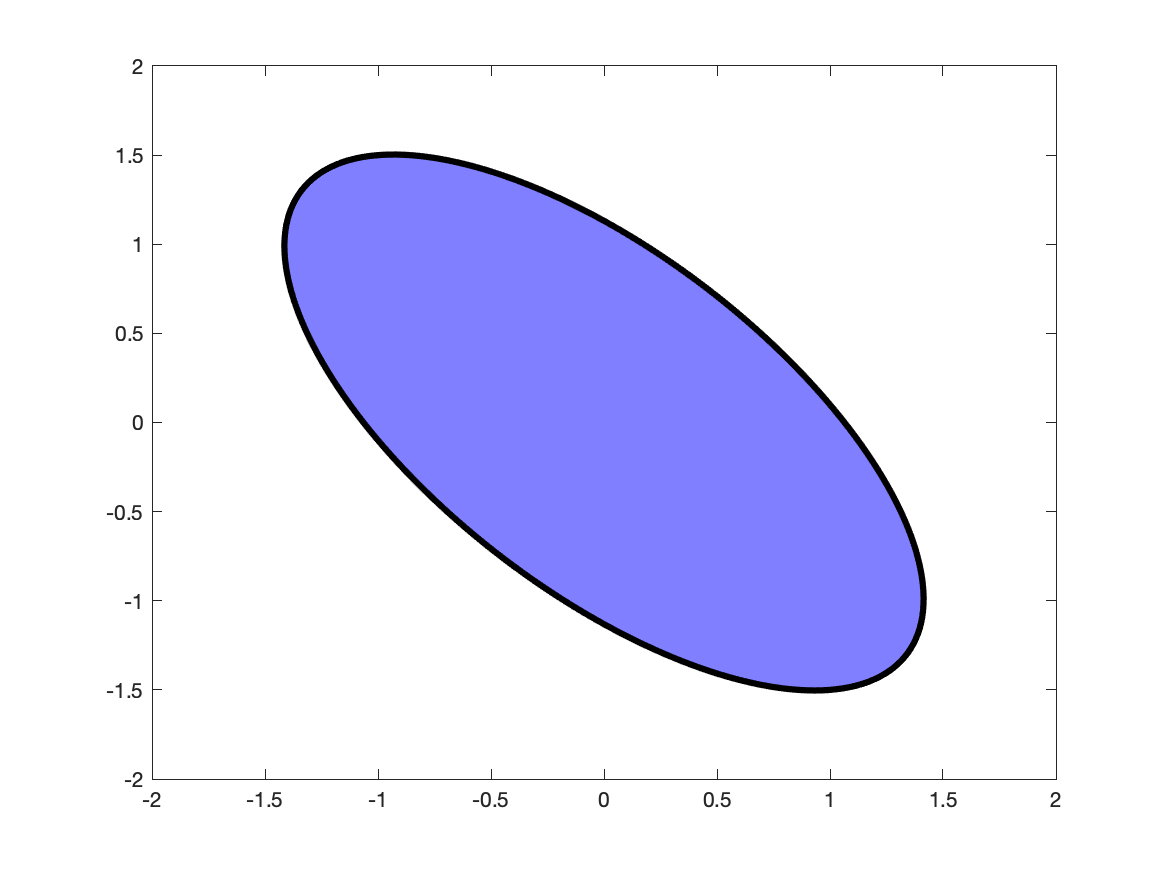}
		\caption{Ellipsoidal set: $\{\data\vec{u}\,\vert\,\firstw \in \mathbb{R}^d, \| \firstw\|_2 \leq 1\}$\centering} \label{fig:spike2_ellipsoid}
	\end{subfigure} \hspace*{\fill}
		\begin{subfigure}[t]{0.32\textwidth}
		\centering
		\includegraphics[width=1\textwidth, height=0.8\textwidth]{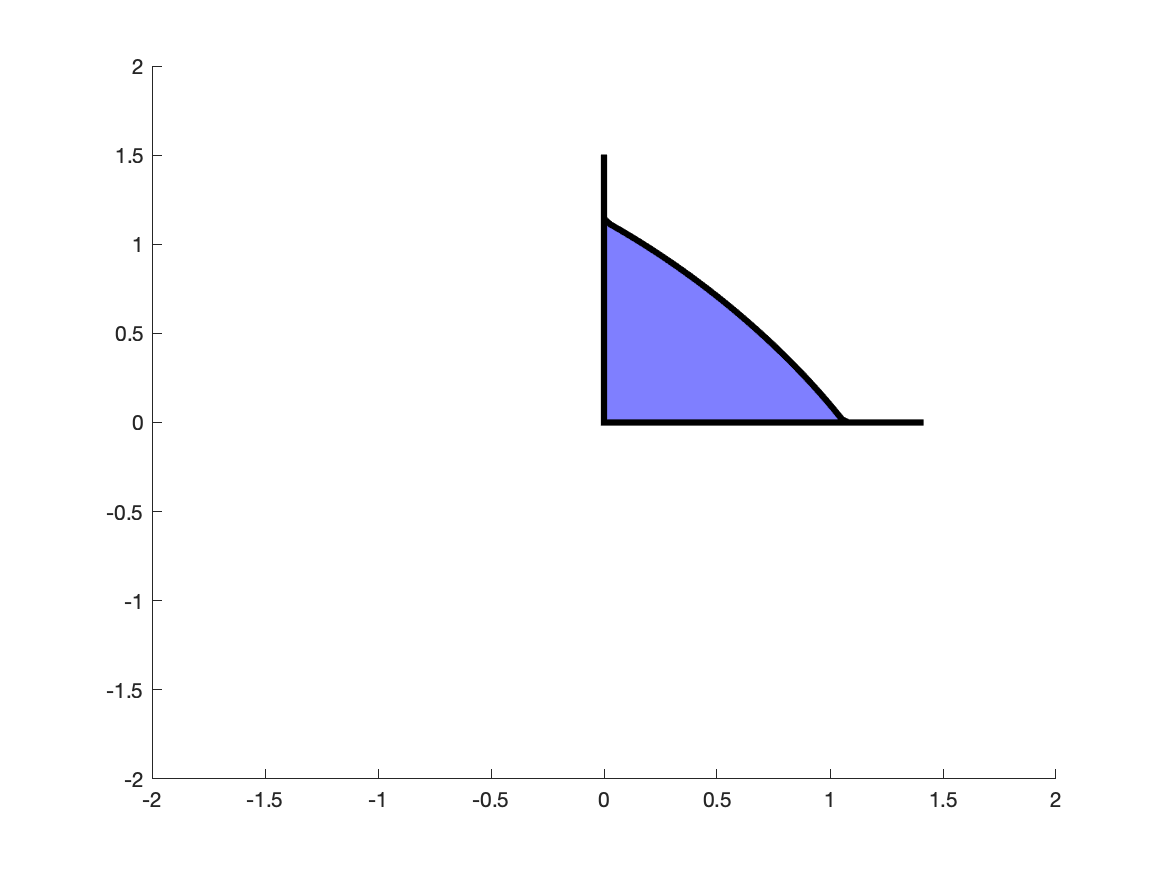}
		\caption{Rectified ellipsoidal set $\rectset$: $\big \{ \relu{ \data \vec{u} } \, | \firstw \in \mathbb{R}^d, \| \firstw\|_2 \leq 1   \big\}$\centering} \label{fig:spike2_recellipsoid}
	\end{subfigure} \hspace*{\fill}
			\begin{subfigure}[t]{0.32\textwidth}
		\centering
		\includegraphics[width=1\textwidth, height=0.8\textwidth]{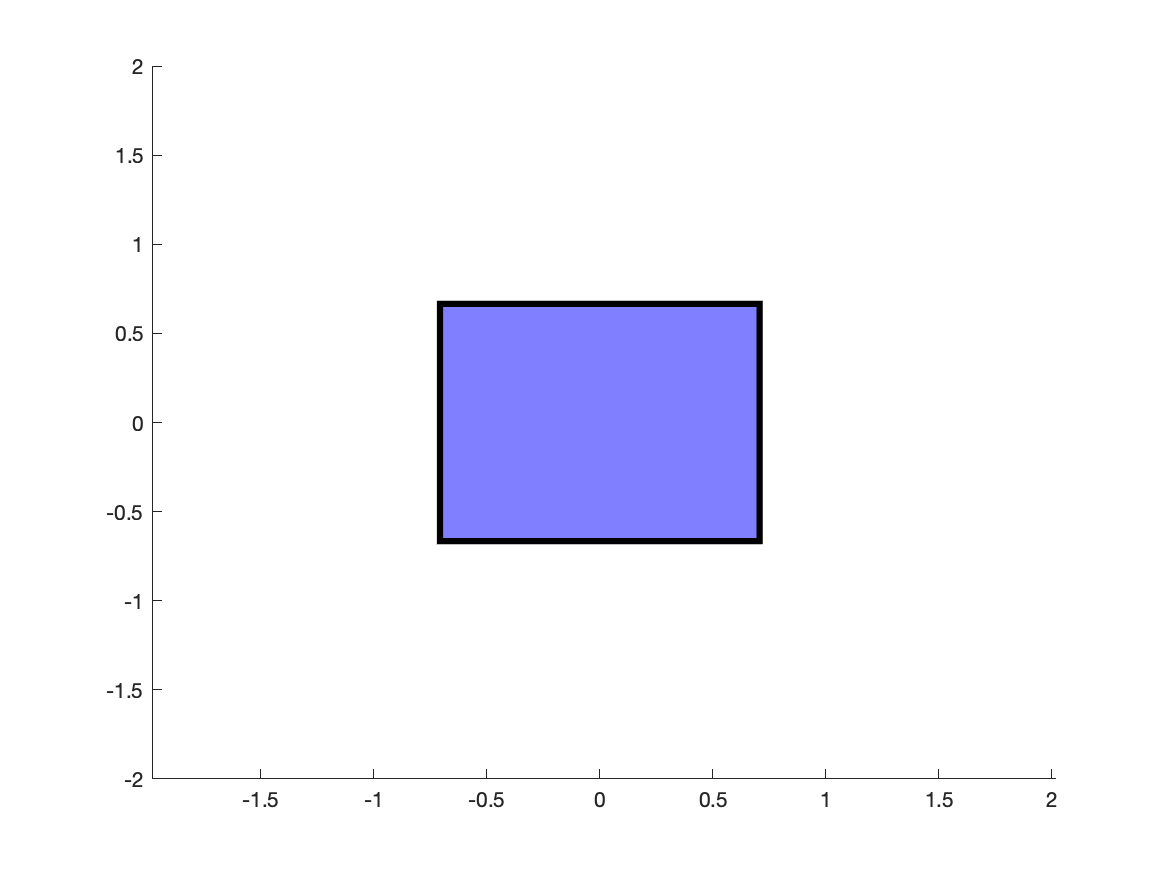}
		\caption{Polar set $\rectset^\circ$: $\{\dual | \dual^T \firstw\le 1\, \forall \firstw\in \rectset\}$\centering} \label{fig:spike2_polar}
	\end{subfigure} \hspace*{\fill}
	\medskip
	\caption{Two dimensional illustration of a the rectified ellipsoid that is not spike-free and its polar set. Note that the polar set is polyhedral since the convex hull of the rectified ellipsoid is polyhedral.}
\label{fig:spike2}
	\end{figure*}

\begin{figure*}[t]
	\centering
	\captionsetup[subfigure]{oneside}
		\begin{subfigure}[t]{0.32\textwidth}
		\centering
		\includegraphics[width=1\textwidth, height=0.8\textwidth]{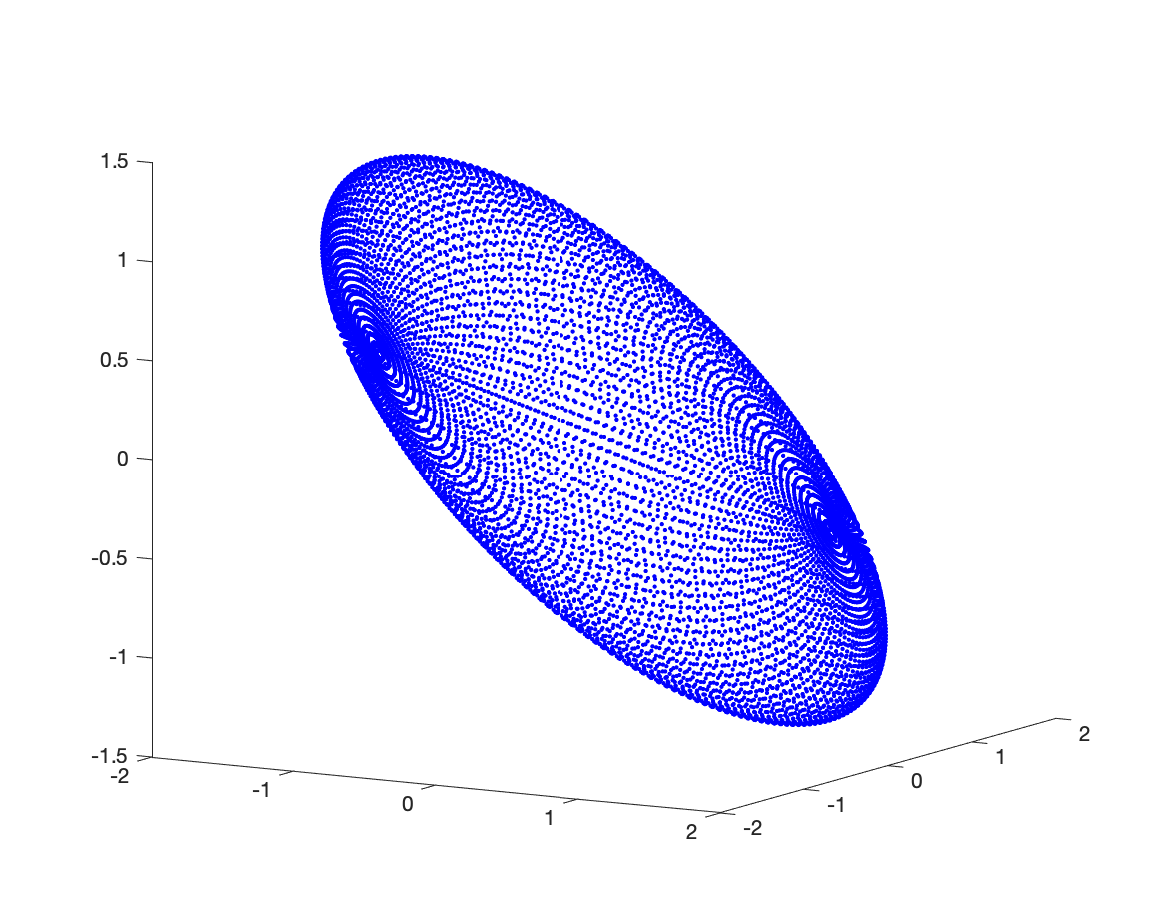}
		\caption{Ellipsoidal set: $\{\data\vec{u}\,\vert\,\firstw \in \mathbb{R}^d, \| \firstw\|_2 \leq 1\}$\centering} \label{fig:3dspike_ellipsoid}
	\end{subfigure} \hspace*{\fill}
		\begin{subfigure}[t]{0.32\textwidth}
		\centering
		\includegraphics[width=1\textwidth, height=0.8\textwidth]{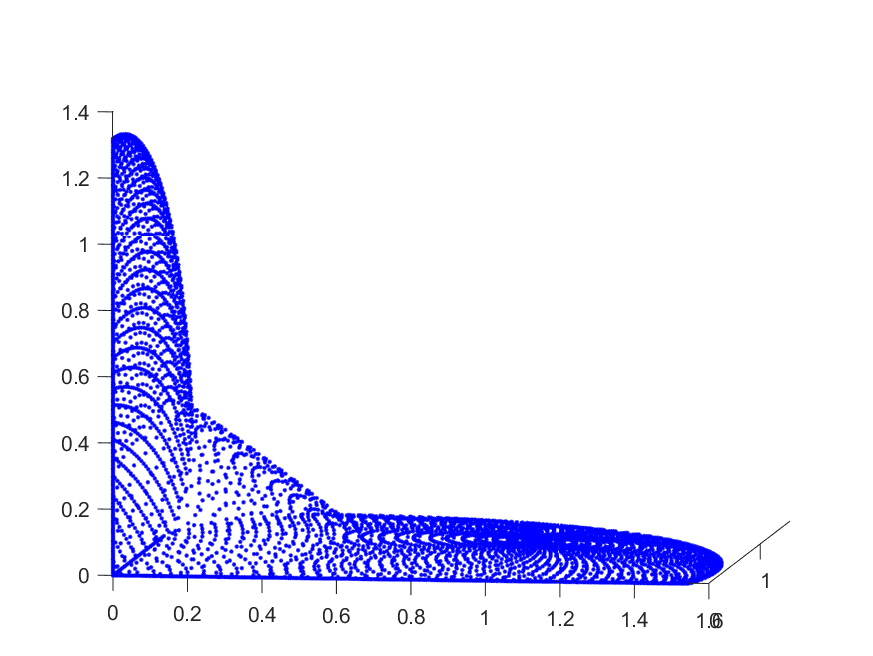}
		\caption{Rectified ellipsoidal set $\rectset$: $\big \{ \relu{ \data \vec{u} } \, | \firstw \in \mathbb{R}^d, \| \firstw\|_2 \leq 1   \big\}$\centering} \label{fig:3dspike_recellipsoid}
	\end{subfigure} \hspace*{\fill}
			\begin{subfigure}[t]{0.32\textwidth}
		\centering
		\includegraphics[width=1\textwidth, height=0.8\textwidth]{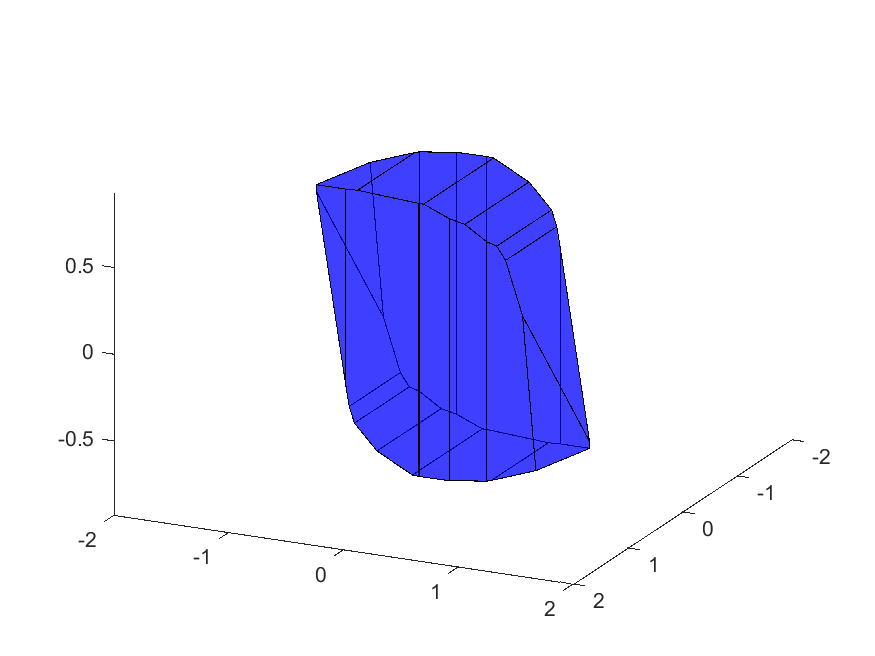}
		\caption{Polar set $\rectset^\circ$: $\{\dual | \dual^T \firstw\le 1\, \forall \firstw\in \rectset\}$\centering} \label{fig:3dspike_polar}
	\end{subfigure} \hspace*{\fill}
	\medskip
	\caption{Three dimensional illustration of the rectified ellipsoid and its polar set. Note that the rectified ellipsoid is the union of lower-dimensional ellipsoids its polar set exhibits a combination of smooth and non-smooth faces.}
\label{fig:3dspikenew}
	\end{figure*}

If $\rectset= \{(\data \firstw )_+ | \firstw \in \mathbb{R}^d, \| \firstw\|_2 \leq 1 \}$ can be expressed as $\mathbb{R}_{+}^n \cap  \{\data \firstw  | \firstw \in \mathbb{R}^d, \| \firstw\|_2 \leq 1 \}$ (see Figure \ref{fig:nospike}), then \eqref{eq:problem_def_geo} can be solved via convex optimization after the rescaling $\firstw = \firstw_1 \secondw_1 $
\begin{align*}
      &\min_{\firstw } \big \|  \data \firstw -\vec{y} \big\|_2^2 \text{ s.t. } \firstw \in \{ \data \firstw \succcurlyeq \vec{0} \} \cup  \{ -\data \firstw \succcurlyeq  \vec{0} \}.
\end{align*}
The following lemma provides a characterization of spike-free matrices 
\begin{lems}\label{lemma:spike-free} 
A matrix $\data$ is spike-free if and only if the following condition holds
\begin{align}
    \forall \vec{u} \in \ball_2, \; \exists\vec{z}\in\ball_2~\mbox{such that}\mbox{ we have } \relu{\data \vec{u}} = \data \vec{z}\,. \label{eq:spikefreecond1}
\end{align}
Alternatively, a matrix $\data$ is spike free if and only if it holds that
\begin{align*}
    \max_{\vec{u}\, :\,\|\vec{u}\|_2\le 1,\, (\vec{I}_n-\data \data^\dagger) (\data\vec{u})_+=\vec{0}} \| \data^\dagger\relu{\data \vec{u}}\|_2 \le 1\,.
\end{align*}
If $\data$ is full row rank, then the above condition simplifies to
\begin{align}
    \max_{\vec{u}\, :\,\|\vec{u}\|_2\le 1} \| \data^\dagger\relu{\data \vec{u}}\|_2 \le 1\,.
    \label{eqn:nospikemaxcondition}
\end{align}
\end{lems}
We note that the condition in \eqref{eqn:nospikemaxcondition} bears a close resemblance to the irrepresentability conditions in Lasso support recovery (see e.g. \citet{zhao2006model}). It is easy to see that diagonal matrices are spike-free. More generally, any matrix of the form $\data=\vec{\Sigma} \vec{V}^T$, where $\vec{\Sigma}$ is diagonal, and $\vec{V}^T$ is any matrix with orthogonal rows, i.e., $\vec{V}^T\vec{V}=\vec{I}_n$, is spike-free. In other cases, $\rectset$ has a non-convex shape as illustrated in Figure \ref{fig:spike1} and \ref{fig:spike2}. Therefore, the ReLU activation might exhibit significantly complicated and non-convex behavior as the dimensionality of the problem increases. Note that $\data\ball_2 \cap \reals^n_+ \subseteq \rectset$ always holds, and therefore the former set is a convex relaxation of the set $\rectset$. We call this set spike-free relaxation of $\rectset$.


As another example for spike-free data matrices, we consider the Singular Value Decomposition of the data matrix $\data = \vec{U} \boldsymbol{\Sigma} \vec{V}^T$ in compact form. We can apply a whitening transformation on the data matrix by defining $\tilde \data = \data \vec{V} \vec{\Sigma}^{-1}$, which is known as zero-phase whitening in the literature. Note that the empirical covariance of the whitened data is diagonal since we have  $\tilde \data \tilde \data^T = \vec{I}_n$. Below, we show whitened data matrices are in fact spike-free. Furthermore, rank-one data matrices with positive left singular vectors are also spike-free as detailed below.
\begin{lems}\label{lemma:whiten}
Let $\data$ be a whitened data matrix with $n\leq d$ that satisfies $\data \data^T =\vec{I}_n$. Then, it holds that
\begin{align*}
      \max_{\vec{u}\, :\,\|\vec{u}\|_2\le 1} \| \data^\dagger\relu{\data \vec{u}}\|_2 \le 1\,,
\end{align*}
where $\data^\dagger= \data^T(\data \data^T)^{-1}$. As a direct consequence, $\data$ is spike-free.
\end{lems}
\begin{lems}\label{lemma:rankone_spikefree}
Let $\data$ be a data matrix such that $\data= \vec{c}\vec{a}^T$, where $\vec{c} \in \mathbb{R}_{+}^n$ and $\vec{a} \in \mathbb{R}^d$. Then, $\data$ is spike-free.
\end{lems}

Although the examples presented here for spike-free data matrices are whitened, one dimensional or rank-one, we believe that the set of spike-free data is far more generic and common. In addition, whitening is a common technique used by deep learning practitioners, e.g., ZCA whitening in image classification is used to improve the validation accuracy of the system as empirically shown in \cite{kmenas_andrewng}. Furthermore, recent work has shown that whitening improves the performance of the state-of-the-art architectures, e.g., ResNets, on benchmark datasets such as CIFAR-10, CIFAR-100, and ImageNet \cite{huang2018decorrelated}. Therefore, we believe that theoretical results on spike-free matrices are valuable for practitioners.

\subsection{Polar convex duality}
It can be shown that the dual of the problem \eqref{eq:problem_def2} is given by\footnote{We refer the reader to Appendix \ref{sec:polar_duality_appendix} for the proof. }
\begin{align}
    &\max_{\dual} \dual^T \vec{y} \mbox{ s.t. } \dual \in \rectset^\circ~ \mbox{,       }-\dual \in \rectset^\circ, \label{eq:dualpolar}
\end{align}
where $\rectset^\circ$ is the polar set \citep{Rockafellar} of $\rectset$ defined as
$    \rectset^\circ = \{\dual | \dual^T \firstw\le 1,\, \forall \firstw\in \rectset\}\,.
$

\subsection{Extreme points} 

A point is called an extreme point of a convex set $\mathcal{C}$ if it does not lie between any two distinct points of $\mathcal{C}$. More precisely, extreme points of a convex set $\mathcal{C}$ is defined as the set of points $\dual \in\mathcal{C}$ such that if $\dual = \frac{1}{2}\dual_1 +\frac{1}{2} \dual_2$, for some $\dual_1, \dual_2 \in \mathcal{C}$, then $\dual=\dual_1=\dual_2$. Let us also define the support function map
\begin{align}
    \sigma_{\mathcal{Q}_{\bf A}}(\dual) := \argmax_{\vec{z} \in \rectset} \dual^T \vec{z}.
\end{align}
Note that the maximum above is achieved at an extreme point of $\mathcal{Q}_{\bf A}$. For this reason, we refer $\sigma_{\mathcal{Q}_{\bf A}}(\dual)$ as the set of extreme points of $\rectset$ along $\dual$. In addition, $\sigma_{\mathcal{Q}_{\bf A}}(\dual)$ is not a singleton in general, but an exposed set. We also remark that the endpoints of the spikes in Figure \ref{fig:spike1} and \ref{fig:spike2} are the extreme points in the ordinary basis directions $\vec{e}_1$ and $\vec{e}_2$.

In the sequel, we show that the extreme points of $\rectset$ are given by data samples and convex mixtures of data samples in one dimensional and multidimensional cases, respectively. Here, we also provide a generic formulation for the extreme points along an arbitrary direction.

\begin{lems} \label{lemma:extreme_l2loss}
In a one dimensional dataset ($d=1$), for any vector $\dual \in \mathbb{R}^n$, an extreme point of $\rectset$ along $\dual$ is achieved when $\firstwscalar_v= \pm 1$ and $b_v=- \sign(\firstwscalar_v)\samplescalar_i $ for a certain index $i\in[n]$.
\end{lems} 
The above lemma shows that the extreme point of the set $\rectset$ along an arbitrary direction $\dual$ yields a set of hidden neurons and biases that take values of $\pm1$ and $\pm \samplescalar_i$, respectively, for an arbitrary index $i \in [n]$. Therefore, the kink of each ReLU activation at the extreme points corresponds to one of the input data samples, i.e., $\samplescalar_i \firstwscalar+\bias=0$ for an arbitrary index $i \in [n]$. Moreover, in the next section (Theorem \ref{theo:equality_dual} and other results) we prove the optimality of these extreme points. Therefore, combined with Theorem \ref{theo:equality_dual}, the above result proves that the optimal network outputs the linear spline interpolation for the input data.

We now generalize the result to higher dimensions by including the extreme points in the span of the ordinary basis vectors. These will improve our spike-free relaxation as a first order correction. In particular, for the spike-free relaxation, we represent the output of ReLU activation as $\relu{\data\firstw}=\data \firstw$ with the constraint $\data \firstw\succcurlyeq \vec{0}$. However, this representation is restrictive since it doesn't allow any non-negative preactivation $\data \firstw$. In order to further improve this relaxation, we also include extreme points in the ordinary basis directions $\{\vec{e}_i\}_{i=1}^n$. Therefore, instead of restricting preactivations to be all nonnegative, we also allow preactivations that have one positive entry. For instance, the behavior in Figure \ref{fig:spike1} and \ref{fig:spike2} is captured by the convex hull of the union of extreme points along $\vec{e}_1$ and $\vec{e}_2$, and the spike-free relaxation.
\begin{lems} \label{lemma:convex_mixtures}
Extreme point in the span of each ordinary basis direction $\vec{e}_{i}$ is given by
\begin{align}\label{eq:mixture_outputs}
    \firstw_i=\frac{\sample_i -\sum_{\substack{j=1\\ j \neq i}}^n \lambda_j\sample_j}{\bigg\|\sample_i -\sum_{\substack{j=1\\ j \neq i}}^n \lambda_j\sample_j\bigg\|_2} \text{  and  } \bias_i= \min_{j\neq i} (-\sample_j^T \firstw_i),
\end{align}
where $\boldsymbol{\lambda}$ is computed via the following problem
\begin{align*}
    \min_{\boldsymbol{\lambda}}  \bigg\|\sample_i -\sum_{\substack{j=1\\ j \neq i}}^n \lambda_j\sample_j\bigg\|_2  \text{ s.t. } \boldsymbol{\lambda} \succcurlyeq \vec{0},\vec{1}^T \boldsymbol{\lambda}=1.
\end{align*}
\end{lems}
Lemma \ref{lemma:convex_mixtures} shows that extreme points of $\rectset$ are given by a convex mixture approximation of the training samples: the hidden neurons are the residuals of the approximation and the corresponding bias values is the negative inner product between the hidden neurons and a training sample, which places a kink at the training sample. Our next result characterizes extreme points along arbitrary directions for the general case.
\begin{lems}\label{lemma:extreme_general_closedform}
For any $\boldsymbol{\alpha} \in \mathbb{R}^n$, the extreme point along the direction of $\boldsymbol{\alpha}$ can be found by
\begin{align}\label{eq:extreme_closedform}
   & \firstw_{\alpha}= \frac{\sum_{i \in \mathcal{S}} (\alpha_i+\lambda_i) \sample_i - \sum_{j \in \mathcal{S}^c} \nu_j \sample_j }{\left\|\sum_{i \in \mathcal{S}} (\alpha_i+\lambda_i) \sample_i - \sum_{j \in \mathcal{S}^c} \nu_j \sample_j \right\|_2} \text{  and  } \bias_{\alpha}  =\begin{cases}
   \max_{i \in \mathcal{S}}(-\sample_i^T \firstw_{\alpha}),  &\text{if } \sum_{i \in \mathcal{S}} \alpha_i \leq 0\\    
  \min_{j \in \mathcal{S}^c}(-\sample_j^T \firstw_{\alpha}), &\text{otherwise}   
\end{cases}
\end{align}
where $\mathcal{S}$ and $\mathcal{S}^c$ denote the set of active and inactive ReLUs, respectively, and $\boldsymbol{\lambda}$ and $\boldsymbol{\nu}$ are obtained via the following convex problem
\begin{align*}
    &\min_{\boldsymbol{\lambda}, \boldsymbol{\nu}} \max_{\firstw,\bias} \firstw^T \bigg (\sum_{i \in \mathcal{S}} (\alpha_i+\lambda_i)\sample_i- \sum_{j \in \mathcal{S}^c} \nu_j \sample_j  \bigg)  \text{ s.t. } \boldsymbol{\lambda}, \boldsymbol{\nu} \succcurlyeq 0, \sum_{i \in \mathcal{S}} (\alpha_i + \lambda_i)= \sum_{j \in \mathcal{S}^c} \nu_j,  \| \firstw \|_2 \le 1.
\end{align*}
\end{lems}

Lemma \ref{lemma:extreme_general_closedform} proves that optimal neurons can be characterized as a linear combination of the input data samples. Below, we further simplify this characterization and obtain a representer theorem for regularized NNs.
\begin{cors}[\textbf{A representer theorem for optimal neurons}] \label{cor:representer}
Lemma \ref{lemma:extreme_general_closedform} implies that each extreme point along the direction $\boldsymbol{\alpha}$ can be written in the following compact form
\begin{align*}
    \firstw_{\boldsymbol{\alpha}}=\frac{\sum_{i \in \mathcal{S}}\alpha_i (\sample_i-\sample_k)} {\| \sum_{i \in \mathcal{S}}\alpha_i (\sample_i-\sample_k)\|_2}\text{ and } \bias=-\sample_k^T \firstw_{\alpha}, \, \mbox{for some } k \mbox{ and subset } \mathcal{S}.
\end{align*}
Therefore, optimal neurons in the training objectives \eqref{eq:problem_statement} and \eqref{eq:problem_def2} all obey the above  representation.
\end{cors}
\begin{remark}
An interpretation of the extreme points provided above is an auto-encoder of the training data: the optimal neurons are convex mixture approximations of subsets of training samples in terms of other subsets of training samples.
\end{remark}

\section{Main Results}\label{sec:main_results}
In the following, we present our main findings based on the extreme point characterization introduced in the previous section.

\subsection{Convex duality}
In this section, we present our first duality result for the non-convex NN training objective given in \eqref{eq:problem_def2}.
\begin{theos} \label{theo:equality_dual}
The dual of the problem in \eqref{eq:problem_def2} is given by 
\begin{align}
    &D^*=\max_{\vec{v}\in \reals^n} \vec{v}^T \vec{y} \hspace{3cm} = && \max_{\vec{v}\in \reals^n} \vec{v}^T \vec{y}\,,\label{eq:problem_dual}\\ \nonumber
    &\text{s.t. } \big\vert\vec{v}^T \relu{\data \firstw} \big \vert \le 1\, \forall \firstw\in \ball_2  &&\text{s.t. } \vec{v} \in \rectset^\circ,-\vec{v} \in \rectset^\circ 
\end{align}
and we have $P^*\ge D^*$. For finite width NNs, there exists an optimal network width $m^*$ that is upperbounded as $m^*\leq n+1$ such that strong duality holds, i.e., $P^*=D^*$, and an optimal $\firstwmat$ for \eqref{eq:problem_def2} satisfies
$    \|(\data \firstwmat^*)_{+}^T \dual^* \|_{\infty} = 1\,,
$
where $\dual^*$ is dual optimal.
\end{theos}
\begin{remark}
Over-parameterization has been extensively studied in the literature and has shown to be one of the key factors for the remarkable generalization performance of deep neural networks \cite{du_overparameterized,li_overparameterized,du2018overparameterized,arora2018overparameterization,brutzkus_overparameterized_linear,neyshabur2018overparameterization,ntk_jacot}. However, most of these studies require an extreme over-parameterization level, e.g., (Du et al, 2018) \cite{du_overparameterized} requires $m=\mathcal{O}(n^6)$, or even an infinite-width as in \cite{ntk_jacot}, which is far from empirical observations and therefore fail to explain the success of neural networks in practice. However, the result in Theorem \ref{theo:equality_dual} require $m^*$ neurons, where $m^* \leq n+1$. Even the upper-bound $n+1$ on $m^*$ is significantly less over-parameterized than the previous theoretical studies. Therefore, we claim that our analysis requires a moderate amount of overparameterization and aligns with practical settings.
\end{remark}
\begin{remark}
Note that \eqref{eq:problem_dual} is a convex optimization problem with infinitely many constraints, and in general not polynomial-time tractable. In fact, even checking whether a point $\dual$ is feasible is NP-hard: we need to solve $\max_{\firstw:\|\firstw\|_2\le 1} \sum_{i=1}^n v_i \relu{\sample_i^T\firstw}$. This is related to the problem of learning halfspaces with noise, which is NP-hard to approximate within a constant factor (see e.g. \citet{guruswami2009hardness,bach2017breaking}).
\end{remark}
Based on the dual form and the optimality condition in Theorem \ref{theo:equality_dual}, we can characterize the optimal neurons as the extreme points of a certain set.
\begin{cors} \label{cor:extreme_points_optimality}
Theorem \ref{theo:equality_dual} implies that the optimal neuron weights are extreme points which solve the following optimization problem
\begin{align*}
    \argmax_{\firstw \in \ball_2} 
    \left \vert {\dual^*}^T \relu{\data\firstw}\, \right \vert.
\end{align*}
\end{cors}
The above corollary shows that the optimal neuron weights are extreme points along $\pm \dual^*$ given by $\sigma_{\mathcal{Q}_{\bf A}}(\pm \dual^*)$ for some dual optimal parameter $\dual^*$.

In the sequel, we first provide a theoretical analysis for the duality gap of finite width NNs and then prove strong duality under certain technical conditions.
\subsubsection{Duality for finite width neural networks}
The following theorem proves that weak duality holds for any finite width NN.
\begin{theos}\label{theo:weak_duality}
Suppose that the optimization problem \eqref{eq:problem_def2} is feasible, i.e., there exists a set $\theta$ such that $\f_{\theta}(\data)=\vec{y}$, then weak duality holds for \eqref{eq:problem_def2}.
\end{theos}
We now prove that strong duality holds for any feasible finite width NN, in which the number of neurons exceeds a critical number less than $n+1$.
\begin{theos}\label{theo:strong_duality}
Let $\{ \data,\vec{y}\}$ be a dataset such that the optimization problem \eqref{eq:problem_def2} is feasible and the width exceeds a critical threshold, i.e., $m \geq m^*$, where $m^*$ is the number of constraints active in the dual problem \eqref{eq:problem_dual} that obeys $m^* \leq n+1$. Then, strong duality holds for \eqref{eq:problem_def2}.
\end{theos}
Since strong duality holds for finite width NNs as proved in Theorem \ref{theo:strong_duality}, we can achieve the minimum of the primal problem in \eqref{eq:problem_def2} through the dual form in \eqref{eq:problem_dual}. Therefore, the NN architecture in \eqref{eq:2layer_function} can be globally optimized via a subset of extreme points defined in Corollary \ref{cor:extreme_points_optimality}.

In the sequel, we first show that we can explicitly characterize the set of extreme points for some specific practically relevant problems. We then prove that strong duality holds for these problems.

\subsection{Structure of one dimensional networks}
We are now ready to present our results on the structure induced by the extreme points for one dimensional problems. The following corollary directly follows from Lemma \ref{lemma:extreme_l2loss}.
\begin{cors}\label{cor:1d_optimality}
Let $\{\samplescalar_i\}_{i=1}^n$ be a one dimensional  training set i.e., $\samplescalar_i \in \reals,~ \forall i\in[n]$. Then, a set of solutions  to \eqref{eq:problem_def2} that achieve the optimal value are extreme points, and therefore satisfy $\{(\firstwscalar_i, \bias_i) \}_{i=1}^{m}$, where  $\firstwscalar_i=\pm 1, \bias_{i}=- \sign(\firstwscalar_i) \samplescalar_{i}$. 
\end{cors}

\begin{figure*}[t]
	\centering
	
			\captionsetup[subfigure]{oneside}
		\begin{subfigure}[t]{0.32\textwidth}
		\centering
		\includegraphics[width=1.1\textwidth, height=.9\textwidth]{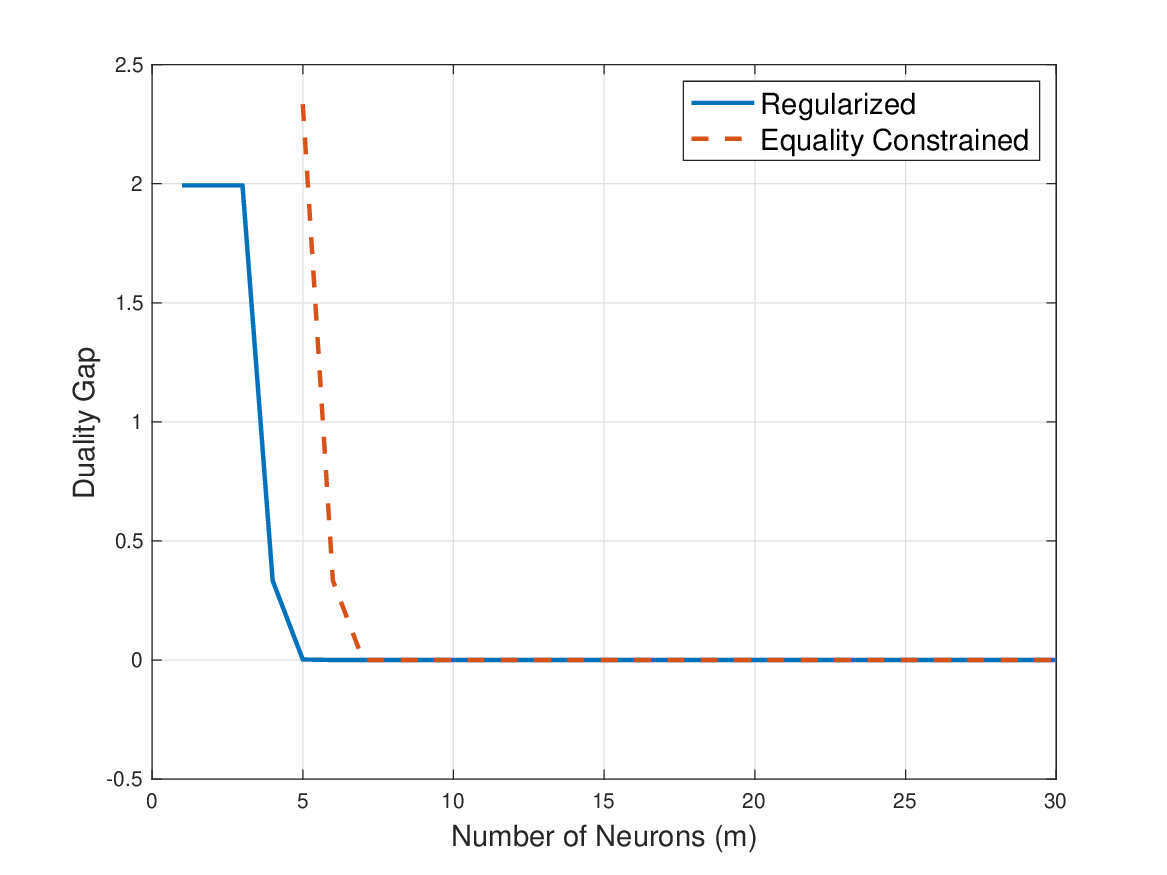}
		\caption{Duality gap for the one dimensional dataset in Figure \ref{fig:neuron}.} \label{fig:dual_gap}
	\end{subfigure} \hspace*{\fill}
			\begin{subfigure}[t]{0.32\textwidth}
		\centering
		\includegraphics[width=1.1\textwidth, height=.9\textwidth]{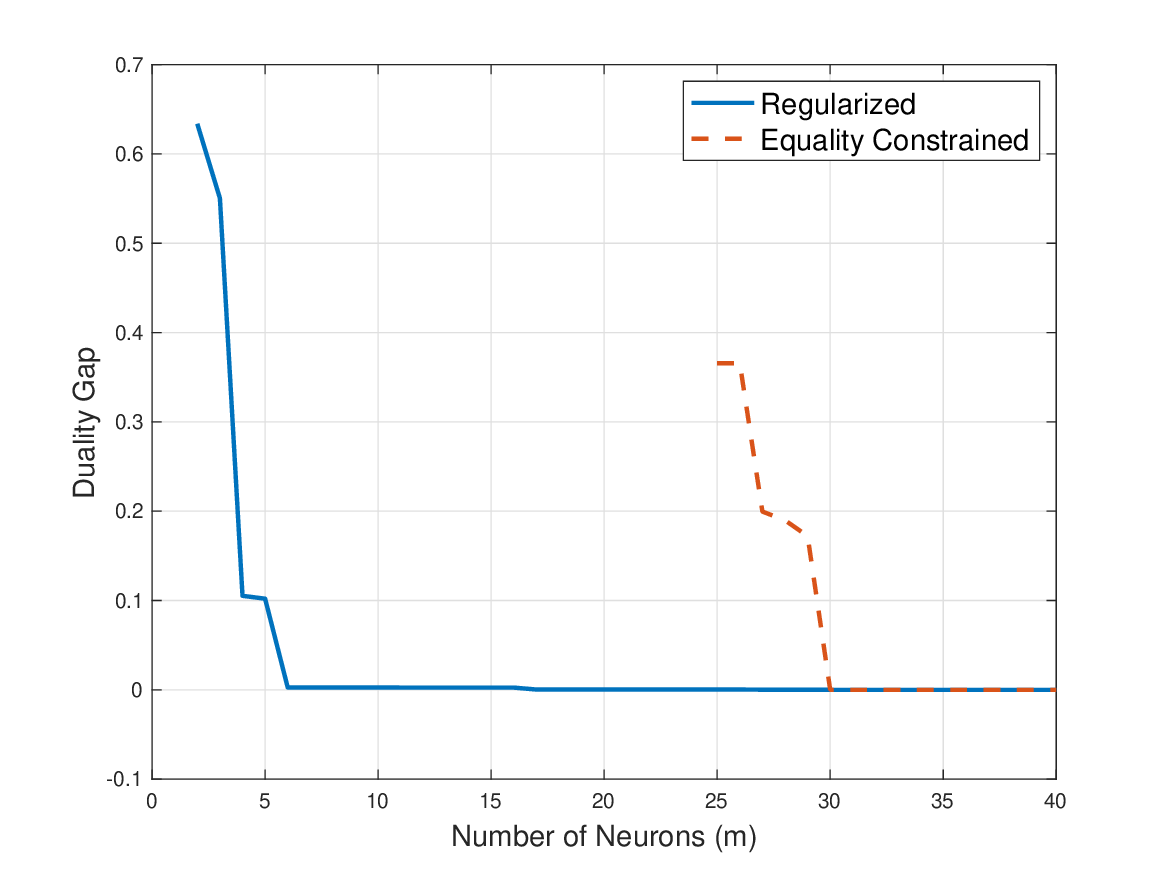}
		\caption{Duality gap for a rank-one dataset with $n=15$ and $d=10$.} \label{fig:dual_gap_rankone}
	\end{subfigure} \hspace*{\fill}
			\begin{subfigure}[t]{0.32\textwidth}
		\centering
		\includegraphics[width=1.1\textwidth, height=.9\textwidth]{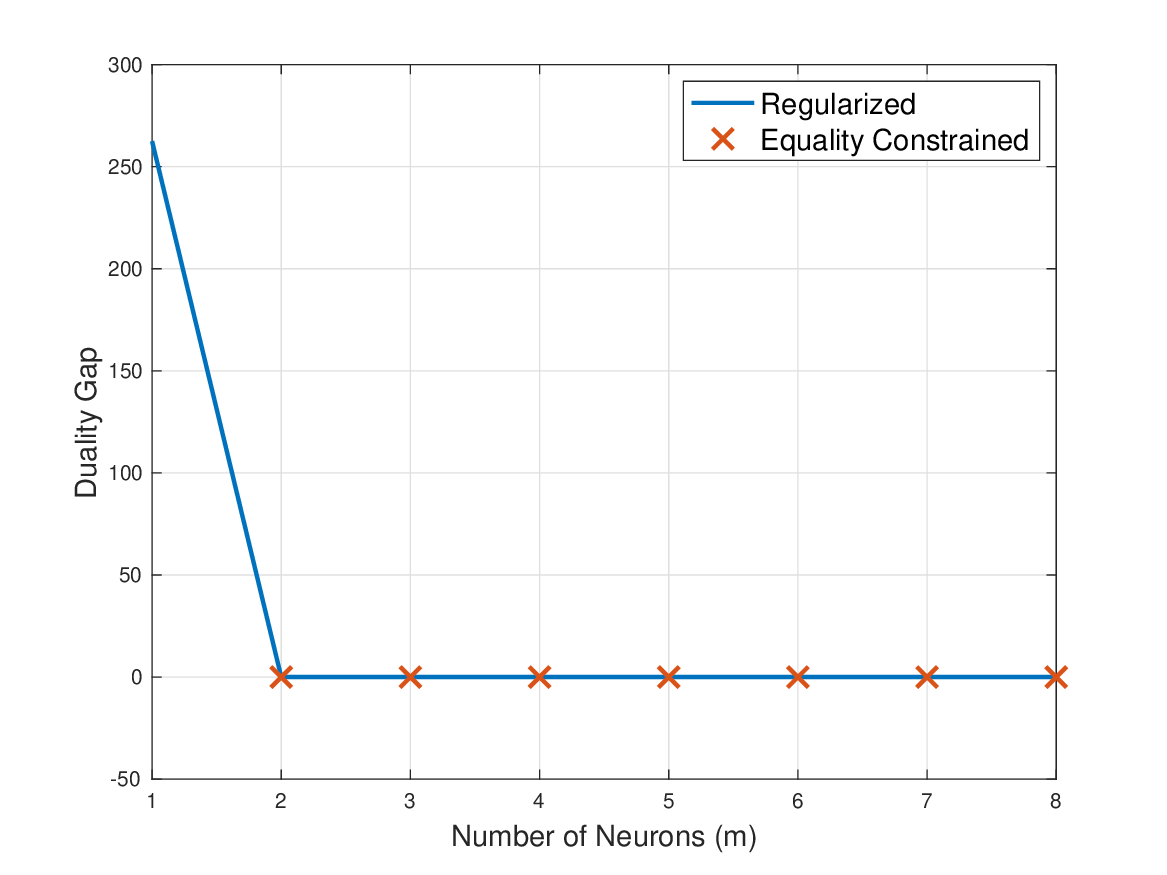}
		\caption{Duality gap for a whitened dataset with $n=30$ and $d=40$.} \label{fig:dual_gap_multi}
	\end{subfigure} \hspace*{\fill}
	\medskip
		\caption{Duality gap for a regression scenario, where we select $\beta=10^{-3}$ for the regularized problem. Here, we consider both the equality constrained case in \eqref{eq:problem_def2} and the regularized case in \eqref{eq:problem_def_regularized}. To solve the primal and dual problems, we use CVX \cite{cvx} with the SDPT3 solver \cite{tutuncu2001sdpt3}.} \label{fig:dual}
	\end{figure*}
	
	\begin{cors} \label{cor:1d_strong_duality}
	For problems involving one dimensional training data, strong duality holds as a result of Corollary \ref{cor:1d_optimality} and Theorem \ref{theo:strong_duality} when $m\ge n+1$.
	\end{cors}
	Corollary \ref{cor:1d_strong_duality} implies that we can globally optimize \eqref{eq:2layer_function} using a subset of finite number of solutions in Corollary \ref{cor:1d_optimality}. In Figure \ref{fig:dual_gap}, we perform a numerical experiment on the dataset plotted in Figure \ref{fig:neuron}. Since strong duality holds for finite width networks with at most $n+1$ neurons as proven in Theorem \ref{theo:equality_dual}, in Figure \ref{fig:dual_gap}, the duality gap vanishes when we reach a certain $m$ value using the parameters formulated in Corollary \ref{cor:1d_optimality}. Notice that this result also validates Theorem \ref{theo:strong_duality}.
	
\begin{figure*}[t]
	\centering
	\captionsetup[subfigure]{oneside,margin={1cm,0cm}}
		\centering
		\includegraphics[width=.7\textwidth, height=0.5\textwidth]{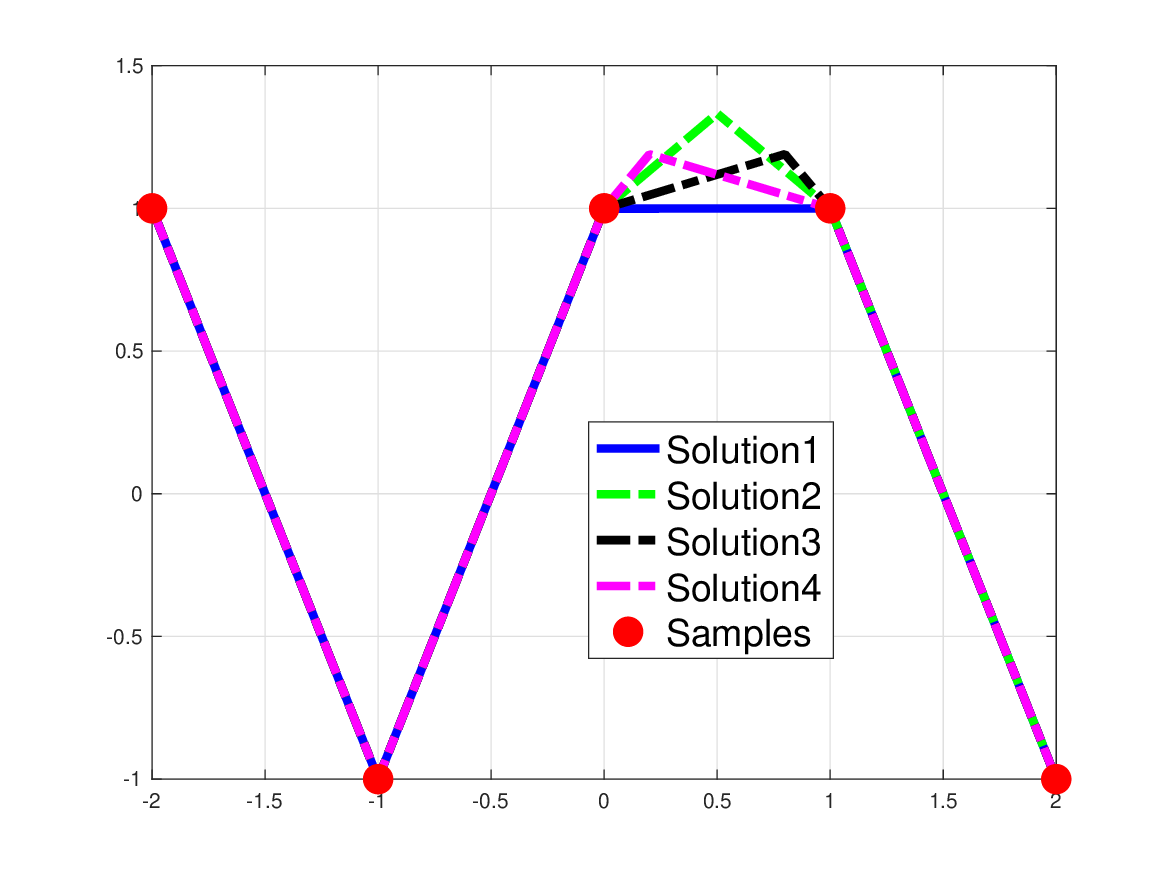}
		\caption{An example verifying our non-uniqueness characterization in Proposition \ref{prop:1d_uniqueness}. Here, we plot four different solutions that analytically achieve the optimal $\ell_2^2$ regularized training cost but produce different predictions. It is interesting to note that the minimum norm criteria is not sufficient to uniquely determine the NN model. The details (including the exact analytical expressions) are presented in Section \ref{sec:proofs}.} 
	\medskip
\label{fig:uniquness}
	\end{figure*}

	Besides, Corollary \ref{cor:1d_optimality} proves that the optimal function output is the linear spline interpolation for the input data, where the kinks of ReLU activations occur at one of the data points. However, in the following, we prove that this set of solutions is not unique so that there might exist other optimal solutions to \eqref{eq:problem_def2} with different function outputs.
	\begin{props}\label{prop:1d_uniqueness}
	The solution provided in Corollary \ref{cor:1d_optimality} is not unique in general. Let us denote the set of active samples for an arbitrary neuron with the parameters $(\firstw,\bias)$ as $\mathcal{S}=\{i| \samplescalar_i \firstwscalar+\bias \geq 0 \}$ and its complementary $\mathcal{S}^c=\{ j|\samplescalar_j \firstwscalar+\bias < 0\}= [n]/\mathcal{S}$. Then, whenever $\sum_{i \in \mathcal{S}} v_i=0$, the value of the bias does not change the objective in the dual constraint. Thus, all the bias values in the range $[\max_{i \in \mathcal{S}}(-\samplescalar_i), \text{ }\min_{j \in \mathcal{S}^c}(-\samplescalar_j)]$ are optimal. In such cases, there are multiple optimal solutions for the training problem. Remarkably, our duality framework enables the construction of all optimal solutions. In Appendix \ref{sec:proofs}, we present an analytic expressions for a counter-example where an optimal solution is not in this form, i.e., not a piecewise linear spline, and illustrate the corresponding function fits in Figure \ref{fig:uniquness}.
	\end{props}
    We remark that \cite{infinite_width} also proved the optimality of the linear spline interpolation as the network output function for one dimensional data and scalar output networks. Moreover, they empirically observed the non-uniqueness of the optimal solutions however did not give any theoretical justifications for this observation. On the contrary, in Proposition \ref{prop:1d_uniqueness}, we completely characterize the set of optimal solutions via convex duality and prove that the linear spline interpolation is not the only optimal solution. In other words, there might exist other optimal solutions with different functional forms. Based on other characterization for the set of optimal solutions, we also provide an empirical evidence for non-uniqueness in Figure 7. Here, we particularly depict three different optimal function fits that are very similar to the linear spline interpolation with the kinks at the input data points except a kink in the range $(0,1)$. Furthermore, since we utilize a more generic approach based on convex duality, our analysis is valid for rank-one data and vector output networks as detailed in the next section and Section \ref{sec:vectoroutput}, respectively.
	
	\subsection{Solutions to rank-one problems}
In this section, we first characterize all possible extreme points for problems involving rank-one data matrices. We then prove that strong duality holds for these problems.
\begin{cors}\label{cor:rankone_extreme}
Let $\data$ be a data matrix such that $\data=\vec{c}\sample^T$, where $\vec{c} \in \mathbb{R}^n$ and $\sample \in \mathbb{R}^d$. Then, a set of solutions  to \eqref{eq:problem_def2} that achieve the optimal value are extreme points, and therefore satisfy $\{(\firstw_i, \bias_i) \}_{i=1}^{m}$, where  $\firstw_i=s_i\frac{\vec{a}}{\|\vec{a}\|_2}, \bias_{i}=- s_i c_{i}\|\vec{a}\|_2$ with $s_i=\pm 1, \forall i \in [m]$. 
\end{cors}
\begin{cors}\label{cor:rankone_strong_duality}
As a result of Theorem \ref{theo:strong_duality} and Corollary \ref{cor:rankone_extreme}, strong duality holds for problems involving rank-one data matrices.
\end{cors}
Corollary \ref{cor:rankone_extreme} and \ref{cor:rankone_strong_duality} indicate that we can globally optimize regularized NNs using a subset of the extreme points in Corollary \ref{cor:rankone_extreme}. We also present a numerical example in Figure \ref{fig:dual_gap_rankone} to confirm the theoretical prediction of Corollary \ref{cor:rankone_strong_duality}.

\subsection{Solutions to spike-free problems}
Here, we show that as a direct consequence of our analysis above, problems involving spike-free data matrices can be equivalently stated as a convex optimization problem. The next result formally presents the convex equivalent problem and further proves that this problem can be globally optimized in polynomial-time with respect to all the problem parameters $n$, $d$, and $m$.

\begin{theos}\label{theo:polynomialtime_convex}\footnote{Proof and extensions are presented in Section \ref{sec:polynomialtime_convex}.}
Let $\data$ be a spike-free data matrix. Then, the non-convex training problem \eqref{eq:problem_def2} can be equivalently formulated as the following convex optimization problem
\begin{align*}
    \min_{\vec{w}_1,\vec{w}_2}  \|\vec{w}_1\|_2+\|\vec{w}_2\|_2 \text{ s.t. } \data(\vec{w}_1-\vec{w}_2)=\vec{y},\;\data \vec{w}_1 \succcurlyeq \vec{0}, \; \data \vec{w}_2 \succcurlyeq \vec{0},
\end{align*}
which can be globally optimized by a standard convex optimization solver in $\mathcal{O}(d^3)$.
\end{theos}
Theorem \ref{theo:polynomialtime_convex} proves that the regularized training problem for spike-free data matrix can be equivalently cast as a convex optimization problem with two neurons. More importantly, this convex problem can be globally optimized by standard convex optimization solvers, e.g., interior-point methods, with a polynomial-time complexity in terms of the number of data samples $n$ and the feature dimension $d$.

\subsection{Closed-form solutions and $\ell_0$-$\ell_1$ equivalence}
A considerable amount of literature have been published on the equivalence of minimal $\ell_1$ and $\ell_0$ solutions in under-determined linear systems, where it was shown that the equivalence holds under assumptions on the data matrices (see e.g. \citet{CandesTao05,Donoho04a,fung2011equivalence}). We now prove a similar equivalence for two-layer NNs. Consider the minimal cardinality problem
\begin{align}\label{eq:problem_Lzero}
    \min_{\theta \in \Theta} \| \secondwvec \|_0 \text{ s.t. } f_{\theta}(\data)=\vec{y}, \| \firstw_j \|_2=1 , \forall j .
\end{align}
The following results provide a characterization of the optimal solutions to the above problem.
\begin{lems}\label{lemma:closedform_2neuron}
Suppose that $n\le d$, $\data$ is full row rank and $\vec{y}$ contains both positive and negative entries, and define $\data ^\dagger = \data^T (\data \data^T)^{-1}$. Then an optimal solution to the problem in \eqref{eq:problem_Lzero} is given by
\begin{align*}
    \firstw_1^* &= \frac{\data^\dagger \relu{\vec{y}}}{\|\data^\dagger \relu{\vec{y}}\|_2}\mbox{ ,  } \secondw_1^* = \|\data^\dagger \relu{\vec{y}}\|_2 \quad \text{ and } \quad
    \firstw_2^* = \frac{\data^\dagger \relu{-\vec{y}}}{\|\data^\dagger \relu{-\vec{y}}\|_2} \mbox{,  } \secondw_2^* = -\|\data^\dagger \relu{-\vec{y}}\|_2\,.
\end{align*}
\end{lems}

\begin{lems} \label{lemma:l1-l0 equivalence}
We have $\ell_1$-$\ell_0$ equivalence, i.e., the optimal solutions of \eqref{eq:problem_Lzero} and \eqref{eq:problem_def2} coincide if the following condition holds
\begin{align*}
    \min_{\vec{v}:\vec{v}^T \relu{\data\firstw_1}=1,\vec{v}^T \relu{\data\firstw_2}=-1}~\max_{\firstw:\|\firstw\|_2\le 1} \big\vert\vec{v}^T\relu{\data\firstw}\big\vert \le 1\,.
\end{align*}
Furthermore, whitened data matrices with $n\le d$ satisfy $\ell_1$-$\ell_0$ equivalence.
\end{lems}

\begin{cors}\label{cor:whitened_strong_duality}
As a result of Theorem \ref{theo:strong_duality} and Corollary \ref{lemma:l1-l0 equivalence}, strong duality holds for problems involving whitened data matrices.
\end{cors}

Lemma \ref{lemma:l1-l0 equivalence} and Corollary \ref{cor:whitened_strong_duality} prove that \eqref{eq:2layer_function} can be globally optimized using the two extreme points in Lemma \ref{lemma:closedform_2neuron}. Therefore, we achieve an equivalence between $\ell_1$-$\ell_0$ problems. Moreover, this result is also consistent with Theorem \ref{theo:polynomialtime_convex} as its special case, i.e., we have full row-rank spike-free data matrix. We also validate Corollary \ref{cor:whitened_strong_duality} via a numerical experiment in Figure \ref{fig:dual_gap_multi}.

\subsection{A cutting plane method} \label{sec:relaxations}

In this section, we introduce a cutting plane based training algorithm for the NN in \eqref{eq:2layer_function}. 
Among infinitely many possible unit norm weights, we need to find the weights that violate the inequality constraint in \eqref{eq:problem_dual}, which can be done by solving the following optimization problems
\begin{align}
\begin{split}
    &\firstw_1^*=\argmax_{\firstw:\| \firstw \|_2 \leq 1} \dual^T (\data \firstw)_+  \hspace{1cm}  \firstw_2^* =\argmin_{\firstw:\| \firstw \|_2 \leq 1} \dual^T (\data \firstw)_+ .
    \end{split}
    \label{eq:vio_nonconvex}
\end{align}
 However, \eqref{eq:vio_nonconvex} is not a convex problem since ReLU is a convex function. There exist several methods and relaxations to find the optimal parameters for \eqref{eq:vio_nonconvex}. As an example, one can use the Frank-Wolfe algorithm \citep{frank_wolfe} in order to approximate the solution iteratively. In the following, we show how to relax the problem using our spike-free relaxation 
\begin{align}
\begin{split}
    &\hat{\firstw}_1=\argmax_{\firstw:\data \firstw  \succcurlyeq \vec{0}, \| \firstw \|_2 \leq 1} \dual^T \data \firstw  \hspace{1cm} \hat{\firstw}_2=\argmin_{\firstw:\data \firstw  \succcurlyeq \vec{0}, \| \firstw \|_2 \leq 1} \dual^T \data \firstw ,
    \end{split} \label{eq:vio_convex}
\end{align}
where we relax the set $\{(\data \firstw)_+| \firstw \in \mathbb{R}^d , \| \firstw\|_2 \leq 1\} $ as $\{\data \firstw  | \firstw \in \mathbb{R}^d, \| \firstw\|_2 \leq 1 \} \cap \mathbb{R}_+^n$. Now, we can find the weights for the hidden layer using \eqref{eq:vio_convex}. In the cutting plane method, we first find a violating neuron using \eqref{eq:vio_convex}. After adding these neurons to $\firstwmat$ as columns, we  solve \eqref{eq:problem_def2}. If we cannot find a new violating neuron then we terminate the algorithm. Otherwise, we find the dual parameter for the updated $\firstwmat$ and then repeat this procedure till we find an optimal solution. We also provide the full algorithm in Algorithm \ref{alg:pseudo1_nobias}\footnote{We also provide the cutting plane method for NNs with a bias term in Appendix \ref{sec:cuttingplane_bias}.}. 

The major advantage of our cutting-plane algorithm (compared to the non-convex methods such as Frank-Wolfe in \cite{bach2017breaking}) is that it can be solved via convex optimization. More specifically, in Algorithm \ref{alg:pseudo1_nobias}, all the problems we need to solve are convex and therefore can be globally and efficiently optimized by standard convex solvers without requiring any exhaustive search to tune hyperparameters, e.g., learning rate and initialization, or heuristics such as dropout. Moreover, since the algorithm incrementally inserts neurons to the hidden layer, there is no need to carefully tune the hidden layer width or use unnecessarily wide architectures to fit the training data. We next prove the convergence of Algorithm \ref{alg:pseudo1_nobias} for spike-free data.

\begin{algorithm}[!h]
		\caption{Cutting Plane based Training Algorithm for Two-Layer NNs (without bias)}
		\begin{algorithmic}[1]
        \State Initialize $\vec{v}=\vec{y}$  
        \While{there exists a violating neuron}
        \State Find $\hat{\vec{u}}_1$ and $\hat{\vec{u}}_2$:
        \begin{align*}
        \begin{split}
            &\hat{\firstw}_1=\argmax_{\firstw:\data \firstw  \succcurlyeq \vec{0}, \| \firstw \|_2 \leq 1} \dual^T \data \firstw  \hspace{1cm} \hat{\firstw}_2=\argmin_{\firstw:\data \firstw  \succcurlyeq \vec{0}, \| \firstw \|_2 \leq 1} \dual^T \data \firstw ,
            \end{split} 
        \end{align*}
        \State $\firstwmat^* \leftarrow [\firstwmat^* \text{ } \hat{\vec{u}}_1 \text{ } \hat{\vec{u}}_2]$ 
        \State Find $\vec{v}$ by solving the dual problem:
        \begin{align*}
            &\max_{\vec{v}\in \reals^n} \vec{v}^T \vec{y} ~\text{ s.t. } ~\big\vert\vec{v}^T \relu{\data \firstw} \big \vert \le 1\, \forall \firstw\in \ball_2  
        \end{align*}
        \State Check the existence of a violating neuron:
        \begin{align*}
            \max_{\firstw:\data \firstw  \succcurlyeq \vec{0}, \| \firstw \|_2 \leq 1} \dual^T \data \firstw  \overset{?}{\geq} 1 \quad \min_{\firstw:\data \firstw  \succcurlyeq \vec{0}, \| \firstw \|_2 \leq 1} \dual^T \data \firstw  \overset{?}{\leq} -1 ,
            \end{align*}
        \EndWhile
        \State Solve the training problem using $\firstwmat^*$:
        \begin{align*}
         \secondwvec^*=\argmin_{\secondwvec} \| \secondwvec \|_1 \text{ s.t. } \relu{\data \firstwmat^*}\secondwvec=\vec{y}
        \end{align*}
        \State Return $\theta=(\firstwmat^*,\vec{w}^*)$
		\end{algorithmic}\label{alg:pseudo1_nobias}	
	\end{algorithm}

\begin{props}\label{prop:cutting_optimality}
When $\data$ is spike-free as defined in Lemma \ref{lemma:spike-free}, the cutting plane based training method globally optimizes \eqref{eq:problem_dual}.
\end{props}
The following theorem shows that spike-free cases are in fact practically relevant. Particularly, random high dimensional i.i.d. Gaussian matrices asymptotically satisfy the spike-free condition as detailed below.
\begin{theos}\label{theo:asymptotic_spikefree}
Let $\data\in \reals^{n\times d}$ be an i.i.d. Gaussian random matrix. Then $\data$ is asymptotically spike-free as $d\rightarrow \infty$. More precisely, we have 
\begin{align*}
    \lim_{d\rightarrow \infty} \mathbb{P} \left[ \max_{\vec{u}\in\ball_2} \|\data^\dagger \relu{\data \vec{u}}\|_2 >1 \right] = 0\,.
\end{align*}
\end{theos}
We now consider improving the basic spike-free relaxation by including the extreme points along the ordinary basis vectors $\{\vec{e}_i\}_{i=1}^n$, which is detailed in Lemma \ref{lemma:convex_mixtures}. The next result characterizes the cases where employing only these extreme points are sufficient to fit the training data.
\begin{theos}\label{theo:convex_hull}
Let $\mathcal{C}_a$ denote the convex hull of $\{ \sample_i\}_{i=1}^n$. If each sample is a vertex of $\mathcal{C}_a$, then a feasible solution to \eqref{eq:problem_def2} can be achieved with $n$ neurons, which are the extreme points in the span of the ordinary basis vectors. Consequently, the weights given in Lemma \ref{lemma:convex_mixtures} achieve zero training error.
\end{theos}
This shows that the extreme points in Lemma \ref{lemma:convex_mixtures} enable the network architecture to achieve zero training error and therefore completely characterize the training data geometry when each data sample is a vertex of the convex hull of all the samples.

Our next result shows that the above claim, i.e., `` each data sample is a vertex of the convex hull of all the samples", is likely to hold for high dimensional random matrices.
\begin{theos}\label{theo:random_convexhull}
 Let $\data \in \mathbb{R}^{n\times d}$ be a data matrix generated i.i.d. from a standard Gaussian distribution $\mathcal{N}(0,1)$. Suppose that  the dimensions of the data matrix obey $d >2n\log(n-1)$. Then, every row of $\data$ is an extreme point of the convex hull of the rows of $\data$ with high probability.
\end{theos}

\section{ Regularized ReLU Networks and Convex Optimization for Spike-Free Matrices}\label{sec:main_regconvex}
In this section, we present extensions of our approach to regularized networks, arbitrary convex loss functions, and vector outputs. More importantly, as a corollary of our analysis in the previous section, we provide exact convex formulations for problems with spike-free data matrices and show that convex equivalent formulations can be globally optimized in polynomial-time by a standard convex solver.

\subsection{Regularized two-layer ReLU networks}\label{sec:regularized}
Here, we first formulate a penalized version of the equality form in \eqref{eq:problem_def2}. We then present duality results for this case.
\begin{theos} \label{theo:regularized_dual}
Optimal hidden neurons $\firstwmat$ for the following regularized problem
\begin{align}
    \label{eq:problem_def_regularized}
    \min_{\theta \in \Theta} \frac{1}{2}\|(\data\firstwmat)_{+}\secondwvec -\vec{y}\|_2^2 + \frac{\beta}{2} \left(\| \secondwvec \|_2^2+ \| \firstwmat \|_F^2\right) 
    =
    \min_{\theta \in \Theta} \frac{1}{2}\|(\data\firstwmat)_{+}\secondwvec -\vec{y}\|_2^2 + \beta \| \secondwvec \|_1 \text{ s.t. } \| \firstw_j \|_2\leq1 , \forall j,
\end{align}
can be found through  the following dual problem
\begin{align*}
    &\max_{\dual}  - \frac{1}{2}\|\dual-\vec{y}\|_2^2+\frac{1}{2} \| \vec{y}\|_2^2 \mbox{ s.t. } \dual \in \beta \rectset^\circ, -\dual \in \beta \rectset^{\circ}\,,
\end{align*}
where $\beta$ is the regularization (weight decay) parameter. Here $\rectset^\circ$ is the convex polar of the rectified ellipsoid $\rectset$.
\end{theos}

\bremark \label{remark:regularized}
Based on Theorem \ref{theo:regularized_dual}, we note that all the weak and strong duality results in Theorem \ref{theo:weak_duality} and \ref{theo:strong_duality} hold for regularized networks. Therefore, Corollary \ref{cor:1d_strong_duality}, \ref{cor:rankone_strong_duality} and \ref{cor:whitened_strong_duality} also apply to regularized networks. Furthermore, the numerical results in Figure \ref{fig:dual} confirm this claim.
\eremark
\begin{cors}\label{cor:kernel_squareloss}
Remark \ref{remark:regularized} also implies that whenever the set of extreme points can be explicitly characterized, e.g., problems involving rank-one and/or whitened data matrices, we can solve \eqref{eq:problem_def2} as a convex $\ell_1$-norm minimization problem to achieve the optimal solutions. Particularly, we first construct a hidden layer weight matrix, i.e., denoted as $\firstwmat^*$, using all possible extreme points. We then solve the following problem
\begin{align}\label{eq:problemdef_kernel}
    \min_{\secondwvec} \| \secondwvec \|_1 \text{ s.t. } \relu{\data \firstwmat^*}\secondwvec=\vec{y}
\end{align}
or the corresponding regularized version
\begin{align*}
    \min_{\secondwvec}  \frac{1}{2}\|\relu{\data \firstwmat^*}\secondwvec-\vec{y}\|_2^2+\beta \| \secondwvec \|_1.
\end{align*}
\end{cors}

Next, we provide the closed-form formulations for the optimal solutions to the regularized training problem \eqref{eq:problem_def_regularized} as in Lemma \ref{lemma:closedform_2neuron}.

\begin{theos} \label{theo:closedform_regularized}
Suppose $\data$ is whitened such that $\data\data^T=\vec{I}_n$, then an optimal solution set for \eqref{eq:problem_def_regularized} can be formulated as follows
\begin{align*}
   \left(\firstwmat^* ,\secondwvec^*\right)= \begin{cases}
   \left(\left[ \frac{\data^\dagger \relu{\vec{y}}}{\|\data^\dagger \relu{\vec{y}}\|_2}, \frac{\data^\dagger \relu{\vec{-y}}}{\|\data^\dagger \relu{\vec{-y}}\|_2}\right], \begin{bmatrix}\|\data^\dagger \relu{\vec{y}}\|_2-\beta\\ -\|\data^\dagger \relu{\vec{-y}}\|_2+\beta\end{bmatrix} \right) &\text{ if } \beta \leq \|\relu{\vec{y}}\|_2,\;  \beta \leq  \|\relu{\vec{-y}}\|_2\\\
     \hfil  \left(\frac{\data^\dagger \relu{\vec{-y}}}{\|\data^\dagger \relu{\vec{-y}}\|_2}, -\|\data^\dagger \relu{\vec{-y}}\|_2+\beta\right) &\text{ if } \beta > \|\relu{\vec{y}}\|_2 , \;  \beta \leq \|\relu{\vec{-y}}\|_2\\
      \hfil \left(\frac{\data^\dagger \relu{\vec{y}}}{\|\data^\dagger \relu{\vec{y}}\|_2}, \|\data^\dagger \relu{\vec{y}}\|_2-\beta\right) &\text{ if } \beta \leq \|\relu{\vec{y}}\|_2,\;  \beta >  \|\relu{\vec{-y}}\|_2\\
    \hfil  \left(\vec{0},0\right) &\text{ if } \beta > \|\relu{\vec{y}}\|_2,\;  \beta >  \|\relu{\vec{-y}}\|_2\\
    \end{cases} .
\end{align*}
\end{theos}
We note that Theorem \ref{theo:closedform_regularized} exactly characterizes how the regularization parameter $\beta$ controls the number of neurons and changes the analytical form of the optimal network parameters. Therefore, for whitened data matrices, there is no need to train a network via backpropagation or other numerical methods, instead, one can directly utilize the closed-form solutions in Theorem \ref{theo:closedform_regularized}.

\subsection{Two-layer ReLU networks with hinge loss}
Now we consider classification problems with the label vector $\vec{y} \in \{+1,-1\}^n$ and hinge loss. 
\begin{theos}\label{theo:hinge_dual}
An optimal $\firstwmat$ for the binary classification task with the hinge loss given by
\begin{align}\label{eq:problem_def_hinge}
    &\min_{\theta \in \Theta} \sum_{i=1}^n\max\{0,1-y_i(\sample_i^T\firstwmat)_{+}\secondwvec \} + \beta \| \secondwvec \|_1  \text{ s.t. } \| \firstw_j \|_2\leq1 , \forall j,
\end{align}
can be found through the following dual
\begin{align*}
    &\max_{\dual}  \dual^T \vec{y} \mbox{ s.t. } 0\le y_i\dualscalar_i \le 1~\forall i\in[n] ,\dual \in \beta \rectset^\circ, -\dual \in \beta \rectset^{\circ}\,.
\end{align*}
\end{theos}
Theorem \ref{theo:hinge_dual} proves that since strong duality holds for two-layer NNs, we can obtain the optimal solutions to \eqref{eq:problem_def_hinge} through the dual form. The following corollary characterizes the solutions obtained via the dual form of \eqref{eq:problem_def_hinge}.
\begin{cors} \label{cor:extreme_points_hinge}
Theorem \ref{theo:hinge_dual} implies that the optimal neuron weights are extreme points which solve the following optimization problem
\begin{align*}
    \argmax_{\firstw \in \ball_2} \left \vert {\dual^*}^T \relu{\data\firstw}\, \right \vert,
\end{align*}
where $\|\dual^*\|_{\infty}\leq1$.
\end{cors}

Consequently, in the one dimensional case, the optimal neuron weights are given by the extreme points. Therefore the optimal network network output is given by the piecewise linear function 
\begin{align*}
    f(\vec{a}) = \sum_{j=1}^m \secondw_j(\vec{a} u_j+b_j)_+\,,
\end{align*}
for some output weights $w_1, \ldots,w_m$ where $u_j=\pm 1$ and $b_j=\mp\samplescalar_j$ for some $j$. 

This explains Figure \ref{fig:hinge}, where the decision regions are determined by the zero crossings of the above piecewise linear function. Moreover, the dual problem reduces to a finite dimensional minimum $\ell_1$-norm Support Vector Machine (SVM), whose solution can be easily determined. As it can be seen in Figure 1c, the piecewise linear fit passes through the data samples which are on the margin, i.e., the network output is $\pm 1$. This corresponds to the maximum margin decision regions and separates the green shaded area from the red shaded area.

\begin{figure*}[t]
	\centering
	\captionsetup[subfigure]{oneside,margin={1cm,0cm}}
		\begin{subfigure}[t]{0.32\textwidth}
		\centering
		\includegraphics[width=1.1\textwidth, height=0.8\textwidth]{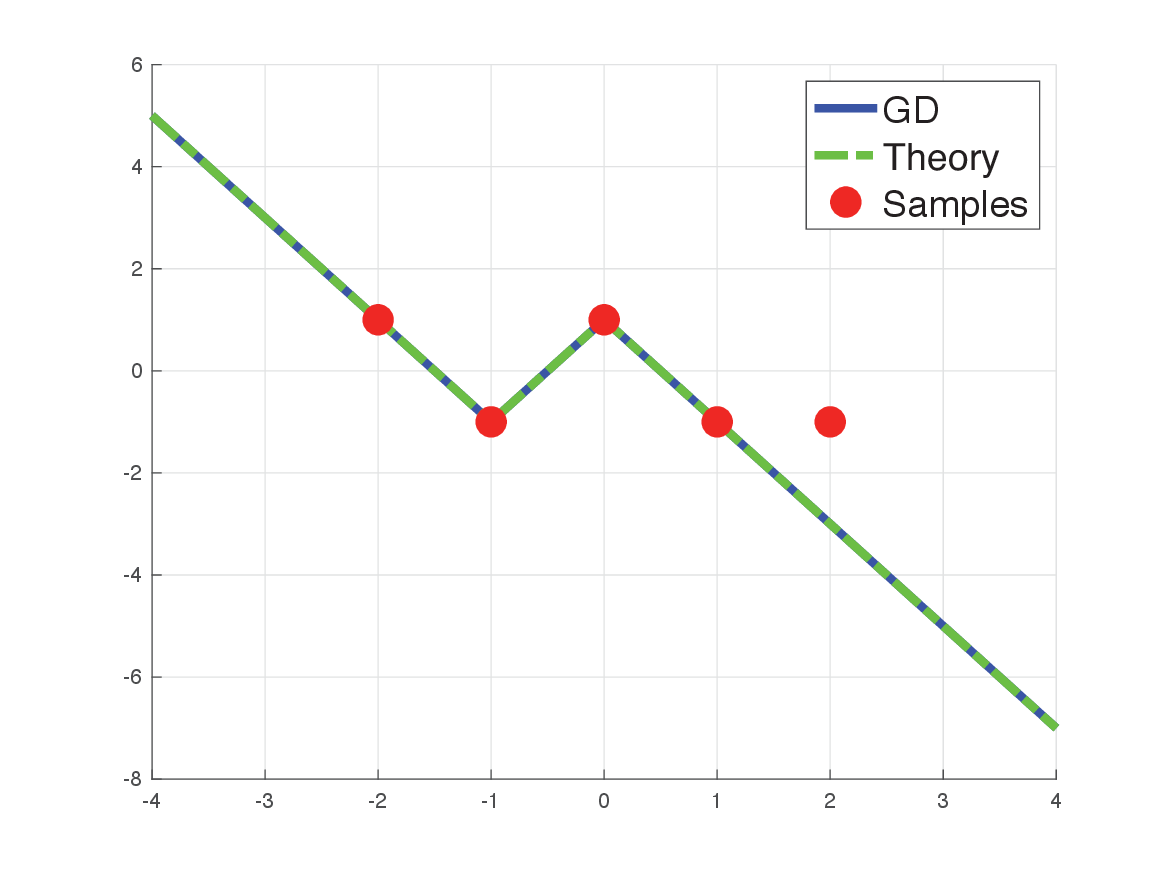}
		\caption{$L_t=1.600\times 10^{-4}$ and $L_{gd}=1.600 \times 10^{-4}$. \\ GD and our theory agrees. \centering} \label{fig:hinge_comp1}
	\end{subfigure} \hspace*{\fill}
			\begin{subfigure}[t]{0.32\textwidth}
		\centering
 		\includegraphics[width=1.1\textwidth, height=0.8\textwidth]{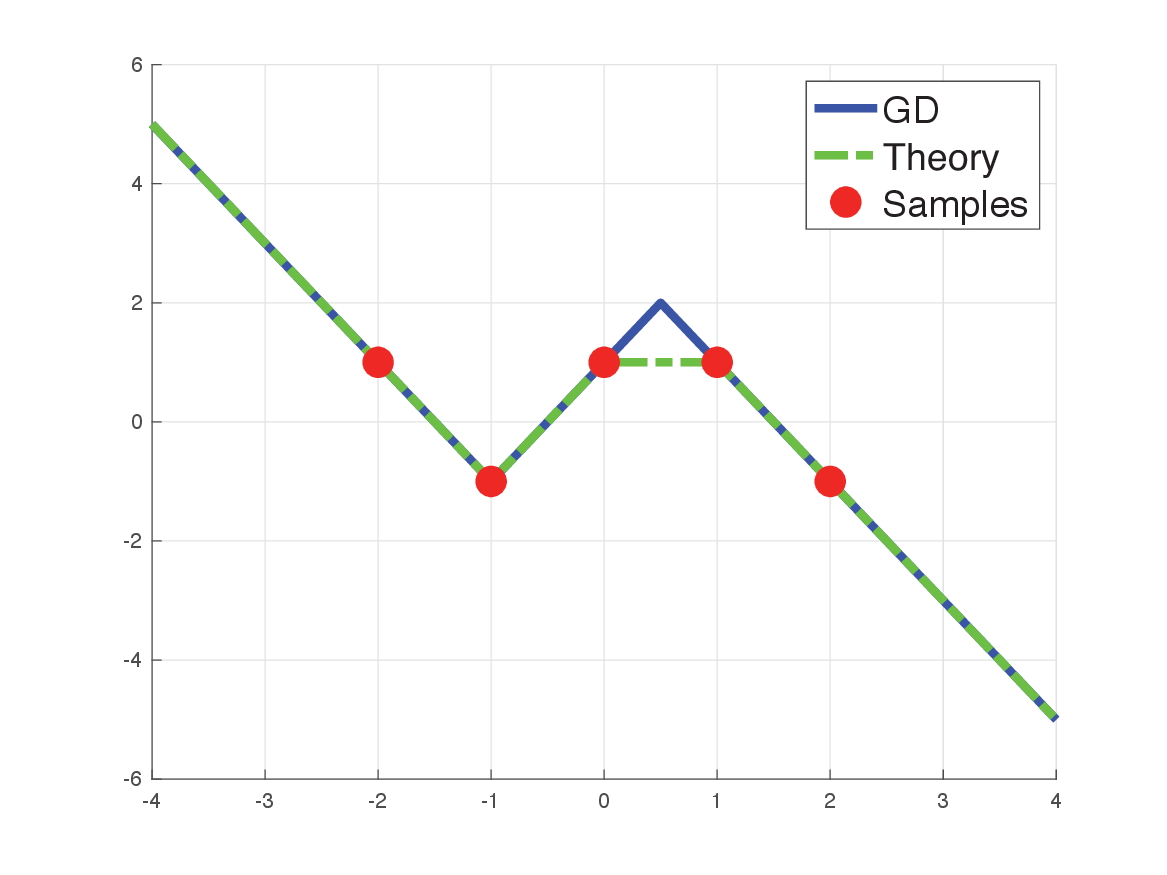}
		\caption{$L_t=1.600\times 10^{-4}$ and $L_{gd}=1.679 \times 10^{-4}$.\centering} \label{fig:hinge_comp2}
	\end{subfigure} \hspace*{\fill}
		\begin{subfigure}[t]{0.32\textwidth}
		\centering
		\includegraphics[width=1.1\textwidth, height=0.8\textwidth]{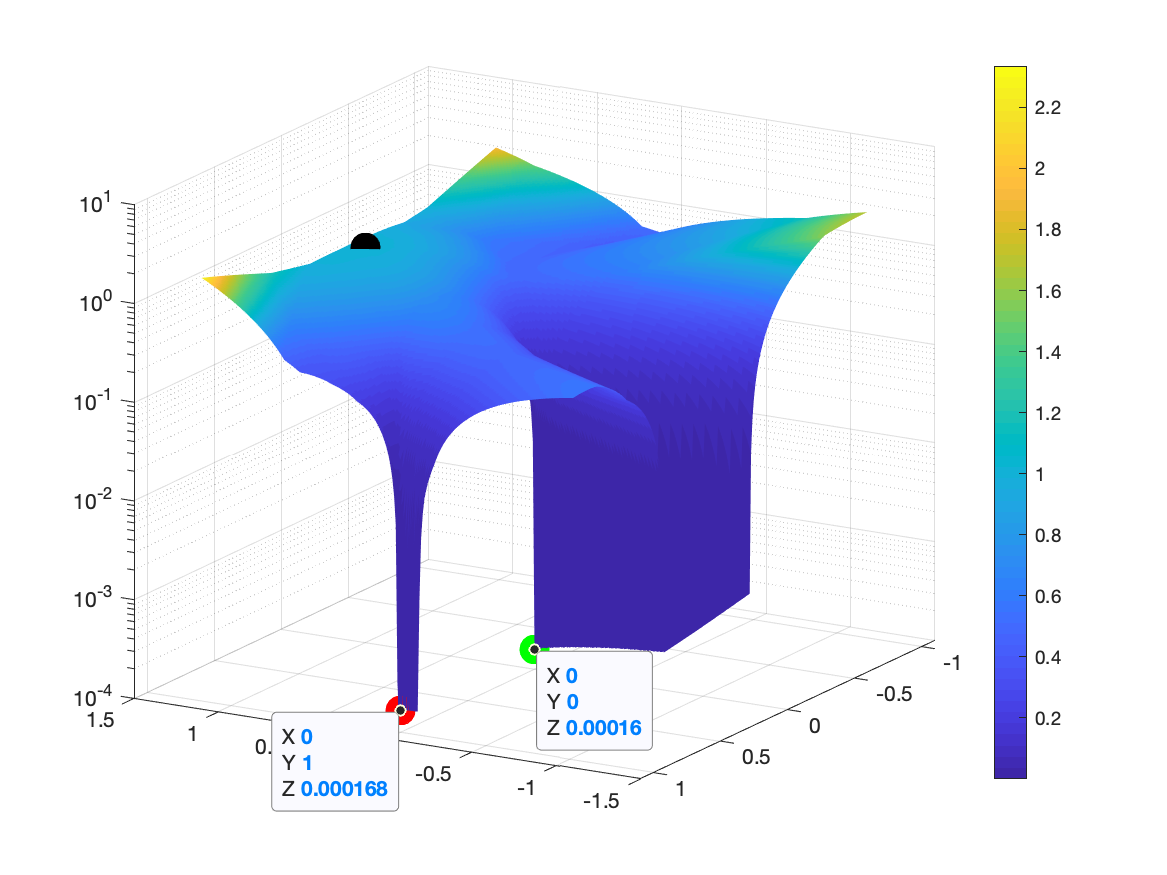}
		\caption{Visualization of the loss landscape in (b) ($L_t=1.600\times 10^{-4}$ and $L_{gd}=1.679 \times 10^{-4}$).\centering} \label{fig:hinge_3d}
	\end{subfigure} \hspace*{\fill}
	\medskip
	\caption{Binary classification using hinge loss, where we apply GD and our approach in Theorem \ref{theo:hinge_svm}. Here, we denote the objective value in \eqref{eq:problem_def_hinge} as $L_t$ and $L_{gd}$ for our theoretical approach (Theorem \ref{theo:hinge_svm}) and GD, respectively. We also note that optimal solution might not be unique as shown in Proposition \ref{prop:1d_uniqueness} and Figure \ref{fig:uniquness}, which explains why GD converges to a solution with a kink in the middle of two data points in (b). In (c), we also provide 3D illustration of the loss surface of the example in (b), where we mark the initial point (black), the GD solution (red), and our solution (green).  }
\label{fig:hinge_fit}
	\end{figure*}

It is easy to see that this is equivalent to applying the kernel map $\kappa(\samplescalar,\samplescalar_j)=(\samplescalar-\samplescalar_j)_+$, forming the corresponding kernel matrix
\begin{align*}
    \vec{K}_{ij} = (\samplescalar_i-\samplescalar_j)_+\,,
\end{align*}
and solving minimum $\ell_1$-norm SVM on the kernelized data matrix. The same steps can also be applied to a rank-one dataset as presented in the following.

\begin{cors}\label{cor:rankone_extreme_hinge}
Let $\data$ be a data matrix such that $\data=\vec{c}\sample^T$, where $\vec{c} \in \mathbb{R}^n$ and $\sample \in \mathbb{R}^d$. Then, a set of solutions  to \eqref{eq:problem_def_hinge} that achieve the optimal value are extreme points, and therefore satisfy $\{(\firstw_i, \bias_i) \}_{i=1}^{m}$, where  $\firstw_i=s_i\frac{\vec{a}}{\|\vec{a}\|_2}, \bias_{i}=- s_i c_{i}\|\vec{a}\|_2$ with $s_i=\pm 1, \forall i \in [m]$. 
\end{cors}

\begin{theos} \label{theo:hinge_svm}
For a rank-one dataset $\data=\vec{c}\sample^T $, applying $\ell_1$-norm SVM on $\relu{\data \firstwmat^{*^T}+\vec{1}\biasvec^{*^T}}$ finds the optimal solution $\theta^*$ to \eqref{eq:problem_def_hinge}, where $\firstwmat^{*}\in \mathbb{R}^{d \times 2n}$ and $\biasvec^{*} \in \mathbb{R}^{2n}$ are defined as $\{\firstw_i^{*}=s_i  \frac{\vec{a}}{\|\vec{a}\|_2} ,\bias_i^{*}=- s_i c_{i}\|\vec{a}\|_2\}_{i=1}^{n} $ with $s_i=\pm 1, \forall i$.
\end{theos}
The proof of Corollary \ref{cor:rankone_extreme_hinge} and Theorem \ref{theo:hinge_svm} directly follows from the proof of Corollary \ref{cor:rankone_extreme}. 

We also verify Theorem \ref{theo:hinge_svm} using a one dimensional dataset in Figure \ref{fig:hinge_fit}. In this figure, we observe that whenever there is a sign change, the corresponding two samples determine the decision boundary, which resembles the idea of support vector. Thus, the piecewise linear fit passes through these samples. On the other hand, when there is no sign change, the piecewise fit does not need to create any kink as in Figure \ref{fig:hinge_comp1}. However, we also note that optimal solution is not unique and there might exist other optimal solutions as shown in Proposition \ref{prop:1d_uniqueness} and Figure \ref{fig:uniquness}. This is exactly what we observe in in Figure \ref{fig:hinge_comp2}, where GD try to converge to a solution with a kink in the middle of two data points unlike our approach. In Figure \ref{fig:hinge_3d}, we also provide a visualization of the loss landscape for this case.

\begin{theos}\label{theo:closedform_2neuron_hinge}
Suppose that $\data$ is whitened such that $\data\data^T=\vec{I}_n$. Then an optimal solution to the problem in \eqref{eq:problem_def_hinge} is given by
\begin{align*}
    &(\firstw_1^*,\secondw_1^*) =\begin{cases}\left(\frac{\data^\dagger \relu{\vec{y}}}{\sqrt{n_+}} ,   \secondw_+\right)  &\text{ if } \beta = \sqrt{n_+} \\
  \hfil \left(\frac{\data^\dagger \relu{\vec{y}}}{\sqrt{n_+}} , \sqrt{n_+} \right)&\text{ if } \beta < \sqrt{n_+} \\
    \hfil \left( \vec{0},0\right) &\text{ if } \beta > \sqrt{n_+}
    \end{cases} \quad  (\firstw_2^*,\secondw_2^*) =\begin{cases} \left(\frac{\data^\dagger \relu{-\vec{y}}}{\sqrt{n_-}} ,\secondw_-\right) &\text{ if } \beta = \sqrt{n_-} \\
  \hfil \left(\frac{\data^\dagger \relu{-\vec{y}}}{\sqrt{n_-}} ,-\sqrt{n_-} \right)&\text{ if } \beta < \sqrt{n_-} \\
    \hfil \left(\vec{0},0\right) &\text{ if } \beta > \sqrt{n_-}
    \end{cases}\,,
\end{align*}
where $\secondw_{+} \in [0,\sqrt{n_+}]$, $\secondw_{-} \in [-\sqrt{n_-},0]$, $n_+$, and $n_-$ are the number of samples with positive and negative labels, respectively.
\end{theos}

 Theorem \ref{theo:closedform_2neuron_hinge} shows that the well-known problem achieving minimum $\ell_2$-norm parameters maximizing the margin between two classes can be solved in closed-form for some generic settings such as problems involving a spike-free and/or whitened data matrix. As in the squared loss case, we observe that the weight decay parameter $\beta$ directly controls the sparsity of the optimal solution. 
 
 \paragraph{Interpretation of the hidden neurons as Fisher's Linear Discriminant vectors:\\}
 \noindent According to Theorem \ref{theo:closedform_2neuron_hinge}, it is interesting to note that for generic full column rank matrices and binary labels $\vec{y}\in \{+1,-1\}^n$, the expression for the first hidden neuron equals $\firstw_1^*=\data^\dagger (\vec{y})_+ = (\data^T \data)^{-1}\data^T(\vec{y})_+ =  \hat{\boldsymbol{\Sigma}}^{-1} \hat{\boldsymbol{\mu}}_1$ where $\hat{ \boldsymbol{\Sigma}}:=\frac{1}{n} \data^T \data$ is the empirical covariance matrix and $\boldsymbol{\mu}_1 = \frac{1}{n} \data^T(\vec{y})_+ = \frac{1}{n} \sum_{i:y_i>0} \sample_i $ is the empirical mean of the samples in the first class. In fact, this formula is identical to Fisher's Linear Discriminant for binary classification.  For whitened data, note that $\hat{\boldsymbol{\Sigma}} $ is a multiple of the identity matrix. This shows that there are two optimal hidden neurons for two individual classes corresponding to the $\{+1,-1\}$ labels. Furthermore, a particular hidden neuron is only active when the weight decay parameter $\beta$ is less than the square root of the number of samples in the corresponding class, i.e., $\beta<\sqrt{n_{-}}$ or $\beta<\sqrt{n_{+}}$. In theorem \ref{theo:closedform_regularized_multiclass}, we provide a closed-form expression for the multi-class case where the number of hidden neurons can be as large as the number of classes and is controlled by the magnitude of $\beta$.

\subsection{Two-layer ReLU networks with general loss functions}
Now we consider the scalar output two-layer ReLU networks with a generic loss 
\begin{align}
    \min_{\theta \in \Theta} \ell((\data\firstwmat)_{+}\secondwvec,\vec{y}) + \frac{\beta}{2} (\| \secondwvec \|_2^2+\|\firstwmat\|_F^2) =\min_{\theta \in \Theta} \ell((\data\firstwmat)_{+}\secondwvec,\vec{y}) + \beta \| \secondwvec \|_1 \text{ s.t. } \| \firstw_j \|_2\leq1 , \forall j,\label{eq:gen_loss}
\end{align}
where $\ell(\cdot,\vec{y})$ is a convex loss function. 
\begin{theos}\label{theo:generic_dual}
The dual of \eqref{eq:gen_loss} is given by
\begin{align*}
    &\max_{\dual}  - \ell^*(\dual) \mbox{ s.t. } \dual \in \beta \rectset^\circ, -\dual \in \beta \rectset^{\circ}\,,
\end{align*}
where $\ell^*$ is the Fenchel conjugate function defined as \citep{boyd_convex}
\begin{align*}
\ell^*(\dual) = \max_{\vec{z}} \vec{z}^T \dual - \ell(\vec{z},\vec{y})\,.
\end{align*}
\end{theos}
Theorem \ref{theo:generic_dual} proves that our extreme point characterization in Lemma \ref{lemma:extreme_l2loss} applies to arbitrary convex loss functions. Therefore, the optimal parameters for \eqref{eq:2layer_function} is a subset of the same extreme point set, i.e., determined by the input data matrix $\data$, independent of the loss function.

\subsection{Polynomial-time convex optimization of problems with spike-free matrices} \label{sec:polynomialtime_convex} \label{sec:spkiefree_polynomial}
Here, we show that two-layer ReLU network training problems involving spike-free data matrices can be equivalently stated as convex optimization problems. More importantly, we prove that the equivalent form can be globally optimized by standard convex optimization solvers with polynomial complexity in terms of the number of data samples $n$ and the data dimension $d$.

We start by restating the two-layer training problem with arbitrary convex loss functions as follows
\begin{align}
    \min_{\theta \in \Theta} \ell((\data\firstwmat)_{+}\secondwvec,\vec{y}) + \beta \| \secondwvec \|_1 \text{ s.t. } \| \firstw_j \|_2\leq1 , \forall j.\label{eq:gen_loss_polynomial}
\end{align}
Then, by Theorem \ref{theo:equality_dual} and \ref{theo:generic_dual}, we have the following dual problem with respect to the output layer weights $\secondwvec$
\begin{align}\label{eq:gen_loss_dual_polynomial}
    &\max_{\dual}  - \ell^*(\dual) \mbox{ s.t. } \max_{\substack{\firstw \in \ball_2\\\data \firstw \succcurlyeq \vec{0} }}\big\vert\vec{v}^T \data \firstw \big \vert \le 1,
\end{align}
where we replace $ (\data \firstw)_+$ with $\{\data\firstw: \data\firstw\succcurlyeq \vec{0} \}$ since $\data$ is spike-free. If we take the dual of \eqref{eq:gen_loss_dual_polynomial} with respect to $\dual$ one more time, i.e., also known as the bidual of \eqref{eq:gen_loss_polynomial}, we obtain the following optimization problem
\begin{align}\label{eq:gen_loss__bidual_polynomial}
    \min_{\vec{w}_1,\vec{w}_2} \ell( \data(\vec{w}_1-\vec{w}_2),\vec{y})+ \beta (\|\vec{w}_1\|_2+\|\vec{w}_2\|_2) \text{ s.t. } \data \vec{w}_1 \succcurlyeq \vec{0}, \; \data \vec{w}_2 \succcurlyeq \vec{0}.
\end{align}
Note that \eqref{eq:gen_loss__bidual_polynomial} is a convex optimization problem with $2d$ variables and $2n$ constraints. Therefore, standard interior-point solvers, e.g., CVX \cite{cvx} with the SDPT3 solver \cite{tutuncu2001sdpt3}, to solve convex optimization problems in Section 4.4. can globally optimize \eqref{eq:gen_loss__bidual_polynomial} with the computational complexity $\mathcal{O}(d^3)$.

\subsection{Extension to vector output neural networks}
\label{sec:vectoroutput}
In this section, we first derive the dual form for vector output NNs and then describe the implementation of the cutting plane algorithm. For presentation simplicity, we consider the weakly regularized scenario with squared loss. However, all the derivations in this section can be extended to regularized problems with arbitrary convex loss functions as proven in the previous section.

Here, we have  $\vec{Y} \in \mathbb{R}^{n \times o}$ and $f(\data) =  \relu{\data \firstwmat} \secondwmat \,$, where $\secondwmat \in \reals^{m\times o}$. Then, the training problem is as follows
\begin{align*}
    &\min_{\theta \in \Theta } \|\secondwmat\|_F^2 + \| \firstwmat\|_F^2 \text{ s.t. } \relu{\data \firstwmat} \secondwmat=\vec{Y}.
\end{align*}

\begin{lems} \label{lemma:reg_equivalence_multiclass}
The following two optimization problems are equivalent:
\begin{align*}
  &\min_{\theta \in \Theta} \|\secondwmat\|_F^2 + \| \firstwmat\|_F^2   \text{ s.t. } \relu{\data \firstwmat} \secondwmat=\vec{Y} \hspace{.5cm} =\hspace{.5cm} \min_{\theta \in \Theta}\sum_{j=1}^m \|\secondwvec_j\|_2  \text{ s.t. } \relu{\data \firstwmat} \secondwmat=\vec{Y}, \; \| \firstw_j \|_2 \le 1, \; \forall j
\end{align*}
\end{lems}

Using Lemma \ref{lemma:reg_equivalence_multiclass}, we get the following equivalent form
\begin{align} \label{eq:problem_def_multiclass}
       &\min_{\theta \in \Theta } \sum_{j=1}^m \|\secondwvec_j\|_2  \text{ s.t. } \relu{\data \firstwmat} \secondwmat=\vec{Y}, \; \| \firstw_j \|_2 \le 1, \; \forall j,
\end{align}
which has the following dual form
\begin{align}\label{eq:dual_multiclass}
    \max_{\dualmat} \trace(\dualmat^T \vec{Y}) \text{ s.t. } \| \dualmat^T (\data \firstw)_+\|_2 \le 1, \; \firstw \in \ball_2.
\end{align}
Then, we again relax the problem using the spike-free relaxation and then we solve the following problem to achieve the extreme points
\begin{align}\label{eq:extreme_vectoroutput}
\begin{split}
    &\hat{\firstw}=\argmax_{\firstw} \|\dualmat^T \data \firstw\|_2 \text{ s.t. } \data \firstw  \succcurlyeq \vec{0}, \| \firstw \|_2 \leq 1
    \end{split} 
\end{align}
Therefore, the hidden layer weights can be determined by solving the above optimization problem.

\subsubsection{Solutions to one dimensional problems}
Here, we consider a vector output problem with a one dimensional data matrix, i.e., $\data=\vec{a}$, where $\vec{a} \in \mathbb{R}^n$. Then, the extreme points of \eqref{eq:dual_multiclass} can be obtained via the following maximization problem
\begin{align}\label{eq:extreme_multiclass_1d}
    \argmax_{\firstwscalar,\bias} \|\dualmat^T (\vec{a}\firstwscalar+ \bias \vec{1})_+\|_2^2 \text{ s.t. } |u|=1.
\end{align}
Using the same steps in Proof of Lemma \ref{lemma:extreme_l2loss}, we can write \eqref{eq:extreme_multiclass_1d} as follows
\begin{align} \label{eq:extreme_multilclass_1dv2}
    \argmax_{\firstwscalar, \bias}  \Big\|\sum_{i \in \mathcal{S}}\dual_i (\samplescalar_i \firstwscalar+\bias)\Big\|_2^2 \text{ s.t. } \samplescalar_i \firstwscalar+\bias \geq 0, \forall i \in \mathcal{S}, \samplescalar_j \firstwscalar+\bias \leq 0, \forall j \in \mathcal{S}^c, |\firstwscalar|= 1.
\end{align}
Notice that $\firstwscalar$ can be either $+1$ or $-1$. Thus, we can solve the problem for each option and then pick the one with higher objective value. First assume that $u=+1$, then \eqref{eq:extreme_multilclass_1dv2} becomes
\begin{align}\label{eq:extreme_multilclass_1d_u1}
    \argmax_{ \bias}  \Big\|\sum_{i \in \mathcal{S}}\dual_i (\samplescalar_i +\bias)\Big\|_2^2 \text{ s.t. } \max_{i \in \mathcal{S}}-\samplescalar_i\leq b \leq \min_{j \in \mathcal{S}^c} -\samplescalar_j.
\end{align}
Since 
\begin{align*}
\Big\|\sum_{i \in \mathcal{S}}\dual_i (\samplescalar_i +\bias)\Big\|_2^2=\Big\|\sum_{i \in \mathcal{S}}\dual_i \samplescalar_i\Big\|_2^2+2b\bigg(\sum_{i \in \mathcal{S}}\dual_i \samplescalar_i\bigg)^T \sum_{i \in \mathcal{S}}\dual_i +b^2 \Big\|\sum_{i \in \mathcal{S}}\dual_i\Big\|_2^2,
\end{align*}
\eqref{eq:extreme_multilclass_1d_u1} is a convex function of $b$. Therefore, the optimal solution to \eqref{eq:extreme_multilclass_1d_u1} is achieved when either $b=\min_{j \in \mathcal{S}^c} -\samplescalar_j$ or $b=\max_{i \in \mathcal{S}}-\samplescalar_i$ holds. Similar arguments also hold for $u=-1$.

\begin{cors}\label{cor:1d_extreme_multiclass}
Let $\{\samplescalar_i\}_{i=1}^n$ be a one dimensional  training set i.e., $\samplescalar_i \in \reals,~ \forall i\in[n]$. Then, the solutions  to \eqref{eq:problem_def_multiclass} that achieve the optimal value satisfy $\{(\firstwscalar_i, \bias_i) \}_{i=1}^{m}$, where  $\firstwscalar_i=\pm 1, \bias_{i}=- \sign(\firstwscalar_i) \samplescalar_{i}$. 
\end{cors}

\subsubsection{Solutions to rank-one problems}
Here, we consider a vector output problem with a rank-one data matrix, i.e., $\data=\vec{c}\sample^T$. Then, all possible extreme points can be characterized as follows
\begin{align*}
        \argmax_{\bias,\firstw:\|\firstw\|_2= 1 } \Big\|{\dualmat}^T \relu{\data \firstw+\bias \vec{1}}\,\Big\|_2^2&=\argmax_{\bias,\firstw:\|\firstw\|_2= 1 } \Big\|{\dualmat}^T \relu{\vec{c} \vec{a}^T \firstw +\bias\vec{1}}\,\Big\|_2^2\\
        &=\argmax_{\bias,\firstw:\|\firstw\|_2= 1 } \Big\|\sum_{i=1}^n \dual_i \relu{c_i \vec{a}^T \firstw+\bias }\,\Big\|_2^2,
\end{align*}
which can be equivalently stated as
\begin{align*}
    &\argmax_{\bias,\firstw:\|\firstw\|_2= 1} \Big\|\sum_{i \in \mathcal{S}}\dual_i c_i\Big\|_2^2(\sample^T \firstw)^2+2b(\sample^T \firstw)\bigg(\sum_{i \in \mathcal{S}}\dual_i c_i\bigg)^T \sum_{i \in \mathcal{S}}\dual_i +b^2 \Big\|\sum_{i \in \mathcal{S}}\dual_i\Big\|_2^2 \\&\text{ s.t. } \begin{cases} &c_i\sample^T \firstw+\bias \geq 0, \forall i \in \mathcal{S} \\
    &c_j\sample^T \firstw+\bias  \leq 0, \forall j \in \mathcal{S}^c\end{cases}
,\end{align*}
which shows that $\firstw$ must be either positively or negatively aligned with $\sample$, i.e.,
$\firstw= s \frac{\sample}{\|\sample\|_2}$, where $s= \pm 1$. Thus, $\bias$ must be in the range of $[\max_{i \in \mathcal{S}}(-s c_i \| \sample\|_2), \text{ }\min_{j \in \mathcal{S}^c}(-s c_j \| \sample\|_2)]$ Using these observations, extreme points can be formulated as follows
\begin{align*}
    \vec{u}_v= \begin{cases}\frac{\sample}{\| \sample\|_2} &\text{if } \sum_{i \in \mathcal{S}} \dualscalar_i c_i \geq 0\\\frac{-\sample}{\| \sample\|_2} & \text{otherwise}\end{cases}
    \text{ and } \bias_v= \begin{cases}\min_{j \in \mathcal{S}^c}(-s_v c_j \| \sample\|_2) &\text{if } \sum_{i \in \mathcal{S}} \dualscalar_i \geq 0\\\max_{i \in \mathcal{S}}(-s_v c_i \| \sample\|_2) & \text{otherwise}\end{cases},
\end{align*}
where $s_v= \sign(\sum_{i \in \mathcal{S}} \dualscalar_i c_i)$.
\begin{cors}\label{cor:rankone_extreme_multiclass}
Let $\data$ be a data matrix such that $\data=\vec{c}\sample^T$, where $\vec{c} \in \mathbb{R}^n$ and $\sample \in \mathbb{R}^d$. Then, the solutions  to \eqref{eq:problem_def_multiclass} that achieve the optimal value are extreme points, and therefore satisfy $\{(\firstw_i, \bias_i) \}_{i=1}^{m}$, where  $\firstw_i=s_i\frac{\vec{a}}{\|\vec{a}\|_2}, \bias_{i}=- s_i c_{i}\|\vec{a}\|_2$ with $s_i=\pm 1, \forall i \in [m]$. %
\end{cors}

\begin{theos}\label{theo:rankone_multiclass}
For one dimensional and/or rank-one datasets, solving the following $\ell_2$-norm convex optimization problem globally optimizes \eqref{eq:problem_def_multiclass}
\begin{align*}
       &\min_{\secondwmat } \sum_{j=1}^m \|\secondwvec_j\|_2  \text{ s.t. } \relu{\data \firstwmat_e} \secondwmat=\vec{Y},
\end{align*}
where $\firstwmat_e \in \mathbb{R}^{d \times m_e}$ is a weight matrix consisting of all possible extreme points.
\end{theos}
Theorem \ref{theo:rankone_multiclass} proves that for one dimensional and/or rank-one datasets, the regularized non-convex training problem in \eqref{eq:problem_def_multiclass} can be equivalently stated as a convex optimization problem where the hidden neurons are fixed. Therefore, the training problem can be globally and efficiently optimized via standard convex optimization solvers.

\subsubsection{Solutions to Whitened Problems} 
Here, we provide the closed-form formulations for the optimal solutions to the regularized training problem \eqref{eq:problem_def_multiclass} as in Theorem \ref{theo:closedform_regularized}.

\begin{theos} \label{theo:closedform_regularized_multiclass}
Let $\{\data,\vec{Y}\}$ be a dataset such that $\data\data^T=\vec{I}_n$ and $\vec{Y}$ is one hot encoded label matrix, then a set of optimal solution $(\firstwmat^*,\secondwmat^*)$ to 
\begin{align*} 
       &\min_{\theta \in \Theta } \frac{1}{2}\|\relu{\data \firstwmat} \secondwmat-\vec{Y}\|_F^2+\frac{\beta}{2}\left(\|\firstwmat\|_F^2+\|\secondwmat\|_F^2\right)
\end{align*}
can be formulated as follows
\begin{align*}
   \left(\firstw_j^* ,\secondwvec_j^*\right)= \begin{cases}
   \left(\frac{\data^\dagger \vec{y}_j}{\|\data^\dagger \vec{y_j}\|_2}, \left(\sqrt{n_j}-\beta\right)\vec{e}_j\right) &\text{ if } \beta \leq \sqrt{n_j}\\\
    \hfil (\vec{0,\vec{0}}) &\text{ otherwise } 
    \end{cases},\quad \forall j \in [o],
\end{align*}
where $n_j$ is the number samples in class $j$ and $\vec{e}_j$ is the $j^{th}$ ordinary basis vector.
\end{theos}

The above result shows that there are at most $o$ hidden neurons, which individually correspond to classifying a particular class. A hidden neuron is only active when the weight decay parameter $\beta$ is less than the square root of the number of samples in the corresponding class. It is interesting to note that the form of the hidden neuron is identical to Fisher's Linear Discriminant for binary classification applied in a one-vs-all fashion.

\subsubsection{Convex optimization for spike-free problems} 
We now analyze the case when the data matrix is spike-free. Following the approach in Section \ref{sec:spkiefree_polynomial}, we first state the dual problem as follows
\begin{align}\label{eq:dual_spikefree_multiclass}
    \max_{\dualmat} \trace(\dualmat^T \vec{Y}) \text{ s.t. } \max_{\substack{\data \firstw \succcurlyeq \vec{0}\\ \firstw \in \ball_2}}\| \dualmat^T \data \firstw\|_2 \le 1, \;  \forall \firstw \in \ball_2.
\end{align}
Then, the the corresponding bidual problem is
\begin{align}\label{eq:bidual_spikefree_multiclass}
    \min_{\{\vec{w}_{1j},\vec{w}_{2j}\}_{j=1}^m} \sum_{j=1}^m \|\vec{w}_{1j}\|_2 \|\vec{w}_{2j}\|_2 \text{ s.t. } \sum_{j=1}^m \data \vec{w}_{1j}\vec{w}_{2j}^T=\vec{Y},\; \data \vec{w}_{1j} \succcurlyeq \vec{0}, \, \forall j,
\end{align}
which can be stated as a convex optimization problem as
\begin{align}\label{eq:bidual_spikefree_convex_multiclass}
    \min_{\vec{M} \in \mathcal{C}}  \|\vec{M}\|_* \text{ s.t. } \data \vec{M}=\vec{Y}, \,
\end{align}
as long as $m\ge m^*:=\text{rank}(\vec{M}^*)$ where $\vec{M}^*$ is an optimal solution, $\mathcal{C}:= \conv\{\vec{w}_1 \vec{w}_2^T : \data\vec{w}_{1} \succcurlyeq \vec{0},\, \vec{w}_1 \in \mathbb{R}^d,\, \vec{w}_2 \in \mathbb{R}^o\}$, $\|\cdot \|_{*}$ denotes the nuclear norm, and $\conv$ represents the convex hull of a set. We remark that the problem in \eqref{eq:bidual_spikefree_convex_multiclass} resembles convex semi non-negative matrix factorizations, such as the ones studied in \citet{semi_nmf,sahiner2021vectoroutput}. These problems are not tractable in polynomial time in the worst case. For instance taking $\data=\vec{I}_n$ simplifies to a copositive program, which is NP-hard for arbitrary $\vec{Y}$.

\subsubsection{$\ell_1$ regularized version of vector output case} 
Notice that since \eqref{eq:extreme_vectoroutput} is a non-convex problem, finding extreme points in general is computationally expensive especially when the data dimensionality is high. Therefore, in this section, we provide an $\ell_1$ regularized version of the problem in \eqref{eq:problem_def_multiclass} so that extreme points can be efficiently achieved using convex optimization tools. Consider the following optimization problem
\begin{align*} 
       &\min_{\theta \in \Theta } \| \firstwmat\|_F^2 +\sum_{j=1}^m \|\secondwvec_j\|_1^2  \text{ s.t. } \relu{\data \firstwmat} \secondwmat=\vec{Y}. 
\end{align*}
Then using the scaling trick in Lemma \ref{lemma:reg_equivalence}, we obtain the following
\begin{align*} 
       &\min_{\theta \in \Theta } \sum_{j=1}^m \|\secondwvec_j\|_1  \text{ s.t. } \relu{\data \firstwmat} \secondwmat=\vec{Y}, \; \| \firstw_j \|_2^2 \le 1, \; \forall j,
\end{align*}
which has the following dual form
\begin{align*}
    &\max_{\vec{V} } \trace(\vec{V}^T \vec{Y}) \text{ s.t. } \|\vec{V}^T \relu{\data \firstw} \|_{\infty} \le 1\,,\forall \firstw\in \ball_2 
\end{align*}
and an optimal $\firstwmat$ satisfies
\begin{align*}
    \|(\data \firstwmat^*)_{+}^T \vec{V}^* \|_{\infty} = 1\,,
\end{align*} 
where $\vec{V}^*$ is the optimal dual variable. Note that we use this particular form as it admits a simpler solution with the cutting-plane method. We again relax the problem using the spike-free relaxation and then we solve the following problem for each $k \in [o]$
\begin{align*}
\begin{split}
    &\hat{\firstw}_{k,1}=\argmax_{\firstw} \dual_k^T \data \firstw \text{ s.t. } \data \firstw  \succcurlyeq \vec{0}, \| \firstw \|_2 \leq 1 \\ 
    &  \hat{\firstw}_{k,2}=\argmin_{\firstw} \dual_k^T \data \firstw \text{ s.t. }\data \firstw \succcurlyeq \vec{0}, \| \firstw \|_2 \leq 1,
    \end{split} 
\end{align*}
where $\dual_k$ is the $k^{th}$ column of $\vec{V}$. After solving these optimization problems, we select the two neurons that achieve the maximum and minimum objective value among $o$ neurons for each problem.  Thus, we can find the weights for the hidden layers using convex optimization.

Consider the minimal cardinality problem
\begin{align*}
    \min_{\theta \in \Theta} \| \secondwmat \|_0 \text{ s.t. } f_{\theta}(\data)=\vec{Y}, \| \firstw_j \|_2=1 , \forall j .
\end{align*}
The following result provides a characterization of the optimal solutions to the above problem.

\begin{lems}\label{lemma:l1_multiclass}
Suppose that $n\le d$, $\data$ is full row rank, and $\vec{Y} \in \mathbb{R}^{n \times o}_+$, e.g., one hot encoded outputs for multiclass classification and we have at least one sample in each class. Then an optimal solution to \eqref{eq:problem_Lzero} is given by
\begin{align*}
    \firstw_k^* &= \frac{\data^\dagger \relu{\vec{y}_k}}{\|\data^\dagger \relu{\vec{y}_k}\|_2}\mbox{ and } \secondwvec_k^* = \|\data^\dagger \relu{\vec{y}_k}\|_2 \vec{e}_k
\end{align*}
for each $k \in [o]$, where $\secondwvec_k$ and $\vec{y}_k$ are the $k^{th}$ row of $\secondwmat$ and column of $\vec{Y}$, respectively.
\end{lems}

\begin{lems}\label{lemma:l1_l0_multiclass}
We have $\ell_1$-$\ell_0$ equivalence if the following condition holds
\begin{align*}
    \min_{\vec{v}:\vec{v}^T \relu{\data\firstw_k}=1,\forall k}~\max_{\firstw:\firstw\in\ball_2} \vec{v}^T\relu{\data\firstw} \le 1\,.
\end{align*}
\end{lems}

\section{Comparison with Neural Tangent Kernel}\label{sec:main_ntk}
Here, we first briefly discuss the recently introduced NTK \citep{ntk_jacot} and other connections to kernel methods. We then compare this approach with our exact characterization in terms of a kernel matrix in Corollary \ref{cor:kernel_squareloss}.
\begin{figure*}[t]
	\centering
	\captionsetup[subfigure]{oneside,margin={1cm,0cm}}
		\begin{subfigure}[t]{0.32\textwidth}
		\centering
		\includegraphics[width=1.2\textwidth, height=0.8\textwidth]{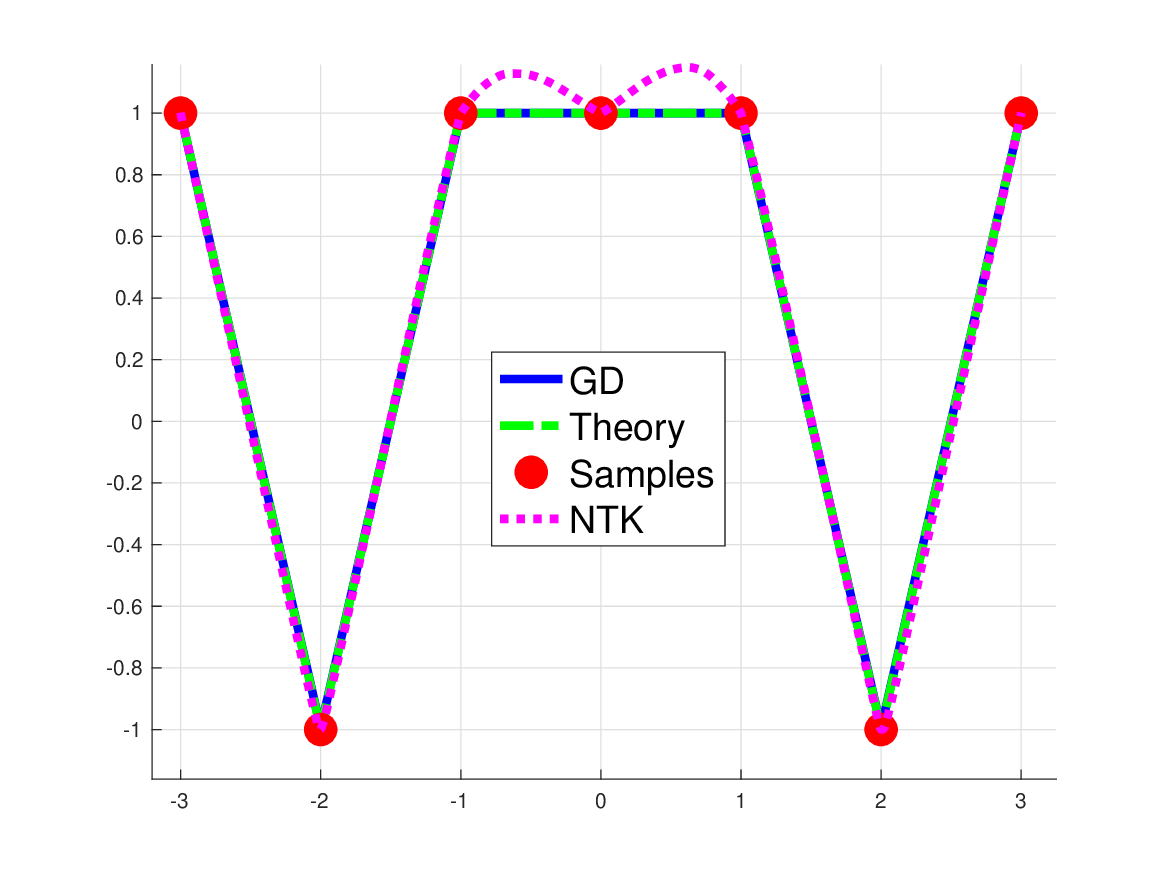}
		\caption{ \centering} \label{fig:ntk1}
	\end{subfigure} \hspace*{\fill}
			\begin{subfigure}[t]{0.32\textwidth}
		\centering
		\includegraphics[width=1.2\textwidth, height=0.8\textwidth]{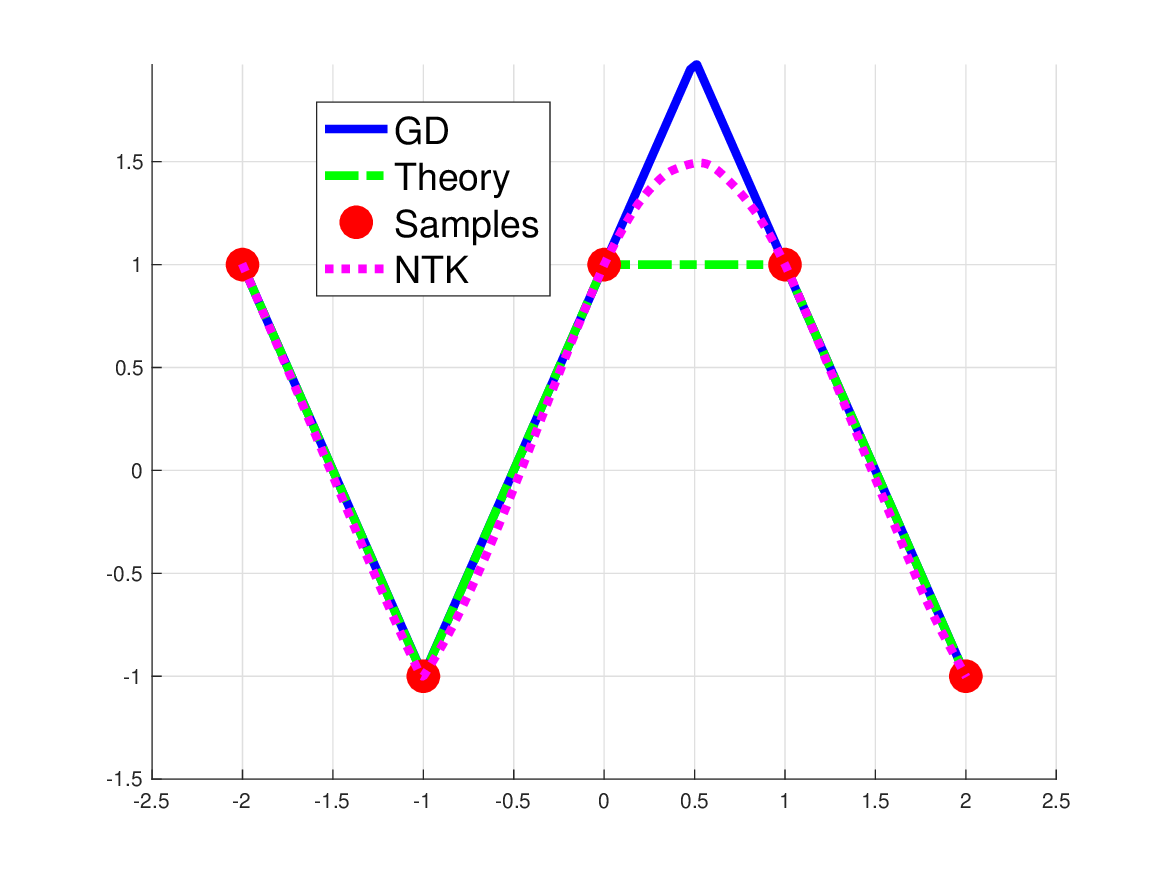}
		\caption{\centering} \label{fig:ntk2}
	\end{subfigure} \hspace*{\fill}
		\begin{subfigure}[t]{0.32\textwidth}
		\centering
		\includegraphics[width=1.2\textwidth, height=0.8\textwidth]{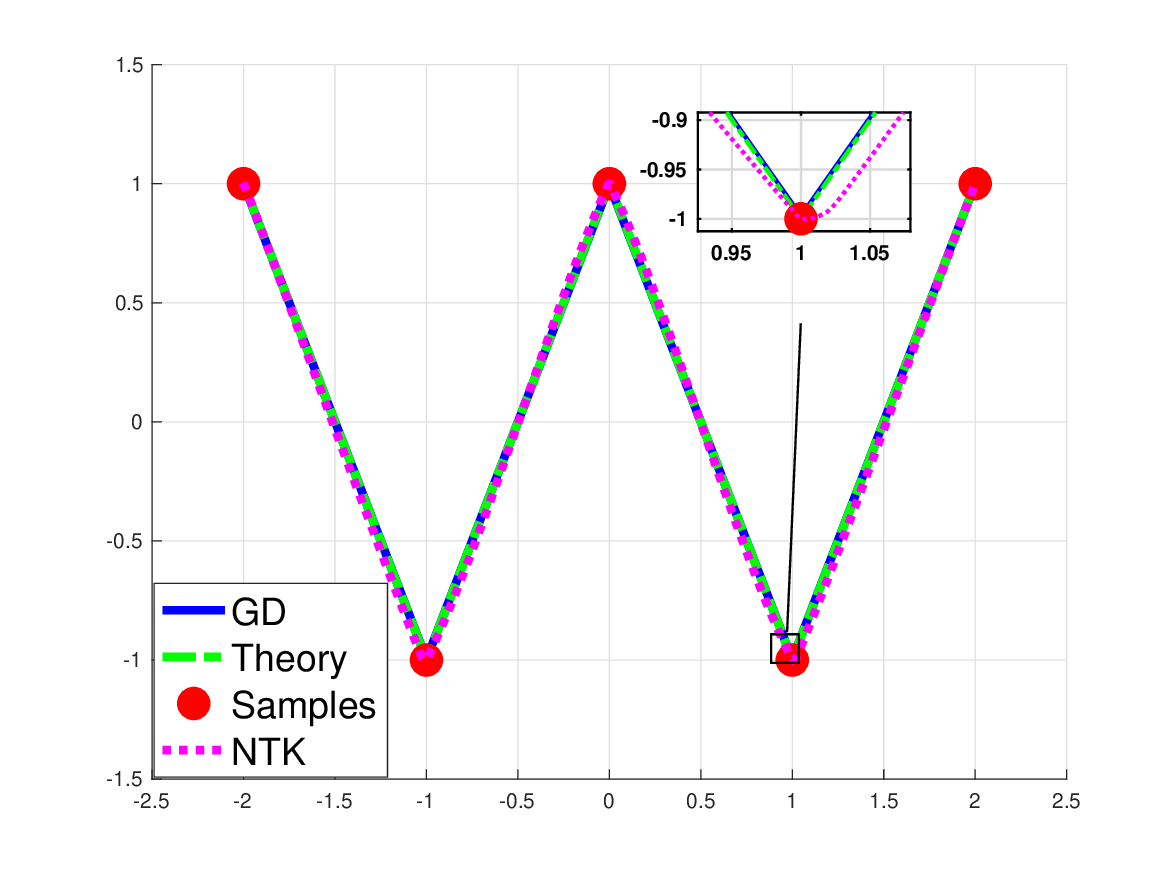}
		\caption{\centering} \label{fig:ntk3}
	\end{subfigure}\hspace*{\fill}
	\medskip
	\caption{One dimensional regression task with square loss, where we apply GD, NTK, and our approach in Corollary \ref{cor:kernel_squareloss}.   }
\label{fig:ntk_all}
	\end{figure*}
	
The connection between kernel methods and infinitely wide NNs has been extensively studied \citep{neal_priorinfinite,gaussian_infinite,gaussian_infinite2}. Earlier studies typically considered untrained networks, or the training of the last layer while keeping the hidden layers fixed and random. Then, by assuming a distribution for initialization of the parameters, the behavior of an infinitely wide NN can be captured by a kernel matrix $\vec{K}(\sample_i,\sample_j)=\mathbb{E}_{\theta \sim \mathcal{D}}[f_{\theta}(\sample_i)f_{\theta}(\sample_j)]$, where $\mathcal{D}$ is the distribution for initialization. However, these  results do not fully align with practical NNs where all the layers are trained simultaneously. Therefore, \citet{ntk_jacot} introduced a new kernel method, i.e., NTK, where all the layers are trained while the width tends to infinity. In this scenario, the network can be characterized by the kernel matrix $\vec{K}(\sample_i,\sample_j)=\mathbb{E}_{\theta \sim \mathcal{D}}[\nabla_{\theta} f_{\theta}(\sample_i)^T\nabla_{\theta}f_{\theta}(\sample_j)]$. This can be interpreted as a linearization of the NN model under a particular scaling assumption, and is closely related to random feature methods. We refer the reader to \citet{lazy_training_bach} for details and limitations of the NTK framework. For one dimensional problems, this kernel characterization can be written as follows \citep{ntk_relu_1d}\footnote{We provide the NTK formulation without a bias term to simplify the presentation. The expression for a case with bias can be found in \citet{ntk_1d}.}
\begin{align*}
    \vec{K}(\samplescalar_i,\samplescalar_j)= |\samplescalar_i||\samplescalar_j| \kappa\left( \frac{\samplescalar_i\samplescalar_j}{|\samplescalar_i||\samplescalar_j|}\right),
\end{align*}
where 
\begin{gather*}
    \kappa(u)= u \kappa_0(u)+\kappa_1(u)\\
    \kappa_0(u)=\frac{1}{\pi}(\pi-\arccos(u)) \; \kappa_1(u)=\frac{1}{\pi}\left(u(\pi-\arccos(u)) +\sqrt{1-u^2}\right).
\end{gather*}
After forming the kernel matrix, one can solve the following $\ell_2$-norm minimization problem to obtain the last layer weights
\begin{align}\label{eq:problemdef_ntk}
    \min \| \secondwvec\|_2^2 \text{ s.t. } \vec{K}\secondwvec=\vec{y}.
\end{align}
We first note that our approach in \eqref{eq:problemdef_kernel} is different than the NTK approach in \eqref{eq:problemdef_ntk} in terms of kernel construction and objective function.

In order to compare the performance of \eqref{eq:problemdef_kernel}, \eqref{eq:problemdef_ntk} and GD, we perform experiments on one dimensional datasets. In Figure \ref{fig:ntk_all}, we observe that NTK outputs smoother functions compared to GD and our approach. Particularly, in Figure \ref{fig:ntk1} and \ref{fig:ntk2}, we clearly see that the output of NTK is not a piecewise linear function unlike our approach and GD. Moreover, even though the output of NTK looks like a piecewise linear function in Figure \ref{fig:ntk3}, we again observe its smooth behavior around the data points. Thus, we conclude that NTK yields output functions that are significantly different than the piecewise linear functions obtained by GD and our approach. We also note that optimal solution might not be unique as proven in Proposition \ref{prop:1d_uniqueness} and Figure \ref{fig:uniquness}, which explains why GD converges to a solution with a kink in the middle of two data points while the kinks obtained by our approach exactly aligns with the data points.

\section{Numerical Experiments}\label{sec:numerical_exps}
\begin{figure*}[t]
	\centering
	\captionsetup[subfigure]{oneside,margin={1cm,0cm}}
		\begin{subfigure}[t]{0.49\textwidth}
		\centering
		\includegraphics[width=1\textwidth, height=0.8\textwidth]{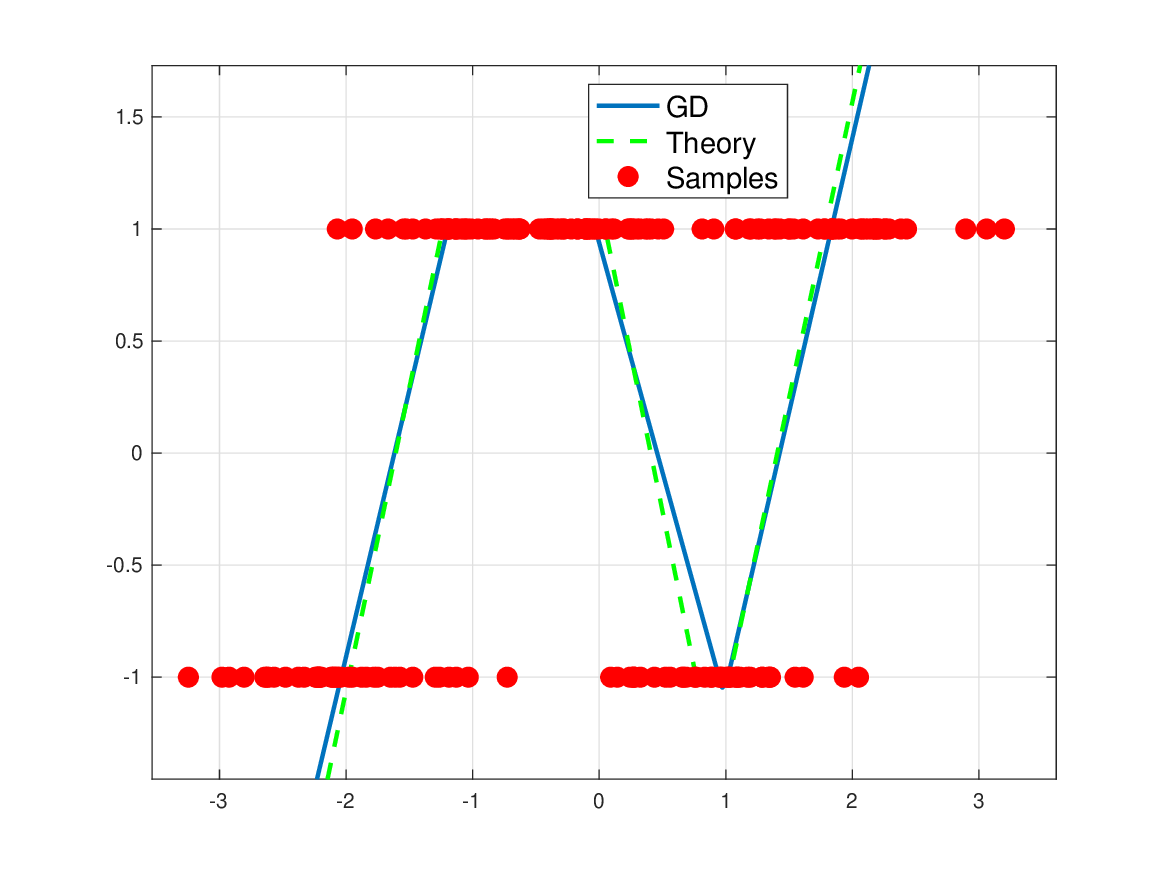}
		\caption{Visualization of the samples and the function fit.\centering} \label{fig:hinge_ex_data2}
	\end{subfigure} \hspace*{\fill}
			\begin{subfigure}[t]{0.49\textwidth}
		\centering
		\includegraphics[width=1\textwidth, height=0.8\textwidth]{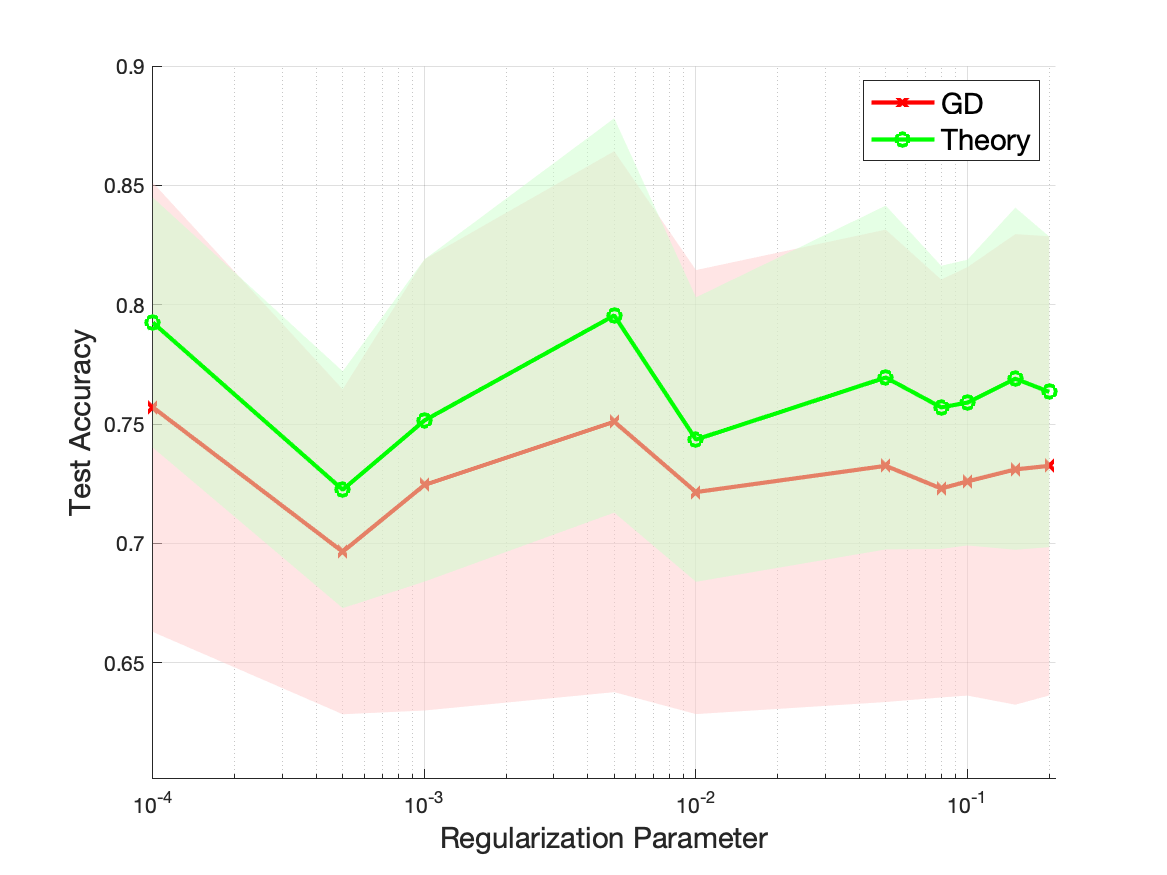}
		\caption{Test accuracy with $90\%$ confidence band.\centering} \label{fig:hinge_ex_test}
	\end{subfigure} \hspace*{\fill}

	\medskip
	\caption{Binary classification using hinge loss, where we apply GD and our approach in Theorem \ref{theo:hinge_svm}, i.e., denoted as Theory.   }
\label{fig:hinge_exp}
	\end{figure*}

	\begin{figure*}[h]
        \centering
        \begin{subfigure}[b]{0.45\textwidth}
            \centering
            \includegraphics[width=\textwidth]{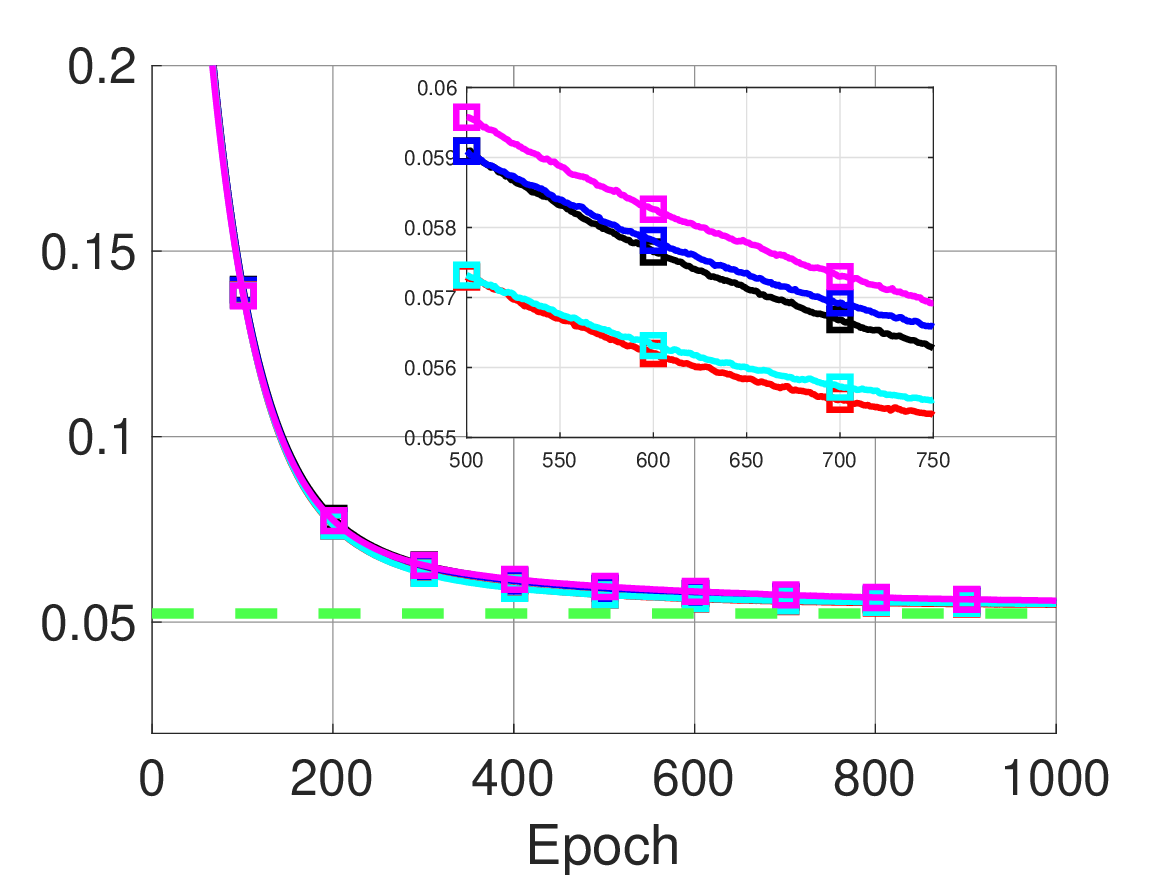}
            \caption{MNIST-Training objective}
        \end{subfigure}
        \hfill
        \begin{subfigure}[b]{0.45\textwidth}   
            \centering 
            \includegraphics[width=\textwidth]{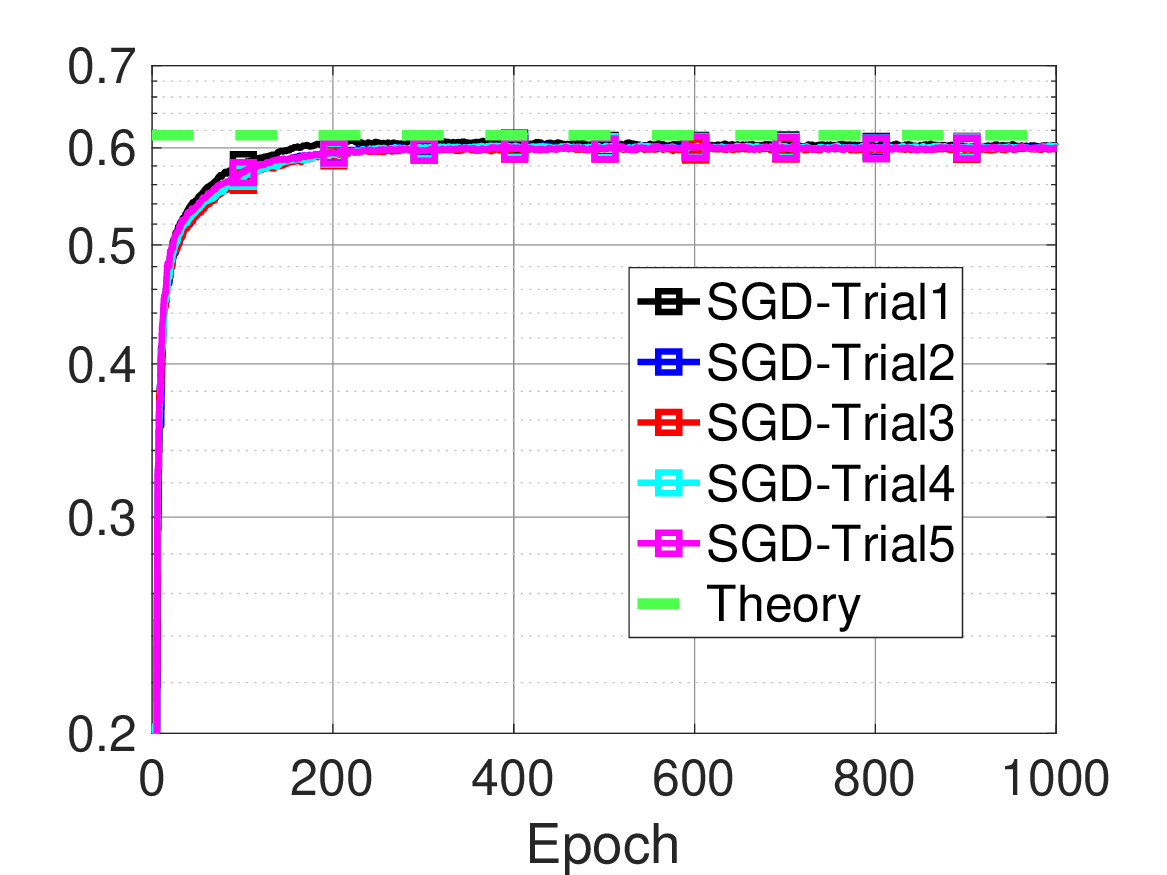}
            \caption{MNIST-Test accuracy}
        \end{subfigure}
	\caption{Training and test performance of 5 independent SGD trials on whitened and sampled MNIST, where $(n,d)=(200,250)$, $K=10$, $\beta=10^{-3}$, $m=100$ and we use squared loss with one hot encoding. For the method denoted as Theory, we use the layer weights in Theorem \ref{theo:closedform_regularized_multiclass}. }\label{fig:mnist}
    \end{figure*}
\begin{figure*}[t]
        \centering
        \begin{subfigure}[b]{0.45\textwidth}
            \centering
            \includegraphics[width=\textwidth]{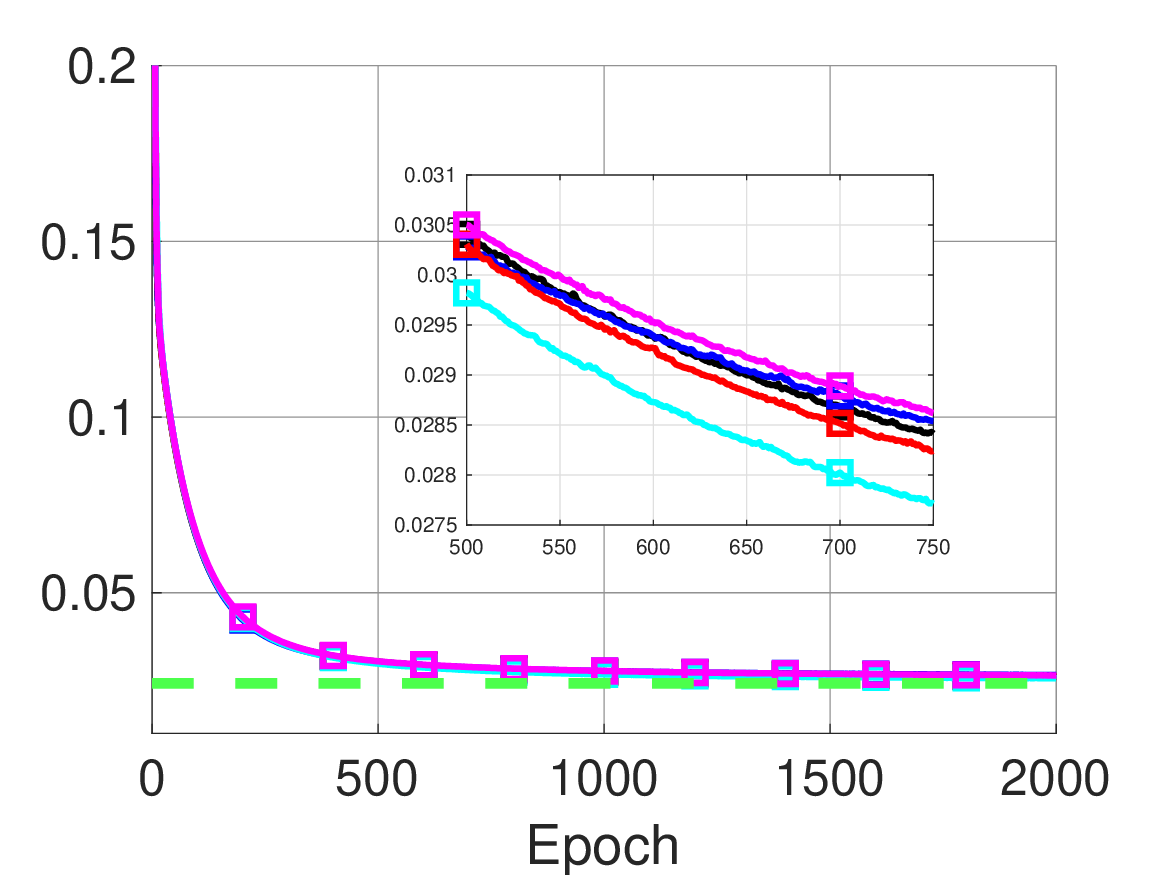}
            \caption{CIFAR10-Training objective}
        \end{subfigure}
        \hfill
        \begin{subfigure}[b]{0.45\textwidth}   
            \centering 
            \includegraphics[width=\textwidth]{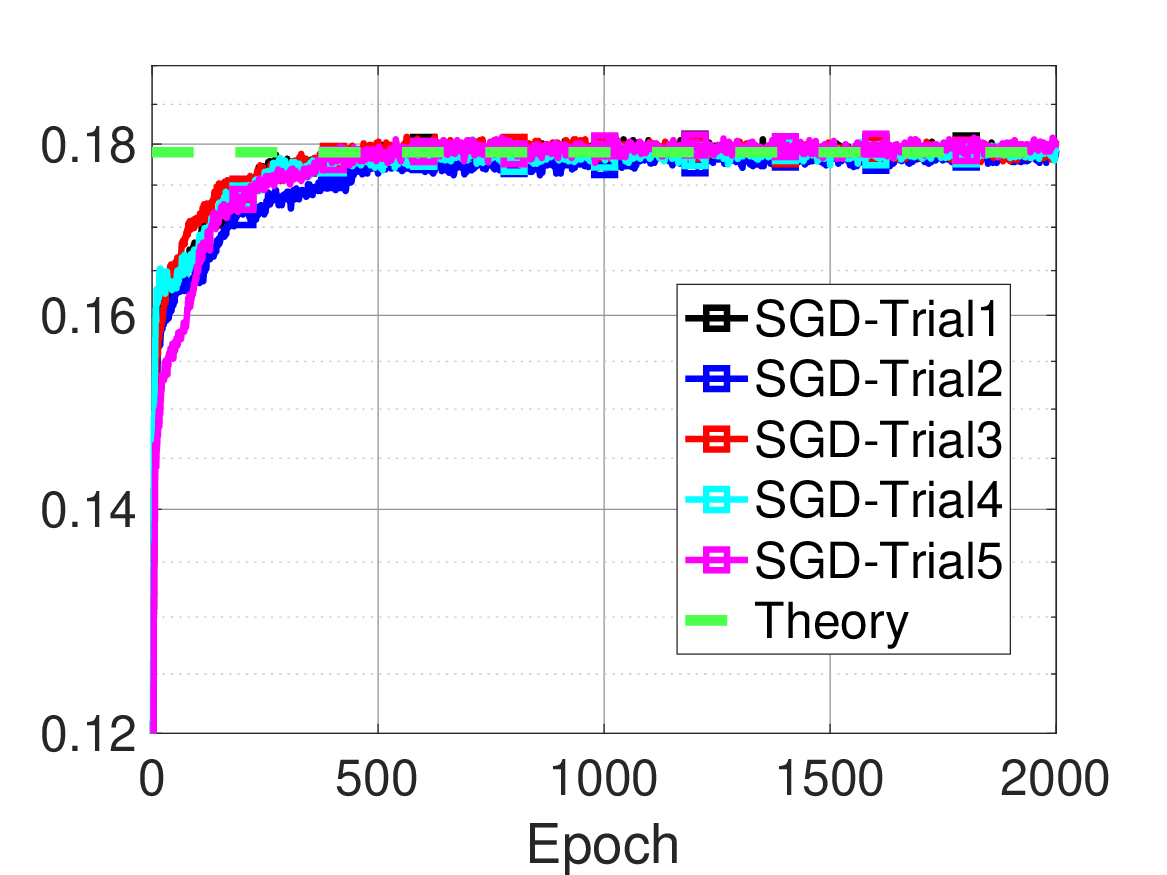}
            \caption{CIFAR10-Test accuracy}
        \end{subfigure}
	\caption{Training and test performance of 5 independent SGD trials on whitened and sampled CIFAR-10, where $(n,d)=(60,60)$, $K=10$, $\beta=10^{-3}$, $m=100$ and we use squared loss with one hot encoding. For Theory, we use the layer weights in Theorem \ref{theo:closedform_regularized_multiclass}. }\label{fig:cifar}
   \end{figure*}
	
We first consider a binary classification experiment using hinge loss on a synthetic dataset\footnote{We provide further details on the numerical experiments in Appendix \ref{sec:additional_experiments}.}. To generate a dataset, we use a Gaussian mixture model, i.e., $\samplescalar_i \sim \mathcal{N} (\mu_j,0.25)$, where the labels are computed using: $y_i=1$, if  $\mu_j \in \{-1,0,2\}$, and $y_i=-1$, if $\mu_j \in \{-2,1\}$. Following these steps, we generate multiple datasets with nonoverlapping training and test splits. We then run our approach in Theorem \ref{theo:hinge_svm}, i.e., denoted as Theory and GD on these datasets. In Figure \ref{fig:hinge_exp}, we plot the mean test accuracy (solid lines) of each algorithm along with a one standard deviation confidence band (shaded regions). As illustrated in this example, our approach achieves slightly better generalization performance compared to GD. We also visualize the sample data distributions and the corresponding function fits in Figure \ref{fig:hinge_ex_data2}, where we provide an example to show the agreement between the solutions found by our approach and GD. 
\begin{table*}
    \caption{Classification Accuracies ($\%$) and test errors \medskip}
  \label{tab:classification}
  \centering
  \resizebox{1\columnwidth}{0.08\columnwidth}{
  \begin{tabular}{lllllllllll}
    \toprule
    \cmidrule(r){1-9}
         & MNIST     & CIFAR-10  & Bank     & Boston & California & Elevators & News20  & Stock\\
    \midrule
    \texttt{One Layer NN (Least Squares)}  &86.04$\%$ &    36.39$\%$ & 0.9258  & 0.3490& 0.8158   &0.5793& 1.0000 & 1.0697  \\
    \texttt{Two-Layer NN (Backpropagation)}  & 96.25$\%$  &41.57 $\%$ &0.6440 &0.1612  & 0.8101 &0.4021&  0.8304& 0.8684       \\
    \texttt{Two-Layer NN Convex}   & 96.94$\%$ &  42.16$\%$  &\textbf{0.5534}       &\textbf{0.1492}& \textbf{0.6344} &\textbf{0.3757}& \textbf{0.8043}& \textbf{0.6184}    \\
    \texttt{Two-Layer Convex-RF} &\textbf{97.72}$\%$&\textbf{80.28$\%$}& - &-&-&-&-&-\\
    \bottomrule
  \end{tabular}}
  \end{table*}

\begin{figure*}[t]
	\centering
	\captionsetup[subfigure]{oneside,margin={1cm,0cm}}
		\centering
		\includegraphics[width=0.8\textwidth, height=0.1\textwidth]{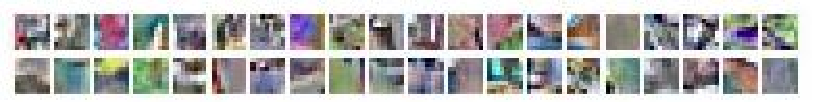}
		\caption{Extreme points found by \eqref{eq:mixture_outputs} applied on image patches of CIFAR-10 yield filters used in the \texttt{Convex-RF} algorithm. Note that the extreme points visually correspond to predictive image patches.} \label{fig:filters}
	\end{figure*}

We then consider classification tasks and report the performance of the algorithms on MNIST \citep{mnist} and CIFAR-10 \citep{cifar10}. In order to verify our results in Theorem \ref{theo:closedform_regularized_multiclass}, we run 5 SGD trials with independent initializations for the network parameters, where we use subsampled versions of the datasets. As illustrated in Figure \ref{fig:mnist} and \ref{fig:cifar}, the network constructed using the closed-form solution achieves the lowest training objective and highest test accuracy for both datasets. In addition to our closed-form solutions, we also propose a convex cutting plane based approach to optimize ReLU networks. In Table \ref{tab:classification}, we observe that our approach denoted as \texttt{Convex}, which is completely based on convex optimization, outperforms the non-convex backpropagation based approach. Note that we use the full datasets for this experiment. Furthermore, we use an alternative approach, denoted as \texttt{Convex-RF} in Table \ref{tab:classification} which uses \eqref{eq:mixture_outputs} on image patches to obtain 40k filters, e.g., Figure \ref{fig:filters} (further details are provided in the next section). This training approach for the hidden layer weights surprisingly increases the accuracy by almost $40 \%$ compared to the convex approach with the cutting plane algorithm. We also evaluate the performances on several regression datasets, namely Bank, Boston Housing, California Housing, Elevators, Stock \citep{regression}, and the Twenty Newsgroups text classification dataset \citep{mitchell1997mcgraw}. In Table \ref{tab:classification}, we provide the test errors for each approach. Here, our convex approach outperforms the backpropagation, and the one layer NN in each case.
\subsection{Unsupervised and interpretable training using extreme points: \texttt{Convex-RF}}
\begin{algorithm}[!h]
		\caption{\texttt{Convex-RF}}
		\begin{algorithmic}[1]
        \State Set $P$, $\epsilon$, and $\beta$
        \State Randomly select $P$ patches from the dataset: $\{\vec{p}_i\}_{i=1}^P$
        \For{i=1:P}
        \State Normalize the patch: $\vec{\bar{p}}_i=\frac{\vec{p}_i-\text{mean}(\vec{p}_i)}{\sqrt{\text{var}(\vec{p}_i)+\epsilon}}$
        \EndFor
        \State Form a patch matrix $\vec{P}=[\vec{\bar{p}}_1 \, \ldots\, \vec{\bar{p}}_P]$
        \State{(Optional) Apply whitening to the patch matrix: \begin{align*}
        &[\vec{V},\vec{D}]=\text{eig}(\text{cov}(\vec{P})) \\
        &\vec{\tilde{P}}= \vec{V}(\vec{D}+\epsilon \vec{I})^{-\frac{1}{2}}\vec{V}^T \vec{P}\end{align*}}
        \For{i=1:P}
        \State Compute a neuron using \eqref{eq:mixture_outputs}: 
        \begin{align*}\firstw_i=\frac{\vec{\tilde{p}}_{i} -\sum_{\substack{j=1\\ j \neq i}}^n \lambda_j\vec{\tilde{p}}_{j}}{\bigg\|\vec{\tilde{p}}_{i} -\sum_{\substack{j=1\\ j \neq i}}^n \lambda_j\vec{\tilde{p}}_{j}\bigg\|_2}\end{align*}
         where $\boldsymbol{\lambda}$ is computed via the following problem
        \begin{align*}
            \min_{\boldsymbol{\lambda}}  \bigg\|\vec{\tilde{p}}_i -\sum_{\substack{j=1\\ j \neq i}}^n \lambda_j\vec{\tilde{p}}_j\bigg\|_2  \text{ s.t. } \boldsymbol{\lambda} \succcurlyeq \vec{0},\vec{1}^T \boldsymbol{\lambda}=1.
        \end{align*}
        \EndFor
        \State{Form the neuron matrix: $\firstwmat=[\firstw_1\, \ldots \firstw_P]$}
        \State{Extract all the patches in $\data$: $\data_p$}
        \State{Compute activations: $\vec{B}=\text{pooling}(\text{ReLU}(\data_p \firstwmat))$}
        \State{Solve a convex $\ell_1$-norm minimization problem: $
            \min_{\secondwvec}  \frac{1}{2}\|\vec{B}\secondwvec-\vec{y}\|_2^2+\beta\|\secondwvec \|_1  
        $}
		\end{algorithmic}\label{alg:convex_rf}	
	\end{algorithm}
For this approach, we use the convolutional neural net architecture in \citet{kmenas_andrewng}. However, instead of using random filters as in \citet{understanding_zhang} or applying the k-means algorithm as in \citet{kmenas_andrewng}, we use the filters that are extracted from the patches using our convex approach in \eqref{eq:mixture_outputs}. Particularly, we first randomly obtain patches from the dataset. We then normalize and whiten (using the ZCA whitening approach) the randomly selected patches. After that we apply \eqref{eq:mixture_outputs} on the the resulting patches to obtain the filter weights via a convex optimization problem with unit simplex constraints.

After obtaining the filter weights in an unsupervised manner, we first compute the activations for each input patch. Here, we use a linear function for activations unlike the triangular activation function in \citet{kmenas_andrewng}. We then apply ReLU and max pooling. Finally, we solve a convex $\ell_1$-norm minimization problem to obtain the output layer weights. Therefore, we achieve a training approach that completely utilizes convex optimization tools and learns the hidden layer weights in an unsupervised manner. The complete algorithm is also presented in Algorithm \ref{alg:convex_rf}.

\begin{remark}
Note that the ZCA whitening step plays a crucial role for similar implementations in \cite{kmenas_andrewng,shankar_kernels,thiry2021unreasonable}. We believe that this is mainly due to the fact that these approaches rely on either direct distance computations or inner products with the samples, where whitening can enable more informative features, i.e., either the distance computation or inner product, by transforming the samples (see Appendix A of \cite{thiry2021unreasonable}). However, in our case, the hidden neurons constructed from the data samples are inherently informative and normalized as detailed in Step 9 of Algorithm \ref{alg:convex_rf}. Particularly, since we compute the distance of each sample to the convex hull of the remaining samples instead of computing pairwise distances, our approach is much more robust against the correlations among patches. In addition, due to the scaling in Lemma \ref{lemma:reg_equivalence}, the hidden neurons are normalized such that inner products tend to be drastically less dependent on the correlations. Therefore, whitening don't change our results and this is why we remark that the whitening step is optional in Algorithm \ref{alg:convex_rf}.
\end{remark}

\section{Concluding Remarks}\label{sec:main_conclusion}
We studied two-layer ReLU networks and introduced a convex analytic framework based on duality to  characterize a set of optimal solutions to the regularized training problem. Our analysis showed that optimal solutions can be exactly characterized as the extreme points of a convex set. More importantly, these extreme points yield simple structures at the network output, which explains why ReLU networks fit structured functions, e.g., a linear spline interpolation for one dimensional datasets. Moreover, by establishing a relation with minimum cardinality problems in compressed sensing, we even provided closed-form solutions for the optimal hidden layer weights in various practically relevant scenarios such as problems involving rank-one or spike-free data matrices. Therefore, for such cases, one can directly use these closed-form solutions and then train only the output layer, i.e., a linear layer, in a similar fashion to kernel regression/classification problems. However, unlike the existing kernel methods, e.g., NTK \citep{ntk_jacot}, our approach is regularized by $\ell_1$-norm encouraging sparse solutions. Thus, we are able to match the performance of a classical finite width ReLU network trained with SGD, for which existing kernel methods fail to provide a satisfactory approximation, by solving a linear minimum $\ell_1$-norm problem with a fixed matrix. Additionally, we provided an iterative algorithm based on the cutting plane method to optimize the network parameters for problems with arbitrary data and then proved its convergence to the global optimum under certain assumptions.

In the light of our results, there are multiple future research directions, which we want to mention as open research problems. First of all, our analysis reveals that the original non-convex training problem has a geometrical structure that can be fully characterized by convex duality. Under certain technical conditions such as spike-freeness, whitened, or rank-one matrices, this geometry is easily understood by closed-form expressions of the extreme points. We conjecture that one can also utilize convex duality to understand the optimization landscape and extraordinary generalization abilities of deep networks. For instance, the recent findings in \citet{lacotte2020all,lederer2020spurious} regarding the global optimality of all local minima in sufficiently wide networks can be understood through the lens of convex analysis. In addition to this, our approach explains why NTK \citep{ntk_jacot} and other kernel methods fail to recover the exact training dynamics of finite width ReLU networks. We also show that one can obtain closed-form solutions for all network parameters, i.e, hidden and output layer weights, in some special cases such as problems with rank-one or whitened data matrices. Hence, there is no need to train a fully connected two-layer ReLU network via SGD in these cases. Based on these observations, we believe that the active learning regime as coined in \citet{lazy_training_bach}, or a new transition regime, where hidden neurons actively learn useful features that generalize well can be analytically characterized. Finally, one can also extend our polynomial-time convex formulations for spike-free problems to develop efficient solvers to globally optimize deeper networks via layerwise learning. 
Even though certain parts of our analysis are restricted certain classes of data matrices such as spike-free, whitened, rank-one or one dimensional, we conjecture that a similar convex analytic framework can be developed for arbitrary data distributions, alternative network architectures and deeper networks.
After a preliminary version of this manuscript appeared in \citet{ergen2020aistats}, our follow-up work \citep{ergen2020convexdeep,pilanci2020convex,ergen2020cnn,sahiner2021hidden} aimed to address some of these questions via the convex analytic tools developed in this present work.


\acks{This work was partially supported by the National Science Foundation
under grants IIS-1838179, ECCS-2037304 and the Army Research Office.}

\newpage

\addcontentsline{toc}{section}{Appendix} 
\part{Appendix} 
\parttoc 

In this section, we present proofs of the main results and further details on the algorithms and numerical results.

\section{Additional details on the numerical experiments}\label{sec:additional_experiments}
In this section, we provide further information about our experimental setup.

In the main paper, we evaluate the performance of the introduced approach on several real datasets. For comparison, we also include the performance of a two-layer NN trained with the backpropagation algorithm and the well-known linear least squares approach. For all the experiments, we use the regularization term (also known as weight decay) to let the algorithms generalize well on unseen data \citep{weightdecay_krogh}. In addition to this, we use the cutting plane based algorithm along with the neurons in \eqref{eq:mixture_outputs} for our convex approach. In order to solve the convex optimization problems in our approach, we use CVX \citep{cvx}. However, notice that when dealing with large datasets, e.g., CIFAR-10, plain CVX solvers might need significant amount of memory. In order to circumvent these issues, we use SPGL1 \citep{spgl1} and SuperSCS \citep{superscs} for large datasets. We also remark that all the datasets we use are publicly available and further information, e.g., training and test sizes, can be obtained through the provided references \citep{mnist,cifar10,regression,news20}. Furthermore, we use the same number of hidden neurons for both our approach and the conventional backpropagation based approach to have a fair comparison.

In order to gain further understanding of the connection between implicit regularization and initial standard deviation of the neuron weights, we perform an experiment that is presented in the main paper, i.e., Figure \ref{figs_intro}. In this experiment, using the backpropagation algorithm, we train two-layers NNs with different initial standard deviations such that each network completely fits the training data. Then, we find the maximum absolute difference between the function fit by the NNs and the ground truth linear interpolation. After averaging our results over many random trials, we obtain Figure \ref{figs_intro}. The same settings are also used for the experiment using hinge loss.

\section{Cutting plane algorithm with a bias term}\label{sec:cuttingplane_bias}
Here, we include the cutting plane algorithm which accommodates a bias term. This is slightly more involved than the case with no bias because of extra constraints. We have the corresponding dual problem as in Theorem \ref{theo:equality_dual} 
\begin{align}\label{eq:dual_problem_withbias}
    &\max_{\vec{v}:\vec{1}^T\vec{v}=0} \vec{v}^T \vec{y} \text{ s.t. } \big\vert\vec{v}^T \relu{\data \firstw+\bias\vec{1}} \big \vert \le 1\,,\forall \firstw\in \ball_2, \forall b \in \reals
\end{align}
and an optimal ($\firstwmat^*$, $\biasvec^*$) satisfies
\begin{align*}
    \|(\data \firstwmat^*+\vec{1}\biasvec{*^T})_{+}^T \dual^* \|_{\infty} = 1\,,
\end{align*} 
where $\dual^*$ is the optimal dual variable.

Among infinitely many possible unit norm weights, we need to find the weights that violate the inequality constraint in the dual form, which can be done by solving the following optimization problems
\begin{align*}
\begin{split}
    &\firstw_1^*=\argmax_{\firstw,\bias} \dual^T (\data \firstw +\bias\vec{1})_+ \text{ s.t. } \| \firstw \|_2 \leq 1 \\ 
    &  \firstw_2^* =\argmin_{\firstw,\bias} \dual^T (\data \firstw +\bias\vec{1})_+ \text{ s.t. } \| \firstw \|_2 \leq 1.
    \end{split}
\end{align*}
 However, the above problem is not convex since ReLU is a convex function. In this case, we can further relax the problem by applying the spike-free relaxation as follows 
\begin{align*}
\begin{split}
    &(\hat{\firstw}_1,\hat{\bias}_1)=\argmax_{\firstw,\bias} \dual^T \data \firstw +\bias \dual^T \vec{1}\text{ s.t. } \data \firstw +\bias \vec{1} \succcurlyeq \vec{0}, \| \firstw \|_2 \leq 1 \\ 
    &  (\hat{\firstw}_2,\hat{\bias}_2)=\argmin_{\firstw,\bias} \dual^T \data \firstw +\bias \dual^T \vec{1} \text{ s.t. }\data \firstw+\bias \vec{1} \succcurlyeq \vec{0}, \| \firstw \|_2 \leq 1,
    \end{split} 
\end{align*}
where we relax the set $\{(\data \firstw+\bias \vec{1})_+| \firstw \in \mathbb{R}^d , \| \firstw\|_2 \leq 1\} $ as $\{\data \firstw+\bias \vec{1}  | \firstw \in \mathbb{R}^d, \| \firstw\|_2 \leq 1 \} \cap \mathbb{R}_+^n$. Now, we can find the weights and biases for the hidden layer using convex optimization. However, notice that depending on the sign of $\vec{1}^T\dual$ one of the problems will be unbounded. Thus, if $\vec{1}^T\dual \neq 0$, then we can always find a violating constraint, which will make the problem infeasible. However, note that we do not include a bias term for the output layer. If we include the output bias term, then  $\vec{1}^T\dual=0$ will be implicitly enforced via the dual problem.

Based on our analysis, we propose the following convex optimization approach to train the two-layer NN. We first find a violating neuron. After adding these parameters to $\firstwmat$ as a column and to $\biasvec$ as a row, we try to solve the original problem. If we cannot find a new violating neuron then we terminate the algorithm. Otherwise, we find the dual parameter for the updated $\firstwmat$. We repeat this procedure until  the optimality conditions are satisfied (see Algorithm \ref{alg:pseudo1_withbias} for the pseudocode). Since the constraint is bounded below and $\hat{\firstw}_j$'s are bounded, Algorithm \ref{alg:pseudo1_withbias} is guaranteed to converge in finitely many iterations Theorem 11.2 of \citet{semiinfinite_goberna}.
\begin{algorithm}[!h]
	\caption{Cutting Plane based Training Algorithm for Two-Layer NNs (with bias)}
	\begin{algorithmic}[1]
    \State Initialize $\dual$ such that $\vec{1}^T \dual=0$
    \While{there exists a violating neuron}
    \State Find $\hat{\firstw}_1$, $\hat{\firstw}_2$, $\hat{\bias}_1$ and $\hat{\bias}_2$
    \State $\firstwmat \leftarrow [\firstwmat \text{ } \hat{\firstw}_1\text{ } \hat{\firstw}_2]$ 
    \State $\biasvec\leftarrow [\biasvec^T \text{ }\hat{\bias}_1 \text{ }\hat{\bias}_2]^T$
    \State{Find $\dual$ using the dual problem}
    \State Check the existence of a violating neuron 
    \EndWhile
    \State Solve the problem using $\firstwmat$ and $\biasvec$
    \State Return $\theta=(\firstwmat,\biasvec,\secondwvec)$
	\end{algorithmic}\label{alg:pseudo1_withbias}	
\end{algorithm}

\section{Infinite size neural networks}
\label{sec:infsize}
Here we briefly review infinite size, i.e., infinite width, two-layer NNs \citep{bach2017breaking}. We refer the reader to \citet{bengio2006convex,margin_theory_tengyu} for further background and connections to our work. Consider an arbitrary measurable input space $\mathcal{X}$ with a set of continuous basis functions $\phi_{\vec{u}}:\mathcal{X}\rightarrow \reals$ parameterized by $\vec{u}\in \ball_2$. We then consider real-valued Radon measures equipped with the uniform norm \citep{Rudin}. For a signed Radon measure $\boldsymbol{\mu}$, we define the infinite size NN output for the input $\vec{x}\in\mathcal{X}$ as
\begin{align*}
    f(\vec{x}) = \int_{\vec{u}\in \ball_2} \phi_\vec{u}(\vec{x}) d\boldsymbol{\mu}(\vec{u})\,.
\end{align*}
The total variation norm of the signed measure $\boldsymbol{\mu}$ is defined as the supremum of $\int_{\vec{u}\in\ball_2}q(\vec{u})d\boldsymbol{\mu}(\vec{u})$ over all continuous functions $q(\vec{u})$ that satisfy $|q(\vec{u})|\le 1$. Now we consider the ReLU basis functions $\phi_{\vec{u}}(\vec{x})=\relu{\vec{x}^T\vec{u}}$. For finitely many neurons, the network output is given by
\begin{align*}
    f(\vec{x}) = \sum_{j=1}^m \phi_{\vec{u}_j}(\vec{x}) w_j\,,
\end{align*}
which corresponds to the signed measure $\boldsymbol{\mu} = \sum_{j=1}^m w_j \delta(\vec{u}-\vec{u}_j)$, where $\delta$ is the Dirac delta function. And the total variation norm $\|\boldsymbol{\mu}\|_{TV}$ of $\boldsymbol{\mu}$ reduces to the $\ell_1$-norm $\|\vec{w}\|_1$. 

The infinite dimensional version of the problem \eqref{eq:problem_def2} corresponds to
\begin{align*}
    &\min \|\boldsymbol{\mu}\|_{TV}\\
    &\mbox{s.t. }~ f(\vec{x}_i) = y_i\,,\forall i\in[n]\,.
\end{align*}
For finitely many neurons, i.e., when the measure $\boldsymbol{\mu}$ is a mixture of Dirac delta basis functions, the equivalent problem is
\begin{align*}
    &\min \|\vec{w}\|_{1}\\
    &\mbox{s.t. }~ f(\vec{x}_i) = y_i\,,\forall i\in[n]\,.
\end{align*}
which is identical to \eqref{eq:problem_def2}\,. Similar results also hold with regularized objective functions, different loss functions and vector outputs.

\section{Proofs of the main results} \label{sec:proofs}
In this section, we present the proofs of the theorems and lemmas provided in the main paper.

\subsection{Proofs for the results in Section \ref{sec:main_prelims}}
\begin{proof}\textbf{of Lemma \ref{lemma:reg_equivalence}}
We first note that similar proofs are also presented in \cite{neyshabur_reg,infinite_width,ergen2019cutting,ergen2020aistats,ergen2020convexdeep,ergen2020workshop,pilanci2020convex,ergen2021bn,vikul2021generative}. For any $\theta \in \Theta$, we can rescale the parameters as $\bar{\firstw}_j=\alpha_j\firstw_j$, $\bar{\bias}_j=\alpha_j \bias_j$ and $\bar{\secondw}_j= \secondw_j/\alpha_j$, for any $\alpha_j>0$. Then, \eqref{eq:2layer_function} becomes
\begin{align*}
    f_{\bar{\theta}}(\data)=\sum_{j=1}^m \bar{\secondw}_j (\data \bar{\firstw}_j+\bar{\bias}_j\vec{1})_{+}=\sum_{j=1}^m \frac{\secondw_j}{\alpha_j} (\alpha_j \data \firstw_j+\alpha_j \bias_j\vec{1})_{+}=\sum_{j=1}^m \secondw_j (\data \firstw_j+\bias_j\vec{1})_{+},
\end{align*}
which proves $f_{\theta}(\data)=f_{\bar{\theta}}(\data)$. In addition to this, we have the following basic inequality
\begin{align*}
    \frac{1}{2}\sum_{j=1}^m (\secondw_j^2+\| \firstw_j\|_2^2) \geq \sum_{j=1}^m (|\secondw_j| \text{ }\| \firstw_j\|_2),
\end{align*}
where the equality is achieved with the scaling choice $\alpha_j=\big(\frac{|\secondw_j|}{\| \firstw_j\|_2}\big)^{\frac{1}{2}}$. Since the scaling operation does not change the right-hand side of the inequality, we can set $\|\firstw_j \|_2=1, \forall j$. Therefore, the right-hand side becomes $\| \secondwvec\|_1$.
\end{proof}

\begin{proof}\textbf{of Lemma \ref{lemma:constraint}}
Consider the following problem
\begin{align*} 
    \min_{\theta \in \Theta} \| \secondwvec \|_1 \text{ s.t. } f_{\theta}(\data)=\vec{y}, \| \firstw_j \|_2 \leq 1 , \forall j,
\end{align*}
where the unit norm equality constraint is relaxed. Let us assume that for a certain index $j$, we obtain  $\| \firstw_j \|_2 < 1$ with $\secondw_j \neq 0$ as the optimal solution of the above problem. This shows that the unit norm inequality constraint is not active for $\firstw_j $, and hence removing the constraint for $\firstw_j$ will not change the optimal solution. However, when we remove the constraint, $\| \firstw_j \|_2 \rightarrow \infty$ reduces the objective value since it yields $\secondw_j=0$. Hence, we have a contradiction, which proves that all the constraints that correspond to a nonzero $\secondw_j$ must be active for an optimal solution.
\end{proof}

\begin{proof}\textbf{of Lemma \ref{lemma:spike-free}}
The first condition immediately implies that $\{\relu{\data\vec{u}}|\vec{u}\in\ball_2\}\subseteq \data\ball_2$. Since we also have $\{\relu{\data\vec{u}}|\vec{u}\in\ball_2\} \subseteq \reals_+^n$, it holds that $\{\relu{\data\vec{u}}|\vec{u}\in\ball_2\}\subseteq \data\ball_2 \cap \reals_+^n$. The projection of $\data\ball_2\cap\reals_+^n$ onto the positive orthant is a subset of $\rectset$, and consequently we have $\rectset=\data\ball_2\cap\reals_+^n$.

\noindent The second conditions follow from the $\min$-$\max$ representation
\begin{align*}
    \max_{\vec{u}\in\ball_2}~~~\min_{\vec{z}:\,\data\vec{z}=(\data\vec{u})_+}\|\vec{z}\|_2\le 1 \iff  \eqref{eq:spikefreecond1}\,,
\end{align*}
by noting that $(\vec{I}_n-\data\data^\dagger)(\data\vec{u})_+=\vec{0}$ if and only if there exists $\vec{z}$ such that $\data\vec{z}=(\data\vec{u})_+$, which in that case provided by $\data^\dagger (\data\vec{u})_+$.
The third condition follows from the fact that the minimum norm solution to $\data \vec{z} = (\data\vec{u})_+$ is given by $\data^{\dagger}(\data\vec{u})_+$ under the full row rank assumption on $\data$, which in turn implies $\vec{I}_n-\data\data^\dagger = \vec{0}$. 
\end{proof}

\begin{proof}\textbf{of Lemma \ref{lemma:whiten}}
We have 
\begin{align*}
    \max_{\vec{u}\,:\,\|\vec{u}\|_2\le 1} \| \data^T (\data\data^T)^{-1}\relu{\data \vec{u}}\|_2 &\le \sigma_{\max}(\data^T (\data\data^T)^{-1}) \max_{\vec{u}\,:\,\|\vec{u}\|_2\le 1} \|\relu{\data\vec{u}}\|_2\\
    & = \sigma^{-1}_{\min}(\data) \max_{\vec{u}\,:\,\|\vec{u}\|_2\le 1} \|\relu{\data\vec{u}}\|_2\\
    &\le \sigma^{-1}_{\min}(\data) \max_{\vec{u}\,:\,\|\vec{u}\|_2\le 1} \|\data\vec{u}\|_2\\
    &\le \sigma^{-1}_{\min}(\data) \sigma_{\max}(\data)\\
    &\le 1\,.
\end{align*}
where the last inequality follows from the fact that $\data$ is whitened.
\end{proof}

\begin{proof}\textbf{of Lemma \ref{lemma:rankone_spikefree}}
Let us consider a data matrix $\data$ such that $\data=\vec{c}\vec{\sample}^T$, where $\vec{c} \in \mathbb{R}^n_+$ and $\sample \in \mathbb{R}^d$. Then, $\relu{\data \firstw}=\vec{c}\relu{\sample^T \firstw}$. If $\relu{\sample^T \firstw}=0$, then we can select $\vec{z}=\vec{0}$ to satisfy the spike-free condition $\relu{\vec{c}\sample^T \firstw}=\data \vec{z}$. If $\relu{\sample^T \firstw}\neq0$, then $\relu{\data \firstw}=\vec{c}\sample^T \firstw=\data \firstw $, where the spike-free condition can be trivially satisfied with the choice of $\vec{z}=\firstw$.
\end{proof}

\begin{proof}\textbf{of Lemma \ref{lemma:extreme_l2loss}} \label{proof:extreme_l2loss}
The extreme point along the direction of $\dual$ can be found as follows
\begin{align}
    \label{eq:1d_extreme}
    \argmax_{\firstwscalar, \bias} \sum_{i=1}^n v_i (\samplescalar_i \firstwscalar+\bias)_+ \text{ s.t. } |\firstwscalar|= 1,
\end{align}
Since each neuron separates the samples into two sets, for some samples, ReLU will be active, i.e., $\mathcal{S}=\{i| \samplescalar_i \firstwscalar+\bias \geq 0 \}$, and for the others, it will be inactive, i.e., $\mathcal{S}^c=\{ j|\samplescalar_j \firstwscalar+\bias < 0\}= [n]/\mathcal{S}$. Thus, we modify \eqref{eq:1d_extreme} as
\begin{align} \label{eq:1d_extreme_full}
    \argmax_{\firstwscalar, \bias} \sum_{i \in \mathcal{S}} v_i (\samplescalar_i \firstwscalar+\bias) \text{ s.t. } \samplescalar_i \firstwscalar+\bias \geq 0, \forall i \in \mathcal{S}, \samplescalar_j \firstwscalar+\bias \leq 0, \forall j \in \mathcal{S}^c, |\firstwscalar|= 1.
\end{align}
In \eqref{eq:1d_extreme_full}, $\firstwscalar$ can only take two values, i.e., $\pm 1$. Thus, we can separately solve the optimization problem for each case and then take the maximum one as the optimal. Let us assume that $\firstwscalar=1$. Then, \eqref{eq:1d_extreme_full} reduces to finding the optimal bias. We note that due to the constraints in \eqref{eq:1d_extreme_full}, $-\samplescalar_i \leq b \leq -\samplescalar_j, \forall i \in \mathcal{S},\forall j \in \mathcal{S}^c$. Thus, the range for the possible bias values is $[\max_{i \in \mathcal{S}}(-\samplescalar_i), \text{ }\min_{j \in \mathcal{S}^c}(-\samplescalar_j)]$. Therefore, depending on the direction $\vec{v}$, the optimal bias can be selected as follows
\begin{align}
\label{eq:bias_1d}
\bias_v =\begin{cases}
   \max_{i \in \mathcal{S}}(-\samplescalar_i),  &\text{if } \sum_{i \in \mathcal{S}} v_i \leq 0\\    
  \min_{j \in \mathcal{S}^c}(-\samplescalar_j), &\text{otherwise}   
\end{cases}.
\end{align}
Similar arguments also hold for $\firstwscalar=-1$ and the $\argmin$ version of \eqref{eq:1d_extreme}. Note that when $\sum_{i \in \mathcal{S}} v_i=0$, the value of the bias does not change the objective in \eqref{eq:1d_extreme_full}. Thus, all the bias values in the range $[\max_{i \in \mathcal{S}}(-\samplescalar_i), \text{ }\min_{j \in \mathcal{S}^c}(-\samplescalar_j)]$ become optimal. In such cases, there might exists multiple optimal solutions for the training problem.
\end{proof}

\begin{proof}\textbf{of Lemma \ref{lemma:convex_mixtures}}
 For the extreme point in the span of $\vec{e}_i$, we need to solve the following optimization problem
\begin{align}
    \argmax_{\firstw,\bias} (\sample_i^T \firstw+\bias) \text{ s.t. }  \sample_j^T \firstw+\bias \leq 0, \forall i \neq j, \| \firstw \|_2=1.
    \label{eq:mixture_primal}
\end{align}
Then the Lagrangian of \eqref{eq:mixture_primal} is
\begin{align}
L(\boldsymbol{\lambda},\firstw,\bias)=   \sample_i^T \firstw+\bias-\sum_{\substack{j=1\\ j \neq i}}^n \lambda_j(\sample_j^T \firstw+\bias),\label{eq:mixture_lagrangian}
\end{align}
where we do not include the unit norm constraint for $\firstw$. For \eqref{eq:mixture_lagrangian}, $\boldsymbol{\lambda}$ must satisfy $\boldsymbol{\lambda} \succcurlyeq \vec{0}$ and $\vec{1}^T \boldsymbol{\lambda}=1$. With these specifications, the problem can be written as
\begin{align}\label{eq:mixture_lagrangianp}
    \min_{\boldsymbol{\lambda}} \max_{\firstw}  \firstw^T \bigg(\sample_i -\sum_{\substack{j=1\\ j \neq i}}^n \lambda_j\sample_j\bigg)  \text{ s.t. } \boldsymbol{\lambda} \succcurlyeq \vec{0},\vec{1}^T \boldsymbol{\lambda}=1, \|\firstw \|_2=1.
\end{align}
Since the $\firstw$ vector that maximizes \eqref{eq:mixture_lagrangianp} is the normalized version of the term inside the parenthesis above, the problem reduces to
\begin{align}\label{eq:mixture_problem}
    \min_{\boldsymbol{\lambda}}  \bigg\|\sample_i -\sum_{\substack{j=1\\ j \neq i}}^n \lambda_j\sample_j\bigg\|_2  \text{ s.t. } \boldsymbol{\lambda} \succcurlyeq \vec{0},\vec{1}^T \boldsymbol{\lambda}=1.
\end{align}
After solving the convex problem \eqref{eq:mixture_problem} for each $i$, we can find the corresponding neurons as follows
\begin{align*}
    \firstw_i=\frac{\sample_i -\sum_{\substack{j=1\\ j \neq i}}^n \lambda_j\sample_j}{\bigg\|\sample_i -\sum_{\substack{j=1\\ j \neq i}}^n \lambda_j\sample_j\bigg\|_2} \text{  and  } \bias_i= \min_{j\neq i} (-\sample_j^T \firstw_i),
\end{align*}
where the bias computation follows from the constraint in \eqref{eq:mixture_primal}.
\end{proof}

\begin{proof}\textbf{of Lemma \ref{lemma:extreme_general_closedform}}
For any $\boldsymbol{\alpha} \in \mathbb{R}^n$, the extreme point along the direction of $\boldsymbol{\alpha}$ can be found by solving the following optimization problem
\begin{align} \label{eq:generic_extreme}
    \argmax_{\firstw, \bias} \boldsymbol{\alpha}^T (\data \firstw + \bias \vec{1})_+ \text{ s.t. } \| \firstw \|_2\le1
\end{align}
where the optimal $(\firstw, \bias)$ groups samples into two sets so that some of them activates ReLU with the indices $\mathcal{S}=\{i| \sample_i^T \firstw +b\geq 0 \}$ and the others deactivate it with the indices $\mathcal{S}^c=\{ j|\sample_j^T \firstw +b < 0  \}=[n]/ \mathcal{S}$. Using this, we equivalently write \eqref{eq:generic_extreme} as 
\begin{align*}
       \max_{\firstw, \bias} \sum_{i \in \mathcal{S}} \alpha_i (\sample_i^T \firstw +b) \text{ s.t. }(\sample_i^T \firstw +b) \geq 0, \forall i\in\mathcal{S},(\sample_j^T \firstw +b) \leq 0, \forall j\in\mathcal{S}^c, \| \firstw \|_2\le 1,
\end{align*}
which has the following dual form
\begin{align*}
    &\min_{\boldsymbol{\lambda}, \boldsymbol{\nu}} \max_{\firstw,\bias} \firstw^T \bigg (\sum_{i \in \mathcal{S}} (\alpha_i+\lambda_i)\sample_i- \sum_{j \in \mathcal{S}^c} \nu_j \sample_j  \bigg)  \text{ s.t. } \boldsymbol{\lambda}, \boldsymbol{\nu} \succcurlyeq \vec{0}, \sum_{i \in \mathcal{S}} (\alpha_i + \lambda_i)= \sum_{j \in \mathcal{S}^c} \nu_j,  \| \firstw \|_2 \le 1.
\end{align*}
Thus, we obtain the following neuron and bias choice for the extreme point
\begin{align*}
    \firstw_{\alpha}= \frac{\sum_{i \in \mathcal{S}} (\alpha_i+\lambda_i) \sample_i - \sum_{j \in \mathcal{S}^c} \nu_j \sample_j }{\|\sum_{i \in \mathcal{S}} (\alpha_i+\lambda_i) \sample_i - \sum_{j \in \mathcal{S}^c} \nu_j \sample_j \|_2} \text{ and } \bias_{\alpha}  =\begin{cases}
   \max_{i \in \mathcal{S}}(-\sample_i^T \firstw),  &\text{if } \sum_{i \in \mathcal{S}} \alpha_i \leq 0\\    
  \min_{j \in \mathcal{S}^c}(-\sample_j^T \firstw), &\text{otherwise}   
\end{cases}.
\end{align*}
\end{proof}

\subsection{Proofs for the results in Section \ref{sec:main_results}}

\begin{proof}\textbf{of Theorem \ref{theo:equality_dual} and Corollary \ref{cor:extreme_points_optimality}}

We first note that the dual of \eqref{eq:problem_def2} with respect to $\secondwvec$ is
\begin{align*}
    \min_{\theta \in \Theta \backslash\{\secondwvec\}} \max_{\dual} \dual^T \vec{y} \text{ s.t. } \|(\data \firstwmat)_{+}^T \dual \|_{\infty} \leq 1, \;\| \firstw_j \|_2\leq1 , \forall j.
\end{align*}
Then, we can reformulate the problem as follows
\begin{align*}
P^*=\min_{\theta \in \Theta \backslash \{\secondwvec\}}\max_{\dual} \dual^T \vec{y}  +\mathcal{I}(\|(\data \firstwmat)_{+}^T \dual \|_{\infty} \leq 1) , \text{ s.t. } \| \firstw_j \|_2\leq1 , \forall j.
\end{align*}
where $\mathcal{I}(\|(\data \firstwmat)_{+}^T \dual \|_{\infty} \leq 1)$ is the characteristic function of the set $\|(\data \firstwmat)_{+}^T \dual \|_{\infty} \leq 1 $, which is defined as
\begin{align*}
    \mathcal{I}(\|(\data \firstwmat)_{+}^T \dual \|_{\infty} \leq 1) = \begin{cases} 0 & \text{ if }\|(\data \firstwmat)_{+}^T \dual \|_{\infty} \leq 1 \\
    -\infty & \mbox{otherwise}\end{cases}.
\end{align*}
Since the set $\|(\data \firstwmat)_{+}^T \dual \|_{\infty}\le 1$ is closed, the function $\Phi(\dual,\firstwmat)=\dual^T \vec{y}  +\mathcal{I}(\|(\data \firstwmat)_{+}^T \dual \|_{\infty} \leq 1)$ is the sum of a linear function and an upper-semicontinuous indicator function and therefore upper-semicontinuous. The constraint on $\firstwmat$ is convex and compact. We use $P^*$ to denote the value of the above $\min$-$\max$ program. Exchanging the order of $\min$ and $\max$ we obtain the dual problem given in \eqref{eq:problem_dual}, which establishes a lower bound $D^*$ for the above problem:
\begin{align*}
P^*\ge D^*&=\max_{\dual}\min_{\theta \in \Theta \backslash \{\secondwvec\}} \dual^T \vec{y}  +\mathcal{I}(\|(\data \firstwmat)_{+}^T \dual \|_{\infty} \leq 1) , \text{ s.t. } \| \firstw_j \|_2\leq1 , \forall j,\\
&= \max_{\dual} \dual^T \vec{y}, \text{ s.t. } \|(\data \firstwmat)_{+}^T \dual \|_{\infty} \leq 1 ~\forall \firstw_j \mbox{ : } \| \firstw_j \|_2\leq1 , \forall j, \\
&= \max_{\dual} \dual^T \vec{y}, \text{ s.t. } \|(\data \firstw)_{+}^T \dual \|_{\infty} \leq 1 ~\forall \firstw \mbox{ : } \| \firstw \|_2\leq 1 , 
\end{align*}
We now show that strong duality holds for infinite size NNs. The dual of the semi-infinite program in \eqref{eq:problem_dual} is given by (see Section 2.2 of \citet{semiinfinite_goberna} and also \citet{bach2017breaking})
\begin{align*}
    &\min \|\boldsymbol{\mu}\|_{TV}\\
    &\mbox{s.t.} \int_{\vec{u}\in \ball_2} \relu{\data\vec{u}}d\boldsymbol{\mu}(\vec{u}) = \vec{y}\,,
\end{align*}
where TV is the total variation norm of the Radon measure $\boldsymbol{\mu}$. This expression coincides with the infinite-size NN as given in Section \ref{sec:infsize}, and therefore strong duality holds. We also remark that even though the above problem involves an infinite dimensional integral form, by Caratheodory's theorem, this integral form can be represented as a finite summation with at most $n+1$ Dirac delta functions \citep{rosset2007}. Next we invoke the semi-infinite optimality conditions for the dual problem in \eqref{eq:problem_dual}, in particular we apply Theorem 7.2 of \citet{semiinfinite_goberna}. We first define the set
\begin{align*}
    \mathbf{K}=\mathbf{cone}\left\{ \left( \begin{array}{c}s\relu{\data \firstw} \\ 1 \end{array}  \right), \firstw \in \ball_2, s\in\{-1,+1\}; \left(\begin{array}{c} \vec{0} \\ -1\end{array}\right) \right\}\,.
\end{align*}
Note that $\mathbf{K}$ is the union of finitely many convex closed sets, since the function $\relu{\data\firstw}$ can be expressed as the union of finitely many convex closed sets. Therefore the set $\mathbf{K}$ is closed. By Theorem 5.3 of \citet{semiinfinite_goberna}, this implies that the set of constraints in \eqref{eq:problem_dual} forms a Farkas-Minkowski system. By Theorem 8.4 of \citet{semiinfinite_goberna}, primal and dual values are equal, given that the system is consistent. Moreover, the system is discretizable, i.e., there exists a sequence of problems with finitely many constraints whose optimal values approach to the optimal value of \eqref{eq:problem_dual}. The optimality conditions in Theorem 7.2 of \citet{semiinfinite_goberna} implies that $\vec{y}=\relu{\data\firstwmat^*}\secondwvec^*$ for some vector $\secondwvec^*$. Since the primal and dual values are equal, we have ${\dual^*}^T \vec{y} ={\dual^*}^T \relu{\data\firstwmat^*}\secondwvec^*= \|\secondwvec^*\|_1$, which shows that the primal-dual pair $\left(\{\secondwvec^*,\firstwmat^*\} ,\dual^*\right)$ is optimal. Thus, the optimal neuron weights $\firstwmat^*$ satisfy $\|(\data \firstwmat^*)_{+}^T \dual^* \|_{\infty} = 1$.
\end{proof}


\begin{proof}\textbf{of Theorem \ref{theo:weak_duality}}
We first assume that zero training error can be achieved with $m_1$ neurons. Then, we obtain the dual of \eqref{eq:problem_def2} with $m=m_1$
\begin{align}\label{eq:weak_dual_problemw}
    P_f^*=\min_{\theta \in \Theta \backslash\{\secondwvec,m\}} \max_{\dual} \dual^T \vec{y} \text{ s.t. } \|(\data \firstwmat)_{+}^T \dual \|_{\infty} \leq 1, \;\| \firstw_j \|_2\leq1 , \forall j \in [m_1].
\end{align}
 Exchanging the order of $\min$ and $\max$ establishes a lower bound for \eqref{eq:weak_dual_problemw}
\begin{align}\label{eq:weak_dual_problemw2}
P_f^*\ge D_f^*=\max_{\dual}\min_{\theta \in \Theta \backslash \{\secondwvec,m\}} \dual^T \vec{y}  +\mathcal{I}(\|(\data \firstwmat)_{+}^T \dual \|_{\infty} \leq 1) , \text{ s.t. } \| \firstw_j \|_2\leq1 , \forall j \in [m_1].
\end{align}
If we denote the optimal parameters to \eqref{eq:weak_dual_problemw} as $\firstwmat^*$ and $\dual^*$, then $|(\data \firstwmat^*)_+^T \dual^*|=\vec{1}$ must hold, i.e., all the optimal neuron weights must achieve the extreme point of the inequality constraint. To prove this, let us consider an optimal neuron $\firstw_j^*$, which has a nonzero weight $\secondw_j\neq 0$ and $|(\data \firstw_j^*)_+^T \dual^*|<1$. Then, even if we remove the inequality constraint for $\firstw_j^*$ in \eqref{eq:weak_dual_problemw}, the optimal objective value will not change. However, if  we remove it, then $\firstw_j^*$ will no longer contribute to $(\data \firstwmat)_+ \secondwvec=\vec{y}$. Then, we can achieve a smaller objective value, i.e., $\| \secondwvec\|_1$, by simply setting $\secondw_j=0$. Thus, we obtain a contradiction, which proves that the inequality constraints that correspond to the neurons with nonzero weight, $\secondw_j\neq0$, must achieve the extreme point for the optimal solution, i.e., $|(\data \firstw_j^*)_+^T \dual^*|=1, \forall j \in [m]$.

Based on this observation, we have 
\begin{align}\label{eq:weak_dual_problemw_result}
   P_f^*=\nonumber &\min_{\theta \in \Theta \backslash \{\secondwvec,m\}}\max_{\dual}  \dual^T \vec{y}  \hspace{3.0cm}\geq &&\min_{\theta \in \Theta \backslash \{\secondwvec\}}\max_{\dual} \dual^T \vec{y}  \\    \nonumber&\text{s.t. } \|(\data \firstwmat)_{+}^T \dual \|_{\infty} \leq 1, \;\| \firstw_j \|_2\leq1 , \forall j \in [m_1] &&\text{s.t. } \|(\data \firstwmat)_{+}^T \dual \|_{\infty} \leq 1, \;\| \firstw_j \|_2\leq1 , \forall j  \\
     \nonumber &\hspace{6.25cm}  \geq  &&\max_{\dual}\min_{\theta \in \Theta \backslash \{\secondwvec\}}  \dual^T \vec{y}  \\    \nonumber&\hfill &&\text{s.t. } \|(\data \firstwmat)_{+}^T \dual \|_{\infty} \leq 1, \;\| \firstw_j \|_2\leq1 , \forall j  \\ \nonumber
       &\hspace{6.25cm}  =  &&\max_{\dual}\min_{\theta \in \Theta \backslash \{\secondwvec,m\}}  \dual^T \vec{y} \\ \nonumber &\hfill &&\text{s.t. } (\|(\data \firstwmat)_{+}^T \dual \|_{\infty} \leq 1, \;\| \firstw_j \|_2\leq1 , \forall j  \in [m_1]\\
       &\hspace{6.25cm}  = &&D_f^*=D^*
\end{align}
where the first inequality is based on the fact that an infinite width NN can always find a solution with the objective value lower than or equal to the objective value of a finite width NN. The second inequality follows from \eqref{eq:weak_dual_problemw2}. More importantly, the equality in the third line follows from our observation above, i.e., neurons that are not the extreme point of the inequality in \eqref{eq:weak_dual_problemw} do not change the objective value. Therefore, by \eqref{eq:weak_dual_problemw_result}, we prove that weak duality holds for a finite width NN, i.e., $P_f^*\geq P^* \geq  D_f^*=D^*$.

\end{proof}


\begin{proof}\textbf{of Theorem \ref{theo:strong_duality}}
By Caratheodory's theorem (see \citet{rosset2007}), the number of constraints active in the dual problem is bounded by $n+1$. Suppose this number is $m^*$, where $m^* \leq n+1$. Thus, we can construct a weight matrix $\firstwmat_e \in \mathbb{R}^{d \times m^*}$ that consists of all the extreme points. Next, the dual of \eqref{eq:problem_def2} with $\firstwmat=\firstwmat_e$
\begin{align}\label{eq:strong_dual_problemw}
    D_f^*= \max_{\dual} \dual^T \vec{y} \text{ s.t. } \|(\data \firstwmat_e)_{+}^T \dual \|_{\infty} \leq 1,
\end{align}
Consequently, we have 
\begin{align}\label{eq:strong_dual_problemw_result}
   P^*=\nonumber &\min_{\theta \in \Theta \backslash \{\secondwvec\}}\max_{\dual} \dual^T \vec{y} \hspace{3.3cm}\geq &&\max_{\dual}\min_{\theta \in \Theta \backslash \{\secondwvec\}}  \dual^T \vec{y}  \\    \nonumber&\text{s.t }\|(\data \firstwmat)_{+}^T \dual \|_{\infty} \leq 1, \;\| \firstw_j \|_2\leq1 , \forall j &&\text{s.t. } \|(\data \firstwmat)_{+}^T \dual \|_{\infty} \leq 1, \;\| \firstw_j \|_2\leq1 , \forall j  \\ \nonumber
       &\hspace{6.25cm}  =  &&\max_{\dual}  \dual^T \vec{y} \\ \nonumber &\hfill &&\text{s.t. } (\|(\data \firstwmat_e)_{+}^T \dual \|_{\infty} \leq 1\\
       &\hspace{6.25cm}  = &&D_f^*=D^*
\end{align}
where the first inequality follows from changing order of min-max to obtain a lower bound and the equality in the second line follows from Corollary \ref{cor:extreme_points_optimality} and our observation above, i.e., neurons that are not the extreme point of the inequality in \eqref{eq:strong_dual_problemw} do not change the objective value. 

From the fact that an infinite width NN can always find a solution with the objective value lower than or equal to the objective value of a finite width NN, we have
\begin{align}\label{eq:strong_original_upperbound}
  P^*_f=&\min_{\theta \in \Theta \backslash \{\firstwmat,m\}} \| \secondwvec\|_1 \hspace{2cm}\geq \hspace{1cm} P^*= \hspace{-1cm}&&\min_{\theta \in \Theta  } \| \secondwvec\|_1 \\\nonumber &\text{s.t. } (\data \firstwmat_e)_{+} \secondwvec  =\vec{y}   &&\text{s.t. } (\data \firstwmat)_{+} \secondwvec  =\vec{y}, \;\| \firstw_j \|_2\leq1 , \forall j,
\end{align}
where $P^*$ is the optimal value of the original problem with infinitely many neurons. Now, notice that the optimization problem on the left hand side of \eqref{eq:strong_original_upperbound} is convex since it is an $\ell_1$-norm minimization problem with linear equality constraints. Therefore, strong duality holds for this problem, i.e., $P^*_f=D^*_f$ and we have $ P^* \geq D^* =D_f^*$. Using this result along with \eqref{eq:strong_dual_problemw_result}, we prove that strong duality holds for a finite width NN, i.e., $P_f^* = P^* = D^* =D_f^*$.
\end{proof}


	\begin{proof}\textbf{of Proposition \ref{prop:1d_uniqueness}} 
	
	We first note that the conditions related to the range of bias values that lead to non-uniqueness directly follows from Proof of Lemma \ref{lemma:extreme_l2loss}. Hence here, we particularly examine the problem in \eqref{eq:problem_def2} when we have a one dimensional dataset, i.e., $\{\samplescalar_i,y_i \}_{i=1}^n$, to provide analytic forms for the counter-examples depicted in Figure \ref{fig:uniquness}. Then, \eqref{eq:problem_def2} can be modified as 
 \begin{align}
   \min_{\theta \in \Theta} \| \secondwvec \|_1 \text{ s.t. } (\sample\firstw^T+\vec{1}\vec{b}^T)_+\secondwvec=\vec{y}, | \firstwscalar_j| \leq 1 , \forall j.\label{eq:problem_def_1d}
\end{align}
Then, using Lemma \ref{lemma:extreme_l2loss}, we can construct the following matrix
\begin{align*}
   \data_e= (\sample \firstw^{*^T}+\vec{1}\biasvec^{*^T})_+,
\end{align*}
where $\firstw^*$ and $\biasvec^*$ consist of all possible extreme points. Using this definition and Corollary \ref{cor:1d_optimality}, we can rewrite \eqref{eq:problem_def_1d} as
\begin{align}
        \label{eq:dual_problem_1dv4}
    \min_{\secondwvec} \|\secondwvec\|_1 \text{ s.t. } \data_e\secondwvec=\vec{y}.
\end{align}
In the following, we first derive optimality conditions for \eqref{eq:dual_problem_1dv4} and then provide an analytic counter example to disprove uniqueness. Then, we also follow the same steps for the regularized version of \eqref{eq:dual_problem_1dv4}.\\

\noindent\textbf{Equality constraint:}
The optimality conditions for \eqref{eq:dual_problem_1dv4} are
\begin{align}\label{eq:1d_optimalityeq_conditions}
    \begin{split}
    & \data_e \secondwvec^*=\vec{y}\\
        &\data_{e,s}^T \dual^* +\sign(\secondwvec_s^*)=0 \\
    &\|\data_{e,s^c}^T \dual^*\|_\infty \leq 1,
    \end{split}
\end{align}
where the subscript $s$ denotes the entries of a vector (or columns for matrices) that correspond to a nonzero weight, i.e. $w_i \neq 0$, and the subscript $s^c$ denotes the remaining entries (or columns). We aim to find an optimal primal-dual pair that satisfies \eqref{eq:1d_optimalityeq_conditions}.

Now, let us consider a specific dataset, i.e., $\sample=[-2\; -1 \; 0 \;1 \; 2]^T$ and $\vec{y}=[1\;-1\;1\;1\;-1]^T$, and yields the following
\begin{align*}
  \data_e= (\sample \firstw^{*^T} + \vec{1}\biasvec^{*^T})=\begin{bmatrix}  0& 0&0&0&1&2&3&4\\
     1& 0&0&0&0&1&2&3\\
    2& 1&0&0&0&0&1&2\\
    3& 2&1&0&0&0&0&1\\
    4& 3&2&1&0&0&0&0 
    \end{bmatrix},
\end{align*}
where $\firstw^{*^T}=[1 \; 1 \;1 \;1 \; -1 \; -1 \;-1 \;-1]$ and $\biasvec^{*^T}=[2 \; 1 \; 0 \; -1 \; -1\;0 \;1 \;2]$. Solving \eqref{eq:dual_problem_1dv4} for this dataset gives
\begin{align*}
 \dual^*=\begin{bmatrix}1 \\-3\\2\\1\\-1\end{bmatrix} \text{ and } \secondwvec^*=\begin{bmatrix}0 \\  6419/5000\\
 -3919/2500\\
 -8581/5000\\
 13581/5000\\
 -1081/2500\\
 -1419/5000\\0\end{bmatrix} \implies  \|\secondwvec^* \|_1=8 .
\end{align*}
We can also achieve the same objective value by using the following matrix
\begin{align*}
   \hat{\data}_e= (\sample \hat{\firstw}^{T} + \vec{1}\hat{\biasvec}^{T})=\begin{bmatrix}  0& 0&0&0&1&2&2.5&4\\
     1& 0&0&0&0&1&1.5&3\\
    2& 1&0&0&0&0&0.5&2\\
    3& 2&1&0.5&0&0&0&1\\
    4& 3&2&1.5&0&0&0&0 
    \end{bmatrix}, 
\end{align*}
where $\hat{\firstw}^{T}=[1 \; 1 \;1 \;1 \; -1 \; -1 \;-1 \;-1]$ and $\hat{\biasvec}^{T}=[2 \; 1 \; 0 \; -0.5 \; -1\;0 \;0.5 \;2]$. Solving \eqref{eq:dual_problem_1dv4} for this dataset yields
\begin{align*}
 \hat{\dual}=\begin{bmatrix}1 \\-11/4\\5/4\\7/4\\-5/4\end{bmatrix} \text{ and } \hat{\secondwvec}=\begin{bmatrix}     0\\4/3\\  0\\-10/3\\8/3\\0\\  -2/3\\0\end{bmatrix} \implies  \|\hat{\secondwvec} \|_1=8 .
\end{align*}
We also note that both solutions satisfy the optimality conditions in \eqref{eq:1d_optimalityeq_conditions}.\\

\noindent\textbf{Regularized case:}
The regularized version of \eqref{eq:dual_problem_1dv4} is as follows
\begin{align}\label{eq:dual_problem_1dreg}
    \min_{\secondwvec} \beta \|\secondwvec \|_1+\frac{1}{2n} \| \data_e \secondwvec -\vec{y}\|_2^2,
\end{align}
where the optimal solution $\secondwvec^*$ satisfies
\begin{align}\label{eq:1d_optimalityreg_conditions}
\begin{split}
        &\frac{1}{n}\data_{e,s}^T (\data_e\secondwvec^* -\vec{y})+\beta \sign(\secondwvec_s^*)=0 \\
    &\|\data_{e,s^c}^{T}(\data_e \secondwvec^*-\vec{y})\|_\infty \leq \beta n,
\end{split}
\end{align}
where the subscript $s$ denotes the entries of a vector (or columns for matrices) that correspond to a nonzero weight, i.e. $w_i \neq 0$, and the subscript $s^c$ denotes the remaining entries (or columns).
Now, let us consider a specific dataset, i.e., $\sample=[-2\; -1 \; 0 \;1 \; 2]^T$ and $\vec{y}=[1\;-1\;1\;1\;-1]^T$. We then construct the following matrix
\begin{align*}
   \data_e= (\sample \firstw^{*^T} + \vec{1}\biasvec^{*^T})=\begin{bmatrix}  0& 0&0&0&0&1&2&2.5&3&4\\
     1& 0&0&0&0&0&1&1.5&2&3\\
    2& 1&0&0&0&0&0&0.5&1&2\\
    3& 2&1&0.5&0&0&0&0&0&1\\
    4& 3&2&1.5&1&0&0&0&0&0 
    \end{bmatrix},
\end{align*}
where $\firstw^{*^T}=[1 \; 1 \;1 \;1\;1 \; -1 \;-1\; -1 \;-1 \;-1]$ and $\biasvec^{*^T}=[2 \; 1 \; 0 \;-0.5\; -1 \; -1\;0\;0.5 \;1 \;2]$. For this dataset with $\beta=10^{-4}$, the optimal value of \eqref{eq:dual_problem_1dreg} can be achieved by the following solutions
\begin{align*}
\begin{split}
    \secondwvec_1=\begin{bmatrix}0 \\ 3197/2400
 \\   -2497/1500
 \\ 0
 \\  -19997/12000
\\   31961/12000
\\    -997/3000
\\0
\\  -3997/12000
\\0\end{bmatrix} \implies \beta \|\secondwvec_1 \|_1+\frac{1}{2n} \| \data_e \secondwvec_1 -\vec{y}\|_2^2= \frac{1999}{2500000}  
\end{split}\\
\begin{split}
    \secondwvec_2=\begin{bmatrix}0 \\    191823/140000
 \\     -990613/840000
 \\    -471683/420000
 \\     -128017/120000
\\       367547/140000
\\     -127357/840000
\\  -87827/420000
\\      -31993/120000
\\0\end{bmatrix} \implies \beta \|\secondwvec_2 \|_1+\frac{1}{2n} \| \data_e \secondwvec_2 -\vec{y}\|_2^2= \frac{1999}{2500000}  
\end{split},
\end{align*}
where each solution satisfies the optimality conditions in \eqref{eq:1d_optimalityreg_conditions}. We also provide a visualization for the output functions of each solution in Figure \ref{fig:uniquness}, namely Solution1 and Solution2. 

\begin{rems} \label{remark:spline_solutions}
In fact, there exist infinitely many solutions to the regularized training problem discussed above, which can be analytically defined by the following weights and biases
\begin{align*}
    \firstw^{*^T}=[1 \; 1 \;1 \;1\;1 \; -1 \;-1\; -1 \;-1 \;-1],\; \biasvec^{*^T}=[2 \; 1 \; 0 \;-c\; -1 \; -1\;0\;c \;1 \;2]
\end{align*}
where $c$ can be arbitrarily chosen to satisfy $0\leq c \leq1$. As a numerical proof, we also provide two additional examples with $c=0.2$ and $c=0.8$, i.e., Solution3 and Solution4, in Figure \ref{fig:uniquness}. These solutions also achieve the same objective value and their last layer weights are as follows
\begin{align*}
    \secondwvec_3=\begin{bmatrix}0 \\        323691/248000\\
 -7349999/5208000\\
 -1039627/1041600\\
 -4660169/5208000\\
    667193/248000\\
 -1810997/5208000\\
  -199753/1041600\\
  -795707/5208000
\\0\end{bmatrix} 
\text{ and }
    \secondwvec_4=\begin{bmatrix}0 \\      167847/116000\\
 -1248409/1218000\\
  -1058987/974400\\
 -6500131/4872000\\
    295631/116000\\
   -25387/1218000\\
    -99563/974400\\
 -2082883/4872000
\\0\end{bmatrix} .
\end{align*}
This numerical observation can also be explained via our extreme point characterization in \eqref{eq:bias_1d}, where the optimal bias can take one of infinitely many possible values in a certain interval when $\sum_{i \in \mathcal{S}} v_i=0$.
\end{rems}

	\end{proof}


\begin{proof}\textbf{of Corollary \ref{cor:rankone_extreme}}
Given $\data=\vec{c}\sample^T$, all possible extreme points can be characterized as follows
\begin{align*}
        \argmax_{\bias,\firstw:\|\firstw\|_2= 1 } |{\dual}^T \relu{\data \firstw+\bias \vec{1}}\,|&=\argmax_{\bias,\firstw:\|\firstw\|_2= 1 } |{\dual}^T \relu{\vec{c} \vec{a}^T \firstw +\bias\vec{1}}\,|\\
        &=\argmax_{\bias,\firstw:\|\firstw\|_2= 1 } \Big|\sum_{i=1}^n \dualscalar_i \relu{c_i \vec{a}^T \firstw+\bias }\,\Big|
\end{align*}
which can be equivalently stated as
\begin{align*}
    \argmax_{\bias,\firstw:\|\firstw\|_2= 1} \sum_{i \in \mathcal{S}} \dualscalar_i c_i\sample^T \firstw+\sum_{i \in \mathcal{S}}\dualscalar_i \bias \text{ s.t. } \begin{cases} &c_i\sample^T \firstw+\bias \geq 0, \forall i \in \mathcal{S} \\
    &c_j\sample^T \firstw+\bias  \leq 0, \forall j \in \mathcal{S}^c\end{cases}
,\end{align*}
which shows that $\firstw$ must be either positively or negatively aligned with $\sample$, i.e.,
$\firstw= s \frac{\sample}{\|\sample\|_2}$, where $s= \pm 1$. Thus, $\bias$ must be in the range of $[\max_{i \in \mathcal{S}}(-s c_i \| \sample\|_2), \text{ }\min_{j \in \mathcal{S}^c}(-s c_j \| \sample\|_2)]$ Using these observations, extreme points can be formulated as follows
\begin{align*}
    \vec{u}_v= \begin{cases}\frac{\sample}{\| \sample\|_2} &\text{if } \sum_{i \in \mathcal{S}} \dualscalar_i c_i \geq 0\\\frac{-\sample}{\| \sample\|_2} & \text{otherwise}\end{cases}
    \text{ and } \bias_v= \begin{cases}\min_{j \in \mathcal{S}^c}(-s_v c_j \| \sample\|_2) &\text{if } \sum_{i \in \mathcal{S}} \dualscalar_i \geq 0\\\max_{i \in \mathcal{S}}(-s_v c_i \| \sample\|_2) & \text{otherwise}\end{cases},
\end{align*}
where $s_v= \sign(\sum_{i \in \mathcal{S}} \dualscalar_i c_i)$.
\end{proof}


\begin{proof}\textbf{of Lemma \ref{lemma:closedform_2neuron}}
Since $\vec{y}$ has both positive and negative entries, we need at least two $\firstw$'s with positive and negative output weights to represent $\vec{y}$ using the output range of ReLU. Therefore the optimal value of the $\ell_0$ problem is at least 2. Note that $\data \data^{\dagger} = \vec{I}_n$ since $\data$ is full row rank. Then let us define the output weights
\begin{align*}
    \secondw_1 &= \|\data^\dagger \relu{\vec{y}}\|_2\\
     \secondw_2 &= -\|\data^\dagger \relu{\vec{-y}}\|_2\,.
\end{align*}
Then note that
\begin{align*}
    \secondw_1\relu{\data\firstw_1}+\secondw_2\relu{\data\firstw_2} &= \relu{\data \data^{\dagger} \relu{\vec{y}}} - \relu{\data \data^{\dagger} \relu{-\vec{y}}}\\
    &= \relu{\relu{\vec{y}}} - \relu{\relu{-\vec{y}}}\\
    & = \relu{\vec{y}} - \relu{-\vec{y}}\\
    & = \vec{y}
\end{align*}
where the second equality follows from $\data \data^{\dagger} = \vec{I}_n$.
\end{proof}
\begin{proof}\textbf{of Lemma \ref{lemma:l1-l0 equivalence}}
We first provide the optimality conditions for the convex program in the following proposition:
\begin{props}\label{prop:optimality}
Let $\firstwmat$ be a weight matrix for \eqref{eq:problem_def2}. Then, $\firstwmat\in\reals^{d\times m}$ is an optimal solution for the regularized training problem if 
\begin{align}\label{eq:optimality_feasible_cons}
\exists\boldsymbol{\alpha}\in\reals^{n}, \secondwvec\in\reals^{m}  \text{ s.t. } \relu{\data\firstwmat}\secondwvec=\vec{y},\quad  \relu{\data\firstwmat}^T \boldsymbol{\alpha} = \sign(\secondwvec)\,
\end{align}
and 
\begin{align}\label{eq:optimality_max_cons}
    \max_{\firstw\,:\, \| \firstw \|_2 \le 1} |\boldsymbol{\alpha}^T (\data \firstw)_+| \leq 1\,.
\end{align}
\end{props}
These conditions follow from linear semi-infinite optimality conditions given in Theorem 7.1 and 7.6 of \citet{semiinfinite_goberna} for Farkas-Minkowski systems. Then the proof  Lemma \ref{lemma:l1-l0 equivalence} directly follows from the solution of minimum cardinality problem given in Lemma \ref{lemma:closedform_2neuron}.

Now we prove the second claim. For whitened data matrices, denoting the Singular Value Decomposition of the input data as $\data = \vec{U}$ where $\vec{U}^T\vec{U}=\vec{U}\vec{U}^T=\vec{I}_n$ since $\data$ is assumed full row rank. Consider the dual optimization problem
\begin{align}
    \max_{|\dual^T(\data\vec{u})_+|\le 1,~\forall \vec{u}\in\ball_2} \dual^T \vec{y} \label{eq:dualproof}
\end{align}
Changing the variable to $\vec{u}^\prime = \vec{U} \vec{u}$ in the dual problem we next show that
\newcommand{\up}{\vec{u}^\prime}
\begin{align}
    \max_{|\dual^T(\up)_+|\le 1,~\forall \up\in\ball_2} \dual^T \vec{y} = \max_{\|(\dual)_+\|_2\le 1,~\|(-\dual)_+\|_2\le 1} \dual^T \vec{y}\,. \label{eq:dualproof2}
\end{align}
where the equality follows from the upper bound
\begin{align*}
    \dual^T(\up)_+ \le (\dual)_+^T (\up)_+ \le \|(\dual)_+\|_2 \|(\up)_+\|_2\le \|(\dual)_+\|_2\,,
\end{align*}
which is achieved when $\up=\frac{(\dual)_+}{\|(\dual)_+\|_2}$. Similarly we have
\begin{align*}
    -\dual^T(\up)_+ \le (-\dual)_+^T (\up)_+ \le \|(-\dual)_+\|_2 \|(\up)_+\|_2\le \|(-\dual)_+\|_2\,,
\end{align*}
which is achieved when $\up=\frac{(-\dual)_+}{\|(-\dual)_+\|_2}$, which verifies the right-hand-side of \eqref{eq:dualproof2}. Now note that
\begin{align*}
    \dual^T \vec{y} \le (\dual)_+^T(\vec{y})_+ + (-\dual)_+^T(-\vec{y})_+\,.
\end{align*}
Therefore the right-hand-side of \eqref{eq:dualproof2} is upper-bounded by $\|(\vec{y})_+\|_2+\|(-\vec{y})_+\|_2$. This upper-bound is achieved by the choice
\begin{align*}
    \dual = \frac{(\vec{y})_+}{\|(\vec{y})_+\|_2} - \frac{(-\vec{y})_+}{\|(-\vec{y})_+\|_2}\,,
\end{align*}
since we have
\begin{align*}
    \dual^T\vec{y} &= \frac{\vec{y}^T(\vec{y})_+}{\|(\vec{y})_+\|_2} - \frac{\vec{y}^T(-\vec{y})_+}{\|(-\vec{y})_+\|_2}=\frac{(\vec{y})_+^T(\vec{y})_+}{\|(\vec{y})_+\|_2} + \frac{(-\vec{y})_+^T(-\vec{y})_+}{\|(-\vec{y})_+\|_2}\\
    &=\|(\vec{y})_+\|_2+\|(-\vec{y})_+\|_2\,.
\end{align*}
Therefore the preceding choice of $\dual$ is optimal. Consequently, the corresponding optimal neuron weights satisfy
\begin{align*}
    \up_1=\frac{(\vec{y})_+}{\|(\vec{y})_+\|_2}\quad\mbox{and}\quad \up_2=\frac{(-\vec{y})_+}{\|(-\vec{y})_+\|_2}\,.
\end{align*}
Changing the variable back via $\vec{u}=\vec{U}^T\up=\data^{\dagger}\up$ we conclude that the optimal neurons are given by
\begin{align*}
    \vec{u}_1=\frac{\data^{\dagger}(\vec{y})_+}{\|(\vec{y})_+\|_2}\quad\mbox{and}\quad \vec{u}_2=\frac{\data^{\dagger}(-\vec{y})_+}{\|(-\vec{y})_+\|_2}\,,
\end{align*}
or equivalently 
\begin{align*}
    \vec{u}_1=\frac{\data^{\dagger}(\vec{y})_+}{\|\data^{\dagger}(\vec{y})_+\|_2}\quad\mbox{and}\quad \vec{u}_2=\frac{\data^{\dagger}(-\vec{y})_+}{\|\data^{\dagger}(-\vec{y})_+\|_2}\,,
\end{align*}
since $\data^\dagger$ is orthonormal and yields the claimed expression. Finally, note that the corresponding output weights are $\|\data^{\dagger}(\vec{y})_+\|_2$ and $\|\data^{\dagger}(-\vec{y})_+\|_2$, respectively.
\end{proof}

\begin{proof}\textbf{of Proposition \ref{prop:cutting_optimality}}
Since the constraint in \eqref{eq:vio_convex} is bounded below and the hidden layer weights are constrained to the unit Euclidean ball, the convergence of the cutting plane method directly follows from Theorem 11.2 of \citet{semiinfinite_goberna}.
\end{proof}

\begin{proof}\textbf{of Theorem \ref{theo:asymptotic_spikefree}}
Given a vector $\vec{u}$ we partition $\data$ according to the subset $S=\{i|\sample_i^T\vec{u}\ge 0\}$, where $\data_S \vec{u}\succcurlyeq 0$ and $-\data_{S^c}\vec{u}\succcurlyeq 0$ into
\begin{align*}
    \data = \left[\begin{array}{c}
        \data_S\\
        \data_{S^c}
    \end{array}\right]\,.
\end{align*}
Here $\data_S$ is the sub-matrix of $\data$ consisting of the rows indexed by $S$, and $S^c$ is the complement of the set $S$. Consequently, we partition the vector $\relu{\data\vec{u}}$ as follows
\begin{align*}
    \relu{\data \vec{u}} = \left[\begin{array}{c}
        \data_S \vec{u} \\
        \vec{0}\
    \end{array}\right]\,.
\end{align*}
Then we use the block matrix pseudo-inversion formula \citep{baksalary2007particular}
\begin{align*}
    \data^{\dagger} = \left[\begin{array}{c c}
        \left(\data_S \vec{P}_{S^c}^\perp\right)^\dagger & \left(\data_{S^c} \vec{P}_{S}^\perp\right)^\dagger
    \end{array}\right]\,,
\end{align*}
where $\vec{P}_{S}$ and $\vec{P}_{S^c}$ are projection matrices defined as follows
\begin{align*}
    \vec{P}_{S} &= \vec{I}_d - \data_S^T \big(\data_S\data_S^T\big)^{-1} \data_S\\
    \vec{P}_{S^c} &= \vec{I}_d - \data_{S^c}^T \big(\data_{S^c}\data_{S^c}^T\big)^{-1} \data_{S^c}\,.
\end{align*}
Note that the matrices $\data_S\data_S^T\in\reals^{|S|\times |S|}$, $\data_{S^c}\data_{S^c}^T\in\reals^{|S^{c}|\times |S^{c}|}$ are full column rank with probability one since the matrix $\data \in\reals^{n\times d}$ is i.i.d. Gaussian where $n<d$. Hence the inverses $\big(\data_S\data_S^T\big)^{-1}$ and $\big(\data_{S^c}\data_{S^c}^T\big)^{-1}$ exist with probability one. Plugging in the above representation in the spike-free condition we get
\begin{align*}
    \data^\dagger \relu{\data \vec{u}} = \left(\data_S \vec{P}_{S^c}^\perp\right)^\dagger \data_S\vec{u}\,.
\end{align*}
Then we can express the probability of the matrix being spike-free as
\begin{align*}
    \mathbb{P} \left[ \max_{\vec{u}\in\ball_2} \|\data^\dagger \relu{\data \vec{u}}\|_2 >1 \right]
    & = \mathbb{P} \left[ \exists \vec{u}\in \ball_2~|~ \|\data^\dagger \relu{\data \vec{u}}\|_2 >1 \right]\\
    &\le \mathbb{P} \left[ \exists \vec{u}\in \ball_2, S \subseteq [n]~|~ \|\left(\data_S \vec{P}_{S^c}^\perp\right)^\dagger \data_S\vec{u}\|_2 >1 \right]\,.
\end{align*}
Finally, observe that $\vec{P}_{S^{c}}\in\reals^{d \times d}$ is a uniformly random projection matrix of subspace of dimension $|S|\le n$. Therefore as $d\rightarrow \infty$, we have $\vec{P}_{S^{c}}^\perp \rightarrow \vec{I}_d$, and consequently
\begin{align*}
     \lim_{d\rightarrow \infty} \|\left(\data_S \vec{P}_{S^c}^\perp\right)^\dagger \data_S\vec{u}\|_2 = \|\data_S^\dagger \data_S\vec{u}\|_2\,,
\end{align*}
with probability one, and we have
\begin{align*}
    \lim_{d\rightarrow \infty} 
     \mathbb{P} \left[ \exists \vec{u}\in \ball_2, S \subseteq [n]~|~ \|\left(\data_S \vec{P}_{S^c}^\perp\right)^\dagger \data_S\vec{u}\|_2 >1 \right]  = 0\,.
\end{align*}
\end{proof}

\begin{proof}\textbf{of Theorem \ref{theo:convex_hull}}
Since each sample $\sample_j$ is a vertex of $\mathcal{C}_a$, we can find a separating hyperplane defined by the parameters $(\firstw_j,\bias_j)$ so that $\sample_j^T \firstw_j+\bias_j > 0$ and $\sample_i^T \firstw_j+\bias_j \leq 0, \forall i \neq j$. Then, choosing $\{(\firstw_j,\bias_j) \}_{j=1}^n$ yields that $(\data \firstwmat+\vec{1}\biasvec^T)_+$ is a diagonal matrix. Using these hidden neurons, we write the constraint of \eqref{eq:problem_def2} in a more compact form as
\begin{align*}
   (\data \firstwmat +\vec{1}\biasvec^T)_+\secondwvec=\vec{y},
\end{align*}
which is a least squares problem with a full rank square data matrix. Therefore, selecting $\secondwvec=((\data \firstwmat+\vec{1}\biasvec^T)_+ )^{\dagger} \vec{y}$ along with $\firstwmat$ and $\biasvec$ achieves a feasible solution for the original problem, i.e., $0$ training error.
\end{proof}
\begin{proof}\textbf{of Theorem \ref{theo:random_convexhull}}
Let us define the distance of the $i^{th}$ sample vector to the convex hull of the remaining sample vectors as $d_i$
\begin{align*}
d_i \triangleq \min_{\substack{\vec{z} \in \mathbb{R}^n\,: \\ \sum_{j\neq i}z_j=1\\
\vec{z} \succcurlyeq 0}} \|\sample_i - \sum_{j\neq i} \sample_j z_j \|_2 &= \min_{\substack{\vec{z} \in \mathbb{R}^n\,: \\ \sum_{j\neq i}z_j=1\\\vec{z} \succcurlyeq 0,z_i=-1}} \|\data^T \vec{z}\|_2 \\
&= \min_{\substack{\vec{z} \in \mathbb{R}^n\,: \\ \sum_{j\neq i}z_j=1\\\vec{z} \succcurlyeq 0,z_i=-1}} \max_{\vec{v}:\|\vec{v}\|_2\le 1} \vec{v}^T \data^T \vec{z}
\end{align*}
Using Gordon's escape from a mesh theorem \citep{gordon1988milman,TalagrandBook}, we obtain the following lower-bound on the expectation of $d_i$
\begin{align}
    \Exs\, d_i &\ge  \Exs \min_{\substack{\vec{z} \in \mathbb{R}^n\,: \nonumber\\ \sum_{j\neq i}z_j=1\\z_j \geq 0,z_i=-1}} \max_{\vec{v}:\|\vec{v}\|_2\le 1} \vec{h}^T\vec{v} \|\vec{z}\|_2 +  \vec{z}^T \vec{g} \nonumber\\
    &=\Exs \min_{\substack{\vec{z} \in \mathbb{R}^n\,: \\ \sum_{j\neq i}z_j=1\\ z_j \geq 0,z_i=-1}} ||\vec{h}||_2\|\vec{z}\|_2 +  \vec{z}^T \vec{g} \nonumber\\
    & \ge \sqrt{d}\|\vec{z}\|_2 - \Exs \max_{j\in[n],j\neq i} g_j + g_i \nonumber \\
    &\ge \frac{\sqrt{d}}{\sqrt{n}} - \sqrt{2\log(n-1)}\,, \label{eq:distlowerbound}
\end{align}

where $\vec{h}\in\mathbb{R}^d$ and $\vec{g} \in \mathbb{R}^n$ are random vectors with i.i.d. standard Gaussian components, and the second inequality follows from a well-known result on finite Gaussian suprema \citep{TalagrandBook}. Therefore, the expected distance of the $i^{th}$ sample to the convex hull is guaranteed to be positive whenever $d>2n\log(n-1)$. Note that the lower bound \eqref{eq:distlowerbound} is vacuous for $d<2n\log(n-1)$ since the random variable $d_i$ can only take non-negative values.

The distance $d_i$ is a Lipschitz function of the random Gaussian matrix $\data$. This can be seen via the following argument
\begin{align*}
\min_{\substack{\vec{z} \in \mathbb{R}^n\,: \\ \sum_{j\neq i}z_j=1\\ z_j \geq 0,z_i=-1}} \|\data^T\vec{z}\|_2 - \min_{\substack{\vec{z} \in \mathbb{R}^n\,: \\ \sum_{j\neq i}z_j=1\\z_j \geq 0,z_i=-1}} \|\tilde \data^T\vec{z}\|_2 &\le \min_{\substack{\vec{z} \in \mathbb{R}^n\,: \\ \sum_{j\neq i}z_j=1\\z_j \geq 0,z_i=-1}} \|\big(\data-\tilde\data^T\big)\vec{z}\|_2 \\
& \le \|(\data- \tilde \data)\|_2  \max_{\substack{\vec{z} \in \mathbb{R}^n\,: \\ \sum_{j\neq i}z_j=1\\z_j \geq 0,z_i=-1}} \|\vec{z}\|_2\\
& \le \|(\data- \tilde \data)\|_F  \max_{\substack{\vec{z} \in \mathbb{R}^n\,: \\ \sum_{j\neq i}z_j=1\\z_j \geq 0,z_i=-1}} \|\vec{z}\|_1\\
& \le 2\|(\data- \tilde \data)\|_F 
\end{align*}Applying the Lipschitz concentration for Gaussian measure \citep{TalagrandBook} yields that 
\begin{align*}
    \mathbb{P}\big[d_i > \sqrt{d}-\sqrt{2n\log(n-1)} -t \big] \ge 1-2e^{-t^2/2}\,.
\end{align*}
Therefore, we have $d_i>0$ for $d>2n\log(n-1)$ with probability exceeding $1-2e^{-t^2/2}$. Taking a union bound over every index $i \in \{0,...,n\}$, we can upper-bound the failure probability by $2ne^{-t^2/2}$. Choosing $t^2=4\log(n-1)$ will yield a failure probability $O(1/n)$ and conclude the proof.
\end{proof}


\subsection{Proofs for the results in Section \ref{sec:main_regconvex}}

\begin{proof}\textbf{of Theorem \ref{theo:regularized_dual}}
 The equivalence of weight decay and $\ell_1$ regularized problems in the theorem statement follows from the scaling argument as in the proofs of Lemma \ref{lemma:reg_equivalence} and \ref{lemma:constraint}. The rest of the proof follows a similar argument as in the proof of Theorem \ref{theo:equality_dual}. We reparameterize \eqref{eq:problem_def_regularized} as follows
\begin{align*}
    \min_{\vec{r},\theta \in \Theta} \frac{1}{2}\|\vec{r}\|_2^2 + \beta \| \secondwvec \|_1 \text{ s.t. } \vec{r}=(\data \firstwmat )_+ \secondwvec -\vec{y}, \;\| \firstw_j \|_2\leq1 , \forall j.
\end{align*}
Then taking the convex dual of the above problem with respect to the second layer weights yields the claimed form.
\end{proof}

\begin{proof}{\bf of Theorem \ref{theo:closedform_regularized}}
 Let us first restate the dual problem as 
\begin{align}\label{eq:dual_regularized1}
    &\max_{\dual}  - \frac{1}{2}\|\dual-\vec{y}\|_2^2+\frac{1}{2} \| \vec{y}\|_2^2 \mbox{ s.t. } \max_{\firstw \in \ball_2} \vert \dual^T \relu{\data \firstw}\vert \leq \beta.
\end{align}
Since $\data$ is whitened such that $\data\data^T=\vec{I}_n$, \eqref{eq:dual_regularized1} can be rewritten as follows
\begin{align*}
    &\max_{\dual}  - \frac{1}{2}\|\dual-\vec{y}\|_2^2+\frac{1}{2} \| \vec{y}\|_2^2 \mbox{ s.t. } \max\left\{\|\relu{\dual}\|_2,\|\relu{-\dual}\|_2 \right\}\leq \beta.
\end{align*}
The problem above has a closed-form solution in the following form
\begin{align*}
    \dual^*= \begin{cases}
    \beta\frac{\relu{\vec{y}}}{\|\relu{\vec{y}}\|_2}-\beta\frac{\relu{-\vec{y}}}{\|\relu{-\vec{y}}\|_2} &\text{ if } \beta \leq \|\relu{\vec{y}}\|_2,\;  \beta \leq  \|\relu{\vec{-y}}\|_2\\
     \hfil \relu{\vec{y}}-\beta\frac{\relu{-\vec{y}}}{\|\relu{-\vec{y}}\|_2} &\text{ if } \beta > \|\relu{\vec{y}}\|_2 , \;  \beta \leq \relu{\vec{-y}}\|_2\\
     \hfil \beta\frac{\relu{\vec{y}}}{\|\relu{\vec{y}}\|_2}-\relu{-\vec{y}} &\text{ if } \beta \leq \|\relu{\vec{y}}\|_2,\;  \beta >  \|\relu{\vec{-y}}\|_2\\
     \hfil \vec{y} &\text{ if } \beta > \|\relu{\vec{y}}\|_2,\;  \beta >  \|\relu{\vec{-y}}\|_2
    \end{cases}
\end{align*}
and the corresponding extreme points for the two-sided constraint in \eqref{eq:dual_regularized1} are
\begin{align*}
    \firstwmat^*= \begin{cases}
   \left( \frac{\data^\dagger \relu{\vec{y}}}{\|\data^\dagger \relu{\vec{y}}\|_2}, \frac{\data^\dagger \relu{\vec{-y}}}{\|\data^\dagger \relu{\vec{-y}}\|_2}\right) &\text{ if } \beta \leq \|\relu{\vec{y}}\|_2,\;  \beta \leq  \|\relu{\vec{-y}}\|_2\\
     \hfil  \frac{\data^\dagger \relu{\vec{-y}}}{\|\data^\dagger \relu{\vec{-y}}\|_2} &\text{ if } \beta > \|\relu{\vec{y}}\|_2 , \;  \beta \leq \relu{\vec{-y}}\|_2\\
      \hfil \frac{\data^\dagger \relu{\vec{y}}}{\|\data^\dagger \relu{\vec{y}}\|_2} &\text{ if } \beta \leq \|\relu{\vec{y}}\|_2,\;  \beta >  \|\relu{\vec{y}}\|_2\\
    \hfil  \vec{0} &\text{ if } \beta > \|\relu{\vec{y}}\|_2,\;  \beta >  \|\relu{\vec{-y}}\|_2\
    \end{cases}.
\end{align*}
Now, we first substitute each case into the primal problem \ref{eq:problem_def_regularized} and then take the derivative with respect to the output layer weights $\secondwvec$. Since this is a linear unconstrained least squares optimization problem, which is convex, we obtain the following closed-form solutions for the output layer weights

\begin{align*}
    \secondwvec^*= \begin{cases}
   \begin{bmatrix} \|\data^\dagger \relu{\vec{y}}\|_2-\beta\\-\|\data^\dagger \relu{\vec{-y}}\|_2+\beta\end{bmatrix} &\text{ if } \beta \leq \|\relu{\vec{y}}\|_2,\;  \beta \leq  \|\relu{\vec{-y}}\|_2\\
      \hfil -\|\data^\dagger \relu{\vec{-y}}\|_2+\beta &\text{ if } \beta > \|\relu{\vec{y}}\|_2 , \;  \beta \leq \|\relu{\vec{-y}}\|_2\\
     \hfil \|\data^\dagger \relu{\vec{y}}\|_2-\beta &\text{ if } \beta \leq \|\relu{\vec{y}}\|_2,\;  \beta >  \|\relu{\vec{-y}}\|_2\\
    \hfil 0 &\text{ if } \beta > \|\relu{\vec{y}}\|_2,\;  \beta >  \|\relu{\vec{-y}}\|_2
    \end{cases}.
\end{align*}

\end{proof}


\begin{proof}\textbf{of Theorem \ref{theo:hinge_dual}}
The proof follows from a similar argument as in the proof of Theorem \ref{theo:equality_dual} and \ref{theo:regularized_dual}. We first put \eqref{eq:problem_def_hinge} into the following form
\begin{align*}
&\min_{\vec{\xi}, \theta \in \Theta} \sum_{i=1}^n \xi_i + \beta \| \secondwvec \|_1  \text{ s.t. } {\xi}_{i} \geq 0, \; \xi_i \geq 1-y_i(\sample_i^T\firstwmat)_{+}\secondwvec, \forall i ,\;\| \firstw_j \|_2\leq1 , \forall j.
\end{align*}
Then taking the dual of the above problem yields the claimed form.
\end{proof}


\begin{proof}{\bf of Theorem \ref{theo:closedform_2neuron_hinge}}
Since $\data$ is whitened, as a direct consequence of Theorem \ref{theo:closedform_regularized} and \ref{theo:hinge_dual}, we rewrite the dual problem as
\begin{align} \label{eq:hinge_dual_regularized1}
    &\max_{\dual}  \dual^T \vec{y} \mbox{ s.t. } \max{\|\relu{\dual}\|_2,\|\relu{-\dual}\|_2}\leq \beta,
\end{align}
which has the following optimal solution
\begin{align*}
    \dual^*=\beta \frac{\relu{\vec{y}}}{\|\relu{\vec{y}}\|_2}-\beta \frac{\relu{-\vec{y}}}{\|\relu{-\vec{y}}\|_2}
\end{align*}
and the corresponding extreme points are
\begin{align*}
    \firstwmat^*&= \begin{bmatrix} \frac{\data^\dagger \relu{\vec{y}}}{\|\data^\dagger \relu{\vec{y}}\|_2} & \frac{\data^\dagger \relu{-\vec{y}}}{\|\data^\dagger \relu{-\vec{y}}\|_2} \end{bmatrix}=\begin{bmatrix} \frac{\data^\dagger \relu{\vec{y}}}{\sqrt{n_+}} & \frac{\data^\dagger \relu{-\vec{y}}}{\sqrt{n_-}} \end{bmatrix}
\end{align*}
where $n_+$ and $n_-$ are the number of samples with positive and negative labels, respectively and the second equality follows from $y_i \in\{\pm1\}$. If we substitute $\firstwmat^*$ into \eqref{eq:problem_def_hinge} and take derivative of the objective with respect to the output layer weight $\secondwvec$, we obtain the following optimality conditions
\begin{align}\label{eq:hinge_regularized_optimality}
\begin{split}
      &0 \in \partial \max\left\{0,\sqrt{n_+}-\secondw_1\right\}\sqrt{n_+}+\beta \partial \vert \secondw_1 \vert \\
      &0 \in \partial \max\left\{0,\sqrt{n_-}+\secondw_2\right\}\sqrt{n_-}+\beta \partial \vert \secondw_2 \vert 
\end{split}\;.
\end{align}
Therefore, the optimal solutions are
\begin{align*}
    \secondw_1=\begin{cases} \secondw \in [0,\sqrt{n_+}] &\text{ if } \beta = \sqrt{n_+} \\
  \hfil \sqrt{n_+} &\text{ if } \beta < \sqrt{n_+} \\
    \hfil 0 &\text{ if } \beta > \sqrt{n_+}
    \end{cases}\;, \quad     \secondw_2=\begin{cases} \secondw \in [-\sqrt{n_-},0] &\text{ if } \beta = \sqrt{n_-} \\
  \hfil -\sqrt{n_-} &\text{ if } \beta < \sqrt{n_-} \\
    \hfil 0 &\text{ if } \beta > \sqrt{n_-}
    \end{cases}\;.
\end{align*}

\end{proof}


\begin{proof}\textbf{of Theorem \ref{theo:generic_dual}}
The proof follows from classical Fenchel duality \citep{boyd_convex}, and a similar argument as in the proof of Theorem \ref{theo:equality_dual} and \ref{theo:regularized_dual}. We first describe \eqref{eq:gen_loss} in an equivalent form as follows
\begin{align*}
    \min_{\vec{z},\theta \in \Theta} \ell(\vec{z},\vec{y}) + \beta \| \secondwvec \|_1 \text{ s.t. } \vec{z}=(\data\firstwmat)_{+}\secondwvec, \; \| \firstw_j \|_2\leq1 , \forall j.
\end{align*}
Then the dual function is
\begin{align*}
    g(\dual)= \min_{\vec{z},\theta \in \Theta} \ell(\vec{z},\vec{y})- \dual^T \vec{z}+ \dual^T (\data\firstwmat)_{+}\secondwvec+ \beta \|\secondwvec \|_1\text{ s.t. } \| \firstw_j \|_2\leq1 , \forall j.
\end{align*}
Therefore, using the classical Fenchel duality \citep{boyd_convex} yields the proposed dual form.
\end{proof}


\begin{proof}\textbf{of Lemma \ref{lemma:reg_equivalence_multiclass}}
For any $\theta \in \Theta$, we can rescale the parameters as $\bar{\firstw}_j=\alpha_j\firstw_j$ and $\bar{\secondwvec}_j= \secondwvec_j/\alpha_j$, for any $\alpha_j>0$, where $\firstw_j$ and $\secondwvec_j$ are the $j^{th}$ column and row of $\firstwmat$ and $\secondwmat$, respectively. Then, \eqref{eq:2layer_function} becomes
\begin{align*}
    f_{\bar{\theta}}(\data)=\sum_{j=1}^m  (\data \bar{\firstw}_j)_{+}\bar{\secondwvec}_j^T=\sum_{j=1}^m  (\alpha_j \data \firstw_j)_{+}\frac{\secondwvec_j^T}{\alpha_j}=\sum_{j=1}^m (\data \firstw_j)_{+}\secondwvec_j^T,
\end{align*}
which proves $f_{\theta}(\data)=f_{\bar{\theta}}(\data)$. In addition to this, we have the following basic inequality
\begin{align*}
    \sum_{j=1}^m (\|\secondwvec_j\|_2^2+\| \firstw_j\|_2^2) \geq 2\sum_{j=1}^m (\|\secondwvec_j\|_2 \text{ }\| \firstw_j\|_2),
\end{align*}
where the equality is achieved with the scaling choice $\alpha_j=\big(\frac{\|\secondwvec_j\|_2}{\| \firstw_j\|_2}\big)^{\frac{1}{2}}$. Since the scaling operation does not change the right-hand side of the inequality, we can set $\|\firstw_j \|_2=1, \forall j$. Therefore, the right-hand side becomes $\sum_{j=1}^m \|\secondwvec_j\|_2 $.
\end{proof}


\begin{proof}\textbf{of Corollary \ref{cor:1d_extreme_multiclass}}
The proof directly follows from the proof of Corollary \ref{cor:1d_optimality}.
\end{proof}


\begin{proof}\textbf{of Corollary \ref{cor:rankone_extreme_multiclass}}
The proof directly follows from the proof of Corollary \ref{cor:rankone_extreme}.
\end{proof}


\begin{proof}\textbf{of Theorem \ref{theo:rankone_multiclass}}
Proof directly follows from Corollary \ref{cor:1d_extreme_multiclass} and \ref{cor:rankone_extreme_multiclass}.
\end{proof}


\begin{proof}{\bf of Theorem \ref{theo:closedform_regularized_multiclass}}
We first apply the scaling in Lemma \ref{lemma:reg_equivalence_multiclass} and then restate the dual problem as 
\begin{align}\label{eq:dual_regularized_multiclass1}
    &\max_{\dualmat}  - \frac{1}{2}\|\dualmat-\vec{Y}\|_F^2+\frac{1}{2} \| \vec{Y}\|_F^2 \mbox{ s.t. } \max_{\firstw \in \ball_2} \| \dualmat^T \relu{\data \firstw}\|_2 \leq \beta.
\end{align}
Since $\data$ is whitened such that $\data\data^T=\vec{I}_n$ and $\vec{Y}$ is one hot encoded, \eqref{eq:dual_regularized_multiclass1} can be rewritten as follows
\begin{align}\label{eq:dual_regularized_multiclass2}
    &D:=\max_{\dualmat}  - \frac{1}{2}\|\dualmat-\vec{Y}\|_F^2+\frac{1}{2} \| \vec{Y}\|_F^2 \mbox{ s.t. }  \max_{\firstw \in \ball_2} \|\dualmat^T \firstw\|_2\leq \beta.
\end{align}
The problem above has a closed-form solution as follows
\begin{align}\label{eq:multiclass_dualparam}
   \dual_j^*= \begin{cases}
    \beta \frac{\vec{y}_j}{\|\vec{y}_j\|_2} &\text{ if } \beta \leq \|\vec{y}_j\|_2\\\
    \hfil\vec{y}_j  &\text{ otherwise } 
    \end{cases},\quad \forall j \in [o].
\end{align}
and the corresponding extreme points of the constraint in \eqref{eq:dual_regularized_multiclass2} are
\begin{align}\label{eq:multiclass_extremepoints}
   \firstw_j^*= \begin{cases}
    \frac{\data^\dagger \vec{y}_j}{\|\data^\dagger \vec{y_j}\|_2} &\text{ if } \beta \leq \|\vec{y}_j\|_2\\\
    \hfil -  &\text{ otherwise } 
    \end{cases},\quad \forall j \in [o].
\end{align}
Now let us first denote the set of indices that yield an extreme point as $\mathcal{E}:=\{j:\beta \leq \|\vec{y}_j\|_2, j \in [o]\}$. Then we compute the objective value for the dual problem in \eqref{eq:dual_regularized_multiclass2} using the optimal parameter in \eqref{eq:multiclass_dualparam}
\begin{align}\label{eq:dual_obj_multiclass2} \nonumber
   D&= - \frac{1}{2}\|\dualmat^*-\vec{Y}\|_F^2+\frac{1}{2} \| \vec{Y}\|_F^2 \nonumber\\
   &=- \frac{1}{2} \sum_{j\in \mathcal{E}} (\beta-\|\vec{y}_j\|_2)^2+\frac{1}{2}\sum_{j=1}^o \|\vec{y}_j\|_2^2 \nonumber\\
   &=-\frac{1}{2} \beta^2 |\mathcal{E}|+\beta\sum_{j \in \mathcal{E}}\|\vec{y}_j\|_2 +\frac{1}{2}\sum_{j \notin \mathcal{E}} \|\vec{y}_j\|_2^2.
\end{align}
Next, we restate the primal problem as follows 
\begin{align}\label{eq:primal_regularized_multiclass2}
    &P:=\max_{\firstwmat,\secondwmat}   \frac{1}{2}\|\relu{\data \firstwmat}\secondwmat-\vec{Y}\|_F^2+\beta \sum_{j=1}^{o} \| \secondwvec_j\|_2 \mbox{ s.t. } \|\firstw_j\|\leq 1, \forall j \in [o],
\end{align}
and then solve it using the optimal hidden layer weights in \eqref{eq:multiclass_extremepoints}, which  yields the following optimal solution 
\begin{align}\label{eq:multiclass_primalparam}
   \left(\firstw_j^* ,\secondwvec_j^*\right)= \begin{cases}
   \left(\frac{\data^\dagger \vec{y}_j}{\|\data^\dagger \vec{y_j}\|_2}, \left(\|\vec{y}_j\|_2-\beta\right)\vec{e}_j\right) &\text{ if } \beta \leq \|\vec{y}_j\|_2\\\
    \hfil - &\text{ otherwise } 
    \end{cases},\quad \forall j \in [o].
\end{align}
Evaluating the primal problem objective with the parameters \eqref{eq:multiclass_primalparam} gives
\begin{align}\label{eq:primal_obj_multiclass2}
    P&= \frac{1}{2}\|\relu{\data \firstwmat^*}\secondwmat^*-\vec{Y}\|_F^2+\beta \sum_{j\in \mathcal{E}} \| \secondwvec_j^*\|_2 \nonumber\\
    &=\frac{1}{2} \left\|\sum_{j\in \mathcal{E}}\left(\|\vec{y}_j\|_2-\beta\right)\frac{\vec{y}_j}{\|\vec{y}_j\|_2}\vec{e}_j^T-\vec{Y} \right\|_F^2+\beta \sum_{j\in \mathcal{E}} (\|\vec{y}_j\|_2 -\beta)\nonumber\\
     &=\frac{1}{2}\sum_{j\in \mathcal{E}} \left\|\beta\frac{\vec{y}_j}{\|\vec{y}_j\|_2}\vec{e}_j^T\ \right\|_F^2+\frac{1}{2}\sum_{j\notin \mathcal{E}} \|\vec{y}_j\vec{e}_j^T\|_F^2+\beta \sum_{j\in \mathcal{E}} \|\vec{y}_j\|_2 -\beta^2 |\mathcal{E}|\nonumber\\
      &=-\frac{1}{2}\beta^2 |\mathcal{E}|+\frac{1}{2}\sum_{j\notin \mathcal{E}} \|\vec{y}_j\|_2^2+\beta \sum_{j\in \mathcal{E}} \|\vec{y}_j\|_2 ,
\end{align}
which has the same value with \eqref{eq:dual_obj_multiclass2}. Therefore, strong duality holds, i.e., $P=D$, and the set of weight in \eqref{eq:multiclass_primalparam} are optimal for the primal problem \eqref{eq:primal_regularized_multiclass2}. We also note that since $\vec{Y}$ is one hot encoded, \eqref{eq:multiclass_primalparam} can be equivalently stated as
\begin{align}\label{eq:multiclass_primalparam_v2}
   \left(\firstw_j^* ,\secondwvec_j^*\right)= \begin{cases}
   \left(\frac{\data^\dagger \vec{y}_j}{\|\data^\dagger \vec{y_j}\|_2}, \left(\sqrt{n_j}-\beta\right)\vec{e}_j\right) &\text{ if } \beta \leq \sqrt{n_j}\\\
    \hfil - &\text{ otherwise } 
    \end{cases},\quad \forall j \in [o],
\end{align}
where $n_j$ is the number samples in the $j^{th}$ class.
\end{proof}


\begin{proof}\textbf{of Lemma \ref{lemma:l1_multiclass}}
The proof is a straightforward generalization of the scalar output case in Lemma \ref{lemma:closedform_2neuron}.
\end{proof}


\begin{proof}\textbf{of Lemma \ref{lemma:l1_l0_multiclass}}
The proof is a straightforward generalization of the scalar output case in Lemma \ref{lemma:l1-l0 equivalence}.
\end{proof}

\section{Polar convex duality}\label{sec:polar_duality_appendix}
In this section we derive the polar duality and present a connection to minimum $\ell_1$ solutions to linear systems. 
 Recognizing the constraint $\dual \in \rectset$ can be stated as
\begin{align*}
    \dual\in \rectset^\circ,\, \dual\in -\rectset^\circ\,,
\end{align*}
which is equivalent to 
\begin{align*}
    \dual\in \rectset^\circ \cap -\rectset^\circ\,.
\end{align*}

\noindent Note that the support function of a set can be expressed as the gauge function of its polar set (see e.g. \citet{Rockafellar}). The polar set of $\rectset^\circ \cap -\rectset^\circ$ is given by
\begin{align*}
    \big(\rectset^\circ \cap -\rectset^\circ\big)^\circ = \conv\{\rectset \cup -\rectset\}\,.
\end{align*}
Using this fact, we express the dual problem \eqref{eq:problem_dual} as
\begin{align}\label{eq:gaugeform}
    D^* = &\inf_{t\in\reals}\, t \\ \nonumber
    &\mbox{s.t. } \vec{y}\in t \conv\big\{\rectset \cup -\rectset\big\}\,,
\end{align}
where $\conv$ represents the convex hull of a set.

Let us restate dual of the two-layer ReLU NN training problem given by
\begin{align}
    &\max_{\dual} \dual^T \vec{y} \mbox{ s.t. } \dual \in \rectset^\circ~ \mbox{,       }-\dual \in \rectset^\circ \label{eq:dualpolar_sup}
\end{align}
where $\rectset^\circ$ is the polar dual of $\rectset$ defined as
$    \rectset^\circ = \{\dual | \dual^T \firstw\le 1\, \forall \firstw\in \rectset\}\,.
$
\brems
The dual problem given in \eqref{eq:dualpolar_sup} is analogous to the convex duality in minimum $\ell_1$-norm solutions to linear systems. In particular, for the latter it holds that
\begin{align*}
    \min_{\secondwvec\,:\,\data\secondwvec=\vec{y}}\|\secondwvec\|_1 = \max_{\dual \in \conv\{\hat{\vec{a}}_1,...,\hat{\vec{a}}_d\}^\circ,~ -\dual \in \conv\{\hat{\vec{a}}_1,...,\hat{\vec{a}}_d\}^\circ} \dual^T \vec{y} \,,
\end{align*}
where $\hat{\vec{a}}_1,...,\hat{\vec{a}}_d$ are the columns of $\data$.
The above optimization problem can also be put in the gauge optimization form as follows.
\begin{align*}
    \min_{\secondwvec\,:\,\data\secondwvec=\vec{y}}\|\secondwvec\|_1 = \inf_{t\in\reals} \, t \mbox{ s.t. }  \vec{y} \in t \conv\{\pm \hat{\vec{a}}_1,..., \pm\hat{\vec{a}}_d \},
\end{align*}
which parallels the gauge optimization form in \eqref{eq:gaugeform}.
\erems

\bibliography{references}

\begin{thebibliography}{67}
\providecommand{\natexlab}[1]{#1}
\providecommand{\url}[1]{\texttt{#1}}
\expandafter\ifx\csname urlstyle\endcsname\relax
  \providecommand{\doi}[1]{doi: #1}\else
  \providecommand{\doi}{doi: \begingroup \urlstyle{rm}\Url}\fi

\bibitem[new()]{news20}
20 newsgroups.
\newblock \url{http://qwone.com/~jason/20Newsgroups/}.

\bibitem[Arora et~al.(2018)Arora, Cohen, and
  Hazan]{arora2018overparameterization}
Sanjeev Arora, Nadav Cohen, and Elad Hazan.
\newblock On the optimization of deep networks: Implicit acceleration by
  overparameterization.
\newblock In \emph{35th International Conference on Machine Learning, ICML
  2018}, pages 372--389. International Machine Learning Society (IMLS), 2018.

\bibitem[Arora et~al.(2019)Arora, Du, Hu, Li, Salakhutdinov, and
  Wang]{arora_cntk}
Sanjeev Arora, Simon~S. Du, Wei Hu, Zhiyuan Li, Ruslan Salakhutdinov, and
  Ruosong Wang.
\newblock On exact computation with an infinitely wide neural net.
\newblock \emph{CoRR}, abs/1904.11955, 2019.
\newblock URL \url{http://arxiv.org/abs/1904.11955}.

\bibitem[Bach(2017)]{bach2017breaking}
Francis Bach.
\newblock Breaking the curse of dimensionality with convex neural networks.
\newblock \emph{The Journal of Machine Learning Research}, 18\penalty0
  (1):\penalty0 629--681, 2017.

\bibitem[Baksalary and Baksalary(2007)]{baksalary2007particular}
Jerzy~K Baksalary and Oskar~Maria Baksalary.
\newblock Particular formulae for the moore--penrose inverse of a columnwise
  partitioned matrix.
\newblock \emph{Linear algebra and its applications}, 421\penalty0
  (1):\penalty0 16--23, 2007.

\bibitem[Bartan and Pilanci(2019)]{bartan2019convex}
Burak Bartan and Mert Pilanci.
\newblock Convex relaxations of convolutional neural nets.
\newblock In \emph{ICASSP 2019-2019 IEEE International Conference on Acoustics,
  Speech and Signal Processing (ICASSP)}, pages 4928--4932. IEEE, 2019.

\bibitem[Bengio et~al.(2006)Bengio, Roux, Vincent, Delalleau, and
  Marcotte]{bengio2006convex}
Yoshua Bengio, Nicolas~L Roux, Pascal Vincent, Olivier Delalleau, and Patrice
  Marcotte.
\newblock Convex neural networks.
\newblock In \emph{Advances in neural information processing systems}, pages
  123--130, 2006.

\bibitem[Bietti and Mairal(2019)]{ntk_relu_1d}
Alberto Bietti and Julien Mairal.
\newblock On the inductive bias of neural tangent kernels.
\newblock \emph{arXiv preprint arXiv:1905.12173}, 2019.

\bibitem[Blanc et~al.(2019)Blanc, Gupta, Valiant, and
  Valiant]{implicit_reg_blanc}
Guy Blanc, Neha Gupta, Gregory Valiant, and Paul Valiant.
\newblock Implicit regularization for deep neural networks driven by an
  ornstein-uhlenbeck like process.
\newblock \emph{CoRR}, abs/1904.09080, 2019.
\newblock URL \url{http://arxiv.org/abs/1904.09080}.

\bibitem[Boyd and Vandenberghe(2004)]{boyd_convex}
Stephen Boyd and Lieven Vandenberghe.
\newblock \emph{Convex optimization}.
\newblock Cambridge university press, 2004.

\bibitem[Brutzkus et~al.(2017)Brutzkus, Globerson, Malach, and
  Shalev{-}Shwartz]{brutzkus_overparameterized_linear}
Alon Brutzkus, Amir Globerson, Eran Malach, and Shai Shalev{-}Shwartz.
\newblock {SGD} learns over-parameterized networks that provably generalize on
  linearly separable data.
\newblock \emph{CoRR}, abs/1710.10174, 2017.
\newblock URL \url{http://arxiv.org/abs/1710.10174}.

\bibitem[Candes and Tao(2005)]{CandesTao05}
E.~J. Candes and T.~Tao.
\newblock Decoding by linear programming.
\newblock \emph{IEEE Trans. Info Theory}, 51\penalty0 (12):\penalty0
  4203--4215, December 2005.

\bibitem[Chizat and Bach(2018)]{lazy_training_bach}
Lenaic Chizat and Francis Bach.
\newblock A note on lazy training in supervised differentiable programming.
\newblock \emph{arXiv preprint arXiv:1812.07956}, 2018.

\bibitem[Coates and Ng(2012)]{kmenas_andrewng}
Adam Coates and Andrew~Y Ng.
\newblock Learning feature representations with k-means.
\newblock In \emph{Neural networks: Tricks of the trade}, pages 561--580.
  Springer, 2012.

\bibitem[Ding et~al.(2008)Ding, Li, and Jordan]{semi_nmf}
Chris~HQ Ding, Tao Li, and Michael~I Jordan.
\newblock Convex and semi-nonnegative matrix factorizations.
\newblock \emph{IEEE transactions on pattern analysis and machine
  intelligence}, 32\penalty0 (1):\penalty0 45--55, 2008.

\bibitem[Donoho(2006)]{Donoho04a}
D.~L. Donoho.
\newblock For most large underdetermined systems of linear equations, the
  minimal $\ell_1$-norm solution is also the sparsest solution.
\newblock \emph{Communications on Pure and Applied Mathematics}, 59\penalty0
  (6):\penalty0 797--829, June 2006.

\bibitem[Du and Lee(2018)]{du2018overparameterized}
Simon~S Du and Jason~D Lee.
\newblock On the power of over-parametrization in neural networks with
  quadratic activation.
\newblock \emph{arXiv preprint arXiv:1803.01206}, 2018.

\bibitem[Du et~al.(2018)Du, Zhai, P{\'{o}}czos, and
  Singh]{du_overparameterized}
Simon~S. Du, Xiyu Zhai, Barnab{\'{a}}s P{\'{o}}czos, and Aarti Singh.
\newblock Gradient descent provably optimizes over-parameterized neural
  networks.
\newblock \emph{CoRR}, abs/1810.02054, 2018.
\newblock URL \url{http://arxiv.org/abs/1810.02054}.

\bibitem[Ergen and Pilanci(2019{\natexlab{a}})]{ergen2019convexshallow}
Tolga Ergen and Mert Pilanci.
\newblock Convex optimization for shallow neural networks.
\newblock In \emph{2019 57th Annual Allerton Conference on Communication,
  Control, and Computing (Allerton)}, pages 79--83. IEEE, 2019{\natexlab{a}}.

\bibitem[Ergen and Pilanci(2019{\natexlab{b}})]{ergen2019cutting}
Tolga Ergen and Mert Pilanci.
\newblock Convex duality and cutting plane methods for over-parameterized
  neural networks.
\newblock In \emph{OPT-ML workshop}, 2019{\natexlab{b}}.

\bibitem[Ergen and Pilanci(2020{\natexlab{a}})]{ergen2020aistats}
Tolga Ergen and Mert Pilanci.
\newblock Convex geometry of two-layer relu networks: Implicit autoencoding and
  interpretable models.
\newblock In \emph{International Conference on Artificial Intelligence and
  Statistics}, pages 4024--4033. PMLR, 2020{\natexlab{a}}.

\bibitem[Ergen and Pilanci(2020{\natexlab{b}})]{ergen2020convexdeep}
Tolga Ergen and Mert Pilanci.
\newblock Revealing the structure of deep neural networks via convex duality.
\newblock 2020{\natexlab{b}}.

\bibitem[Ergen and Pilanci(2020{\natexlab{c}})]{ergen2020workshop}
Tolga Ergen and Mert Pilanci.
\newblock Convex programs for global optimization of convolutional neural
  networks in polynomial-time.
\newblock In \emph{OPT-ML workshop}, 2020{\natexlab{c}}.

\bibitem[Ergen and Pilanci(2021)]{ergen2020cnn}
Tolga Ergen and Mert Pilanci.
\newblock Implicit convex regularizers of {\{}cnn{\}} architectures: Convex
  optimization of two- and three-layer networks in polynomial time.
\newblock In \emph{International Conference on Learning Representations}, 2021.
\newblock URL \url{https://openreview.net/forum?id=0N8jUH4JMv6}.

\bibitem[Ergen et~al.(2021)Ergen, Sahiner, Ozturkler, Pauly, Mardani, and
  Pilanci]{ergen2021bn}
Tolga Ergen, Arda Sahiner, Batu Ozturkler, John~M. Pauly, Morteza Mardani, and
  Mert Pilanci.
\newblock Demystifying batch normalization in relu networks: Equivalent convex
  optimization models and implicit regularization.
\newblock \emph{CoRR}, abs/2103.01499, 2021.
\newblock URL \url{https://arxiv.org/abs/2103.01499}.

\bibitem[Frank and Wolfe(1956)]{frank_wolfe}
Marguerite Frank and Philip Wolfe.
\newblock An algorithm for quadratic programming.
\newblock \emph{Naval Research Logistics Quarterly}, 3\penalty0
  (1‐2):\penalty0 95--110, 1956.
\newblock \doi{10.1002/nav.3800030109}.
\newblock URL
  \url{https://onlinelibrary.wiley.com/doi/abs/10.1002/nav.3800030109}.

\bibitem[Fung and Mangasarian(2011)]{fung2011equivalence}
GM~Fung and OL~Mangasarian.
\newblock Equivalence of minimal $\ell_0$-and $\ell_p$-norm solutions of linear
  equalities, inequalities and linear programs for sufficiently small p.
\newblock \emph{Journal of optimization theory and applications}, 151\penalty0
  (1):\penalty0 1--10, 2011.

\bibitem[Goberna and L\'{o}pez-Cerd\'{a}(1998)]{semiinfinite_goberna}
Miguel~Angel Goberna and Marco L\'{o}pez-Cerd\'{a}.
\newblock \emph{Linear semi-infinite optimization}.
\newblock 01 1998.
\newblock \doi{10.1007/978-1-4899-8044-1_3}.

\bibitem[Gordon(1988)]{gordon1988milman}
Yehoram Gordon.
\newblock On milman's inequality and random subspaces which escape through a
  mesh in rn.
\newblock In \emph{Geometric Aspects of Functional Analysis}, pages 84--106.
  Springer, 1988.

\bibitem[Grant and Boyd(2014)]{cvx}
Michael Grant and Stephen Boyd.
\newblock {CVX}: Matlab software for disciplined convex programming, version
  2.1.
\newblock \url{http://cvxr.com/cvx}, March 2014.

\bibitem[Gupta et~al.(2021)Gupta, Bartan, Ergen, and
  Pilanci]{vikul2021generative}
Vikul Gupta, Burak Bartan, Tolga Ergen, and Mert Pilanci.
\newblock Convex neural autoregressive models: Towards tractable, expressive,
  and theoretically-backed models for sequential forecasting and generation.
\newblock In \emph{ICASSP 2021 - 2021 IEEE International Conference on
  Acoustics, Speech and Signal Processing (ICASSP)}, pages 3890--3894, 2021.
\newblock \doi{10.1109/ICASSP39728.2021.9413662}.

\bibitem[Guruswami and Raghavendra(2009)]{guruswami2009hardness}
Venkatesan Guruswami and Prasad Raghavendra.
\newblock Hardness of learning halfspaces with noise.
\newblock \emph{SIAM Journal on Computing}, 39\penalty0 (2):\penalty0 742--765,
  2009.

\bibitem[Huang et~al.(2018)Huang, Yang, Lang, and Deng]{huang2018decorrelated}
Lei Huang, Dawei Yang, Bo~Lang, and Jia Deng.
\newblock Decorrelated batch normalization.
\newblock 2018.

\bibitem[Jacot et~al.(2018)Jacot, Gabriel, and Hongler]{ntk_jacot}
Arthur Jacot, Franck Gabriel, and Cl{\'e}ment Hongler.
\newblock Neural tangent kernel: Convergence and generalization in neural
  networks.
\newblock In \emph{Advances in neural information processing systems}, pages
  8571--8580, 2018.

\bibitem[Krizhevsky et~al.(2014)Krizhevsky, Nair, and Hinton]{cifar10}
Alex Krizhevsky, Vinod Nair, and Geoffrey Hinton.
\newblock The {CIFAR}-10 dataset.
\newblock \url{http://www. cs. toronto. edu/kriz/cifar. html}, 2014.

\bibitem[Krogh and Hertz(1992)]{weightdecay_krogh}
Anders Krogh and John~A Hertz.
\newblock A simple weight decay can improve generalization.
\newblock In \emph{Advances in neural information processing systems}, pages
  950--957, 1992.

\bibitem[Lacotte and Pilanci(2020)]{lacotte2020all}
Jonathan Lacotte and Mert Pilanci.
\newblock All local minima are global for two-layer relu neural networks: The
  hidden convex optimization landscape.
\newblock \emph{arXiv preprint arXiv:2006.05900}, 2020.

\bibitem[LeCun()]{mnist}
Yann LeCun.
\newblock The {MNIST} database of handwritten digits.
\newblock \url{http://yann. lecun. com/exdb/mnist/}.

\bibitem[Lederer(2020)]{lederer2020spurious}
Johannes Lederer.
\newblock No spurious local minima: on the optimization landscapes of wide and
  deep neural networks, 2020.

\bibitem[Ledoux and Talagrand(2013)]{TalagrandBook}
Michel Ledoux and Michel Talagrand.
\newblock \emph{Probability in Banach Spaces: isoperimetry and processes}.
\newblock Springer Science \& Business Media, 2013.

\bibitem[Lee et~al.(2017)Lee, Bahri, Novak, Schoenholz, Pennington, and
  Sohl-Dickstein]{gaussian_infinite2}
Jaehoon Lee, Yasaman Bahri, Roman Novak, Samuel~S Schoenholz, Jeffrey
  Pennington, and Jascha Sohl-Dickstein.
\newblock Deep neural networks as gaussian processes.
\newblock \emph{arXiv preprint arXiv:1711.00165}, 2017.

\bibitem[Li and Liang(2018)]{li_overparameterized}
Yuanzhi Li and Yingyu Liang.
\newblock Learning overparameterized neural networks via stochastic gradient
  descent on structured data.
\newblock \emph{CoRR}, abs/1808.01204, 2018.
\newblock URL \url{http://arxiv.org/abs/1808.01204}.

\bibitem[Maennel et~al.(2018)Maennel, Bousquet, and Gelly]{quantized_hartmut}
Hartmut Maennel, Olivier Bousquet, and Sylvain Gelly.
\newblock Gradient descent quantizes relu network features.
\newblock \emph{arXiv preprint arXiv:1803.08367}, 2018.

\bibitem[Matthews et~al.(2018)Matthews, Rowland, Hron, Turner, and
  Ghahramani]{gaussian_infinite}
Alexander G de~G Matthews, Mark Rowland, Jiri Hron, Richard~E Turner, and
  Zoubin Ghahramani.
\newblock Gaussian process behaviour in wide deep neural networks.
\newblock \emph{arXiv preprint arXiv:1804.11271}, 2018.

\bibitem[Mitchell and Learning(1997)]{mitchell1997mcgraw}
Tom~M Mitchell and Machine Learning.
\newblock Mcgraw-hill science.
\newblock \emph{Engineering/Math}, 1:\penalty0 27, 1997.

\bibitem[Neal(1996)]{neal_priorinfinite}
Radford~M Neal.
\newblock Priors for infinite networks.
\newblock In \emph{Bayesian Learning for Neural Networks}, pages 29--53.
  Springer, 1996.

\bibitem[Neyshabur et~al.(2014)Neyshabur, Tomioka, and Srebro]{neyshabur_reg}
Behnam Neyshabur, Ryota Tomioka, and Nathan Srebro.
\newblock In search of the real inductive bias: On the role of implicit
  regularization in deep learning.
\newblock \emph{arXiv preprint arXiv:1412.6614}, 2014.

\bibitem[Neyshabur et~al.(2018)Neyshabur, Li, Bhojanapalli, LeCun, and
  Srebro]{neyshabur2018overparameterization}
Behnam Neyshabur, Zhiyuan Li, Srinadh Bhojanapalli, Yann LeCun, and Nathan
  Srebro.
\newblock Towards understanding the role of over-parametrization in
  generalization of neural networks.
\newblock \emph{arXiv preprint arXiv:1805.12076}, 2018.

\bibitem[Ongie et~al.(2020)Ongie, Willett, Soudry, and Srebro]{Ongie2020A}
Greg Ongie, Rebecca Willett, Daniel Soudry, and Nathan Srebro.
\newblock A function space view of bounded norm infinite width relu nets: The
  multivariate case.
\newblock In \emph{International Conference on Learning Representations}, 2020.
\newblock URL \url{https://openreview.net/forum?id=H1lNPxHKDH}.

\bibitem[Parhi and Nowak(2019)]{parhi_minimum}
Rahul Parhi and Robert~D Nowak.
\newblock Minimum ``norm'' neural networks are splines.
\newblock \emph{arXiv preprint arXiv:1910.02333}, 2019.

\bibitem[Pilanci and Ergen(2020)]{pilanci2020convex}
Mert Pilanci and Tolga Ergen.
\newblock Neural networks are convex regularizers: Exact polynomial-time convex
  optimization formulations for two-layer networks.
\newblock In Hal~Daumé III and Aarti Singh, editors, \emph{Proceedings of the
  37th International Conference on Machine Learning}, volume 119 of
  \emph{Proceedings of Machine Learning Research}, pages 7695--7705. PMLR,
  13--18 Jul 2020.
\newblock URL \url{http://proceedings.mlr.press/v119/pilanci20a.html}.

\bibitem[Rockafellar(1970)]{Rockafellar}
R.~T. Rockafellar.
\newblock \emph{Convex Analysis}.
\newblock Princeton University Press, Princeton, 1970.

\bibitem[Rosset et~al.(2007)Rosset, Swirszcz, Srebro, and Zhu]{rosset2007}
Saharon Rosset, Grzegorz Swirszcz, Nathan Srebro, and Ji~Zhu.
\newblock L1 regularization in infinite dimensional feature spaces.
\newblock In \emph{International Conference on Computational Learning Theory},
  pages 544--558. Springer, 2007.

\bibitem[Rudin(1964)]{Rudin}
Walter Rudin.
\newblock \emph{Principles of Mathematical Analysis}.
\newblock Mc{G}raw-Hill, New York, 1964.

\bibitem[Sahiner et~al.(2021{\natexlab{a}})Sahiner, Ergen, Ozturkler, Bartan,
  Pauly, Mardani, and Pilanci]{sahiner2021hidden}
Arda Sahiner, Tolga Ergen, Batu Ozturkler, Burak Bartan, John Pauly, Morteza
  Mardani, and Mert Pilanci.
\newblock Hidden convexity of wasserstein gans: Interpretable generative models
  with closed-form solutions, 2021{\natexlab{a}}.

\bibitem[Sahiner et~al.(2021{\natexlab{b}})Sahiner, Ergen, Pauly, and
  Pilanci]{sahiner2021vectoroutput}
Arda Sahiner, Tolga Ergen, John~M. Pauly, and Mert Pilanci.
\newblock Vector-output relu neural network problems are copositive programs:
  Convex analysis of two layer networks and polynomial-time algorithms.
\newblock In \emph{International Conference on Learning Representations},
  2021{\natexlab{b}}.
\newblock URL \url{https://openreview.net/forum?id=fGF8qAqpXXG}.

\bibitem[Savarese et~al.(2019)Savarese, Evron, Soudry, and
  Srebro]{infinite_width}
Pedro Savarese, Itay Evron, Daniel Soudry, and Nathan Srebro.
\newblock How do infinite width bounded norm networks look in function space?
\newblock \emph{CoRR}, abs/1902.05040, 2019.
\newblock URL \url{http://arxiv.org/abs/1902.05040}.

\bibitem[Shankar et~al.(2020)Shankar, Fang, Guo, Fridovich-Keil, Ragan-Kelley,
  Schmidt, and Recht]{shankar_kernels}
Vaishaal Shankar, Alex Fang, Wenshuo Guo, Sara Fridovich-Keil, Jonathan
  Ragan-Kelley, Ludwig Schmidt, and Benjamin Recht.
\newblock Neural kernels without tangents.
\newblock In Hal~Daumé III and Aarti Singh, editors, \emph{Proceedings of the
  37th International Conference on Machine Learning}, volume 119 of
  \emph{Proceedings of Machine Learning Research}, pages 8614--8623. PMLR,
  13--18 Jul 2020.
\newblock URL \url{http://proceedings.mlr.press/v119/shankar20a.html}.

\bibitem[Themelis and Patrinos(2019)]{superscs}
Andreas Themelis and Panagiotis Patrinos.
\newblock Supermann: a superlinearly convergent algorithm for finding fixed
  points of nonexpansive operators.
\newblock \emph{IEEE Transactions on Automatic Control}, 2019.

\bibitem[Thiry et~al.(2021)Thiry, Arbel, Belilovsky, and
  Oyallon]{thiry2021unreasonable}
Louis Thiry, Michael Arbel, Eugene Belilovsky, and Edouard Oyallon.
\newblock The unreasonable effectiveness of patches in deep convolutional
  kernels methods.
\newblock 2021.

\bibitem[Torgo()]{regression}
L.~Torgo.
\newblock Regression data sets.
\newblock \url{http://www.dcc.fc.up.pt/~ltorgo/Regression/DataSets.html}.

\bibitem[T{\"u}t{\"u}nc{\"u} et~al.(2001)T{\"u}t{\"u}nc{\"u}, Toh, and
  Todd]{tutuncu2001sdpt3}
RH~T{\"u}t{\"u}nc{\"u}, KC~Toh, and MJ~Todd.
\newblock Sdpt3—a matlab software package for semidefinite-quadratic-linear
  programming, version 3.0.
\newblock \emph{Web page http://www. math. nus. edu. sg/mattohkc/sdpt3. html},
  2001.

\bibitem[van~den Berg and Friedlander(2007)]{spgl1}
E.~van~den Berg and M.~P. Friedlander.
\newblock {SPGL1}: A solver for large-scale sparse reconstruction, June 2007.
\newblock http://www.cs.ubc.ca/labs/scl/spgl1.

\bibitem[Wei et~al.(2018)Wei, Lee, Liu, and Ma]{margin_theory_tengyu}
Colin Wei, Jason~D Lee, Qiang Liu, and Tengyu Ma.
\newblock On the margin theory of feedforward neural networks.
\newblock \emph{arXiv preprint arXiv:1810.05369}, 2018.

\bibitem[Williams et~al.(2019)Williams, Trager, Silva, Panozzo, Zorin, and
  Bruna]{ntk_1d}
Francis Williams, Matthew Trager, Claudio Silva, Daniele Panozzo, Denis Zorin,
  and Joan Bruna.
\newblock Gradient dynamics of shallow univariate relu networks.
\newblock \emph{arXiv preprint arXiv:1906.07842}, 2019.

\bibitem[Zhang et~al.(2016)Zhang, Bengio, Hardt, Recht, and
  Vinyals]{understanding_zhang}
Chiyuan Zhang, Samy Bengio, Moritz Hardt, Benjamin Recht, and Oriol Vinyals.
\newblock Understanding deep learning requires rethinking generalization.
\newblock \emph{arXiv preprint arXiv:1611.03530}, 2016.

\bibitem[Zhao and Yu(2006)]{zhao2006model}
Peng Zhao and Bin Yu.
\newblock On model selection consistency of lasso.
\newblock \emph{Journal of Machine learning research}, 7\penalty0
  (Nov):\penalty0 2541--2563, 2006.

\end{thebibliography}
\end{document}